\documentclass{article}

\usepackage{microtype}
\usepackage{graphicx}
\usepackage{subcaption}
\usepackage{booktabs} 
\usepackage{makecell}
\usepackage{amsfonts} 
\usepackage{amssymb}
\usepackage{bm}
\usepackage{tcolorbox}
\usepackage{xcolor}
\usepackage{enumitem}
\usepackage{bm}
\usepackage{booktabs}
\usepackage{multirow}
\usepackage{subcaption}
\usepackage{lipsum}
\usepackage{wrapfig}
\usepackage{float}
\usepackage{titlesec}
\usepackage[table]{xcolor}
\usepackage{subcaption} 
\newcommand{\method}{\texttt{NeuroFaith}}

\usepackage{hyperref}

\usepackage[preprint]{icml2026}

\usepackage{amsmath}
\usepackage{amssymb}
\usepackage{mathtools}
\usepackage{amsthm}

\usepackage[capitalize,noabbrev]{cleveref}

\theoremstyle{plain}

\theoremstyle{definition}

\theoremstyle{remark}

\usepackage[textsize=tiny]{todonotes}

\newcommand{\printappendixcontents}{%
  \subsection*{Appendix Table of Contents}
  \small
  \noindent
  \begin{itemize}[leftmargin=*, labelwidth=0pt, labelsep=0pt]
    \itemsep 0.05cm
    \item[\textbf{A.}] \quad \hyperlink{sec:appendix_scientific_library}{Scientific Libraries} \dotfill \pageref{sec:appendix_scientific_library}
    \item[\textbf{B.}] \quad \hyperlink{sec:appendix_llm_implementation_details}{LLM Implementation Details} \dotfill \pageref{sec:appendix_llm_implementation_details}
    \item[\textbf{C.}] \quad \hyperlink{sec:appendix_datasets}{Datasets} \dotfill \pageref{sec:datasets}
    \item[\textbf{D.}] \quad \hyperlink{sec::appendix_implementation_detail}{\method\ Implementation Details} \dotfill \pageref{sec::appendix_implementation_detail}
    \begin{itemize}[leftmargin=1cm]
      \item[\textbf{D.1}] \quad \hyperlink{sec::method_CE}{Concept Extraction} \dotfill \pageref{sec::method_CE}
      \item[\textbf{D.2}] \quad \hyperlink{sec::method_MI}{Mechanistic Interpretation} \dotfill \pageref{sec::method_MI}
      \item[\textbf{D.3}] \quad \hyperlink{sec::linear_appendix}{Linear Latent Faithfulness Detection} \dotfill \pageref{sec::linear_appendix}
    \end{itemize}
    \item[\textbf{E.}] \quad \hyperlink{sec::caract}{Detailed Taxonomy of Self-NLE in Two-hop Reasoning} \dotfill \pageref{sec::caract}
    \begin{itemize}[leftmargin=1cm]
      \item[\textbf{E.1}] \quad \hyperlink{sec::appendix_2_hop_detailed_results}{2-hop reasoning detailed results.} \dotfill \pageref{sec::appendix_2_hop_detailed_results}
    \end{itemize}
    \item[\textbf{F.}] \quad \hyperlink{sec:sota_comparison}{\method\ Faithfulness Measure Comparison} \dotfill \pageref{sec:sota_comparison}
    \begin{itemize}[leftmargin=1cm]
      \item[\textbf{F.1}] \quad \hyperlink{sec:sota_qualitative}{2-hop Reasoning Faithfulness Qualitative Comparison} \dotfill \pageref{sec:appendix_quantitative_comparison}
      \item[\textbf{F.2}] \quad \hyperlink{sec:appendix_quantitative_comparison}{2-hop Reasoning Faithfulness Quantitative Comparison} \dotfill \pageref{sec:appendix_quantitative_comparison}
    \end{itemize}
    \item[\textbf{G.}] \quad \hyperlink{sec::examples_classification}{Classification Examples} \dotfill \pageref{sec::examples_classification}
    \item[\textbf{H.}] \quad \hyperlink{sec::examples}{2-hop Reasoning Taxonomy Examples} \dotfill \pageref{sec::examples}
  \end{itemize}
  \normalsize
}

\icmltitlerunning{Evaluating LLM Self-Explanation Faithfulness via Internal Representation Alignment}

\begin{document}

\twocolumn[
  \icmltitle{\method: Evaluating LLM Self-Explanation Faithfulness via\\Internal Representation Alignment}



  \icmlsetsymbol{equal}{*}

 \begin{icmlauthorlist}
  \icmlauthor{Milan Bhan}{equal,sorbonne,ekimetrics}
  \icmlauthor{Jean-Noël Vittaut}{sorbonne}
  \icmlauthor{Nicolas Chesneau}{ekimetrics}
  \icmlauthor{Sarath Chandar}{mila}
  \icmlauthor{Marie-Jeanne Lesot}{sorbonne}
\end{icmlauthorlist}

\icmlaffiliation{sorbonne}{Sorbonne Université, CNRS, LIP6, F-75005 Paris, France}
\icmlaffiliation{ekimetrics}{Ekimetrics, Paris, France}
\icmlaffiliation{mila}{Mila - Quebec AI Institute, Université de Montréal, Montreal, Canada}

\icmlcorrespondingauthor{Milan Bhan}{milan.bhan@lip6.fr}

  \icmlkeywords{Machine Learning, ICML}

  \vskip 0.3in
]



\printAffiliationsAndNotice{}  

\begin{abstract}
  Large Language Models (LLMs) can generate plausible free text self-explanations to justify their answers. However, these natural language explanations may not accurately reflect the model's actual reasoning process, pinpointing a lack of faithfulness. Existing faithfulness evaluation methods rely primarily on behavioral tests or computational block analysis without examining the semantic content of internal neural representations. This paper proposes \method, a flexible framework that measures the faithfulness of LLM free text self-explanation by identifying key concepts within explanations and mechanistically testing whether these concepts actually influence the model's predictions. We show the versatility of \method\ across 2-hop reasoning and classification tasks. Additionally, we develop a linear faithfulness probe based on \method\ to detect unfaithful self-explanations from representation space and improve faithfulness through steering. \method\ provides a principled approach to evaluating and enhancing the faithfulness of LLM free text self-explanations, addressing critical needs for trustworthy AI systems.
\end{abstract}

\section{Introduction}

Autoregressive Large Language Models (LLMs) can generate plausible self Natural Language Explanation (self-NLE) to support their answers~\citep{teach_me_explain,huang_can_2023}. Generating self-NLE consists of prompting the LLM to output an explanation in a \textit{predict-then-explain} setting, where the model first generates a response to a question and then produces a self-NLE as a justification. Unlike their non-generative predecessors, modern LLMs are trained to generate both answers and free text self-NLE that appear credible despite potentially containing persuasive hallucinations~\citep{sahoo2024comprehensive}. This way, despite their logical and coherent appearance favoring trust in the model~\citep{trust_faithfulness}, LLM-generated self-NLE turn out to not systematically reflect the actual underlying decision-making process of the model, creating a tension between self-NLE \textit{plausibility} and \textit{faithfulness}~\citep{faithfulness_plausibility}.

\begin{figure}[t]
    \includegraphics[width=0.48\textwidth]{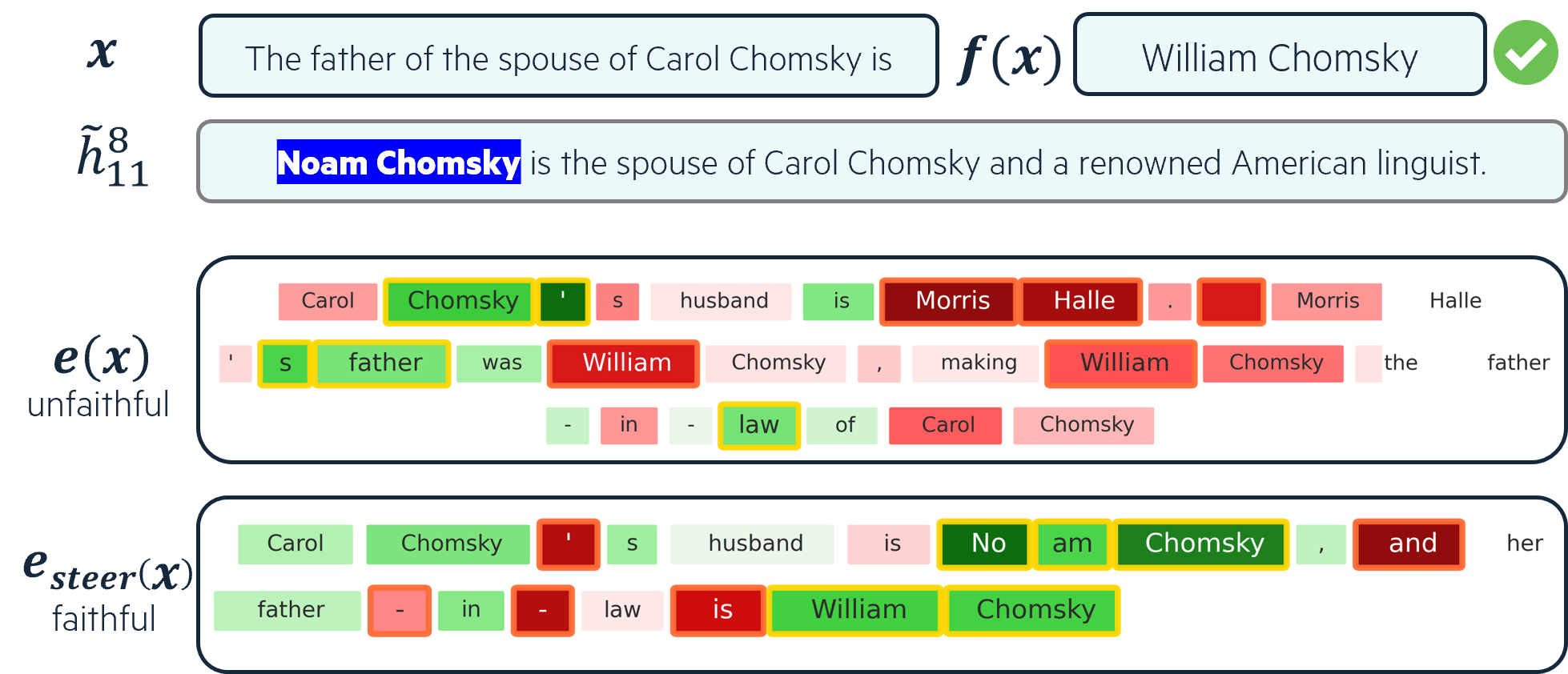}
    \caption{2-hop reasoning example with LLM answer and decoded hidden representation ($\tilde{h}$). Faithfulness linear probe is applied before ($e$) and after ($e_{steer}$) steering. Red: unfaithful activations (e.g. Morris Halle); Green: faithful activations (e.g. Noam Chomsky).}
    \label{fig:example}
\end{figure}

Faithfulness, as defined by \citet{jacovi2020towards}, measures "how accurately the explanation reflects the true reasoning process of the model", a definition widely adopted in the literature~\citep{XAI_nlp_survey} and which we also follow throughout this work.  Unfaithful self-NLE can have serious consequences in critical domains~\citep{farah2023assessment}, where explanations that appear plausible but lack faithfulness might lead end-users to over-rely on model predictions and make unfair~\citep{luo2022evaluating} or harmful~\citep{kayser2024fool} decisions. As the use of LLMs is expanding across diverse fields, the combination of their widespread adoption and the simplicity of generating self-NLE through prompting make evaluating their faithfulness increasingly critical.

Assessing self-NLE faithfulness presents profound difficulties due to the free-form nature of natural language explanations, unlike more structured explainability methods such as attribution~\citep{wiegreffe_faithfulness} or counterfactual approaches~\citep{self_consistency_madsen}. Numerous methods have been developed to measure self-explanation faithfulness (see e.g.~\citet{XAI_nlp_survey}). However, (1) they mostly perform behavioral tests and do not examine LLM internal reasoning processes~\citep{atanasova_faithfulness_2023,siegel_faithfulness,siegel_faithfulness_V2,walk_the_talk_faithfulness}, and (2) they identify the computational blocks that contribute to prediction and self-NLE without conducting semantic analysis of the neural representations within these blocks~\citep{wiegreffe_faithfulness,parcalabescu_faithfulness,yeo_faithfulness}. These shortcomings have led these methods to be characterized as measuring \textit{self-consistency} rather than genuine faithfulness~\citep{parcalabescu_faithfulness}, since they fail to establish direct connections between explanations and the model's reasoning processes. To overcome these limitations, we introduce \method, a flexible framework for directly measuring LLM self-NLE faithfulness. Our main contributions are as follows: \begin{enumerate}[noitemsep,topsep=0pt,leftmargin=*]
    \item \method\ measures the alignment between LLM internal reasoning and its self-explanations by identifying key concepts within the explanation and mechanistically testing whether these concepts actually influence the model's predictions.
    \item We demonstrate the versatility of \method\ by applying it to both 2-hop reasoning and classification tasks.
    \item We develop a linear faithfulness probe based on \method\ to efficiently detect unfaithful LLM self-explanations from representation space and improve faithfulness through steering.
\end{enumerate}

"Figure~\ref{fig:example} illustrates \method\ on a 2-hop reasoning question: despite answering correctly, the model's initial self-NLE ($e$) is unfaithful. Linear steering improves faithfulness by aligning the steered self-NLE ($e_{steer}$) with the decoded internal concept (Noam Chomsky) being required for reasoning solving. This paper is organized as follows: Section~\ref{sec:mot_rl} presents how existing approaches measure the faithfulness of LLM self-NLE. Section~\ref{sec:framework} introduce the \method\ framework. In Sections~\ref{sec:2_hop} and~\ref{sec:classif}, we instantiate \method\ for two different tasks, 2-hop reasoning and classification respectively.  We finally show in Section~\ref{sec:linear_detection} that self-NLE faithfulness, as measured by \method, can be linearly detected in LLM representation space and improved through steering.

\section{Related Work}
\label{sec:mot_rl}

Numerous approaches have been proposed to measure self-NLE faithfulness~\citep{XAI_nlp_survey}. One approach~\citep{tutek_faithfulness} assesses the effect of unlearning~\citep{liu2025rethinking} the parametric knowledge encoded in the reasoning steps of the self-NLE. The higher the change in prediction between the original model and the model having unlearned the reasoning steps, the more faithful the self-NLE. We group the remaining approaches in two categories.

\paragraph{Counterfactual Interventions.} NLE faithfulness can be assessed through behavioral tests that measure how perturbations in the input text affect both predictions and self-NLE~\citep{atanasova_faithfulness_2023}. Counterfactual Intervention (CI) methods employ auxiliary models to generate counterfactual texts designed to change the LLM outcome. The LLM is then prompted to produce a self-NLE to justify its new prediction. The self-NLE is deemed faithful if it aligns with the specific CI that caused the prediction change. These CI approaches mostly differ in two ways: how they measure consistency between the intervention and the resulting self-NLE~\citep{atanasova_faithfulness_2023,siegel_faithfulness,siegel_faithfulness_V2}, and the granularity of the CI intervention~\citep{walk_the_talk_faithfulness}. These approaches face several limitations: (1) the CI may not be solely responsible for the change in prediction, as the model might base its new prediction on another part of the input text after intervention, (2) CI methods treat the model as a black box by only examining input-output relationships without analyzing the internal neural processes that generate predictions, which departs from the commonly adopted definition of explanation faithfulness.

\paragraph{Attribution Agreement.} Another way to measure self-NLE faithfulness is to compute post-hoc Attribution Agreement (AA) between the prediction and the NLE~\citep{parcalabescu_faithfulness, wiegreffe_faithfulness,yeo_faithfulness}. AA methods compute attribution scores for both the model's predictions and its self-NLE and then measure the correlation between these scores to assess faithfulness. Higher correlation values indicate greater faithfulness in the model's self-NLE. AA approaches vary in the post-hoc attribution method employed (gradient-based~\citep{sundararajan_axiomatic_2017}, SHAP~\citep{lundberg_unified_2017} or activation patching~\citep{meng2022locating}). While AA methods assess whether the same LLM computational blocks were used when generating both the prediction and the self-NLE, they overlook the semantic content of neural representations, leaving the model's reasoning process only partially treated.

In the following, we propose a framework that directly examines the correspondence between self-NLE and the model's actual reasoning process by conducting concept-level mechanistic analysis of internal hidden states during the forward pass that generates the prediction.

\section{\method: A Framework for Measuring the Faithfulness of LLM self-NLE}
\label{sec:framework}
This section introduces the core principles of \method, our proposed flexible framework for measuring the faithfulness of LLM self-natural language explanations (self-NLE). Based on the premise that faithful explanations should accurately reflect the model's internal reasoning process, \method\ quantifies how well a self-NLE aligns with the model's internal reasoning processes by extracting concepts from self-NLE and mechanistically evaluating their importance for the model prediction. Sections~\ref{sec:2_hop} and~\ref{sec:classif} provide detailed instantiations of how to apply \method\ to 2-hop reasoning and classification tasks respectively. In the following, we denote an $L$-layer auto-regressive Transformer-based LLM $f$, a set of input texts \textbf{X} and a text of interest $x \in \textbf{X}$. We denote $f(x)$ the model's answer and $e(x)$ the corresponding self-NLE produced by $f$ for input $x$.  
\method\ is a 3-step framework illustrated in Figure~\ref{fig:overall}, summarized below and detailed in the next subsections. 

\begin{enumerate}[noitemsep,topsep=0pt,leftmargin=*]
    \item \textbf{Concept Extraction}. We extract from $e(x)$ a set of concepts $\left\{c_{i}\right\}_{i=1}^{p}$  quoted as important for $f(x)$.
    \item \textbf{Concept-wise Mechanistic Interpretation}. For each concept $c_{i}$, we generate a post-hoc interpretation $\texttt{I}_{\Gamma}(c_{i})$ based on a relevant subpart of the model called circuit~$\Gamma$ and a mechanistic interpretability method called interpreter $\texttt{I}$ to assess its impact on $f(x)$.
    \item \textbf{Faithfulness Measurement}. We compute faithfulness $F(x,e)$ by validating that the detected concepts $\left\{ c_{i} \right\}_{i=1}^{p}$ in the self-NLE have mechanistic effects on $f(x)$ as determined by their interpretations $\left\{ \texttt{I}_{\Gamma} (c_i) \right\}_{i=1}^{p}$.
\end{enumerate}

\subsection{Concept Extraction}
\label{sec:concept_extraction}
Given an input text $x$, a prediction $f(x)$ and its self-NLE $e(x)$, the first step involves parsing the self-NLE to extract a set of concepts $\left\{c_{i}\right\}_{i=1}^{p}$ that influence $f(x)$.

\begin{figure*}[t]{\centering}
\begin{center}
\includegraphics[width=\linewidth]{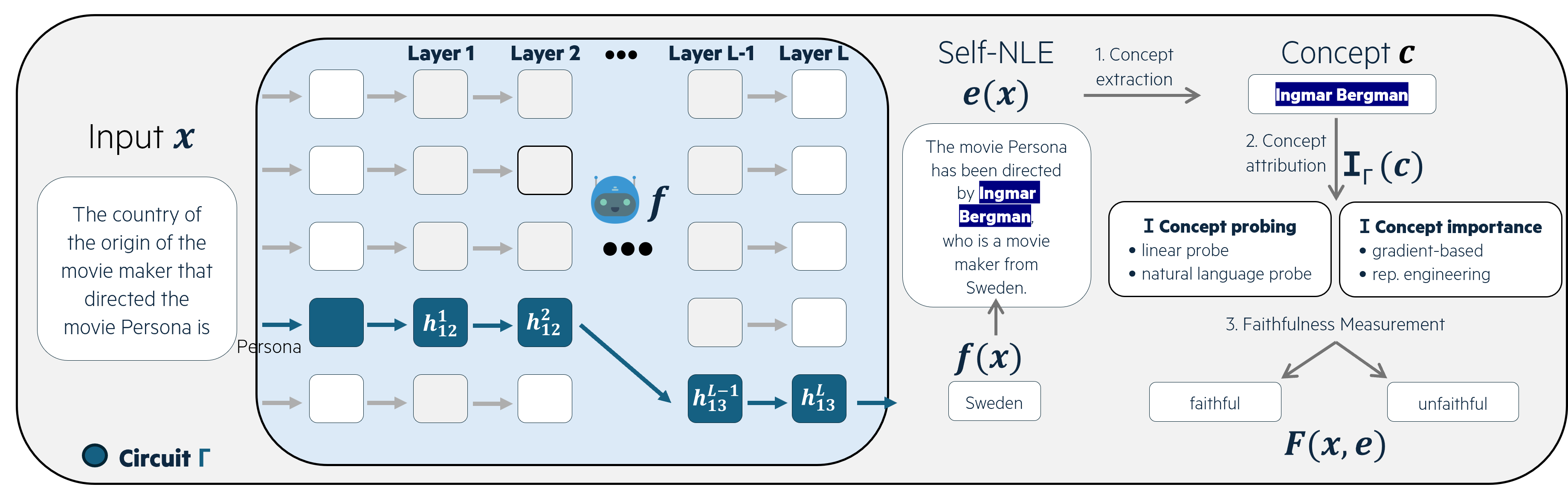}
\caption{\method\ overview. \method\ (1) extracts concepts from the self-NLE (Section~\ref{sec:concept_extraction}), (2) assesses the mechanistic influence of these concepts (Section~\ref{sec::mechanistinc_interpret}) to finally (3) measure faithfulness (Section~\ref{sec:faith_measure}).}  
\label{fig:overall}
\end{center}
\end{figure*}

\textbf{Concept Definition.} Recent research has shown that interpretable binary high-level features, referred to as \textit{concepts}, appear to be linearly encoded within LLM representation space~\citep{elhage_linear_representation,linear_rep_hypothesis}. This computational understanding aligns with definitions from cognitive science~\citep{ruiz2024theoretical}, where concepts are considered as mental entities essential to thought, enabling information integration and categorization~\citep{what_is_a_concept}. This alignment makes concepts an ideal granularity level for model interpretation. For example in Figure~\ref{fig:overall}, the concept "\textit{Ingmar Bergman}" directly influences the prediction "\textit{Sweden}" because Ingmar Bergman is Swedish and directed the movie Persona. 

\textbf{Concept Extraction.} Concepts can be extracted either through human investigation or using an auxiliary LLM under LLM-as-a-Judge settings~\citep{gu2024survey}. Human annotation enables targeted analysis of specific concepts that are hypothesized to be important for the model's reasoning. Human annotation provides high-quality, domain-expert identification of relevant influential concepts but is costly and may not scale to large datasets. LLM-as-a-Judge approaches offer scalability but may  miss relevant concepts that human experts would identify. We provide detailed examples of prompts used to extract concepts from self-NLE with an auxiliary LLM in Appendix~\ref{sec::method_CE}.

\subsection{Concept-wise Mechanistic Interpretation}
\label{sec::mechanistinc_interpret}

The second step evaluates whether the concepts $\left\{c_{i}\right\}_{i=1}^{p}$ are actually important for the model's internal processing. We compute an attribution score $\texttt{I}_{\Gamma}(c)$ for each concept $c \in \left\{c_{i}\right\}_{i=1}^{p}$ using an interpretability method (interpreter~$\texttt{I}$) applied to model hidden states within a relevant subpart of the model (circuit $\Gamma$). We assume each concept can be detected in the model representation space.

\paragraph{Circuit.} Rather than examining every network component, we focus our mechanistic analysis on computational subgraphs that most significantly influence $f$ prediction named \textit{circuits}~\citep{elhage2021mathematical}. Circuits are sparse oriented subgraphs with interpretable functional roles, where nodes represent computational units and edges represent computation paths~\citep{rauker2023toward}. Formally, we define a circuit as $\Gamma = (\left \{ (k,\ell)\right \},  \mathcal{G})$ where $\left \{ (k,\ell)\right \}$ specifies coordinate pairs (token index, layer) for information flow in $f$, and $\mathcal{G}$ defines the granularity of each computational node. The granularity $\mathcal{G}$ represents transformer sublayers: residual stream output (\texttt{RS}), multi-head attention (\texttt{MHA}), or multi-layer perceptron (\texttt{MLP})~\citep{rai2024practical}. This granularity avoids focusing on individual neurons due to their polysemanticity~\citep{elhage_linear_representation}. For example in Figure~\ref{fig:overall}, the circuit is highlighted in blue, starting at token index 12 and layer 1 and finishes at token index 13 and layer  L. Circuits can be obtained through manual investigation of task-specific patterns~\citep{MI_survey,wanginterpretability,hopping_too_late_multi_hop} or automated discovery via activation patching~\citep{meng2022locating,acdc}, backward attribution~\citep{ferrando2024information}, or Transcoders~\citep{dunefskytranscoders}. Additional details about circutis are in Appendix~\ref{sec::method_MI}.

\paragraph{Mechanistic Interpreter.} To mechanistically interpret a concept $c$ and its impact on $f(x)$ across circuit $\Gamma$, we use an interpreter \texttt{I} to compute an attribution score $\texttt{I}_\Gamma(c)$ that evaluates whether $c$ impacts $f(x)$. \method\ implements two approaches, each suited to different objectives. \textbf{Probing} methods determine whether $c$ is represented within $f$ hidden states in $\Gamma$. This involves either generating natural language interpretations of hidden states (using \texttt{Selfie}~\citep{selfie} or \texttt{Patchscopes}~\citep{patchscopes}) to detect $c$, or training linear probes~\citep{belinkov2022probing} when $c$ is linearly separable with sufficient labels to train the probe. Denoting $h^\ell_k$ the hidden state 
at token $k$ and layer $\ell$, we define $\texttt{I}_{\Gamma}(c) = \max_{(k,\ell) \in \Gamma} \texttt{I}(h^\ell_k)$, assessing $c$ as important if detected in at least one hidden state from $\Gamma$. \textbf{Concept importance} methods approximate the causal influence of $c$ on $f(x)$~\citep{geiger2025causal}. They use gradient-based methods like \texttt{TCAV}~\citep{tcav} or Representation Engineering~\citep{zou2023representation} to directly manipulate concept-related activations across $\Gamma$ and measure behavioral changes, providing causal evidence for concept importance. Section~\ref{sec:classif} details this procedure in the case of classification.

\subsection{Faithfulness Measurement}
\label{sec:faith_measure}

We consider the self-NLE $e(x)$ as faithful when the extracted concepts $\left\{ c_i \right\}_{i=1}^{p}$ are demonstrably important for $f$'s  internal processing according to $\texttt{I}_\Gamma(c)$. 
We propose to define the faithfulness of $e(x)$ as the proportion of concepts it contains being mechanistically important, i.e. having  positive attribution score: $F(x,e) = \frac{1}{p} \sum_{i=1}^{p} \mathbf{1}_{\texttt{I}_\Gamma(c_i) > 0}$. 

The faithfulness score has different meanings depending on the chosen interpreter \texttt{I}: \textbf{Probing-based faithfulness} measures the detection rate of explanation concepts within the model's internal representations along circuit $\Gamma$. High probing-based faithfulness indicate that most mentioned concepts are properly decoded from the model's hidden states. \textbf{Importance-based faithfulness} estimate causal relevance of explanation concepts for the prediction $f(x)$. High importance-based faithfulness indicate that most mentioned concepts actually influence $f$ reasoning process. This framework directly captures concept-level alignment between the self-NLE $e(x)$ and $f$ internal reasoning process. \method\ faithfulness scores either indicate that the self-NLE accurately reflects internal processing or that the explanation mentions concepts that are not mechanistically important. This way, \method\ directly aligns with the established definition of explanation faithfulness introduced above~\citep{jacovi2020towards}, offering a principled evaluation of self-NLE faithfulness.

\begin{table*}[t]
\centering
\caption{2-hop reasoning accuracy, self-NLE correctness, latent first hop correctness and self-NLE faithfulness across models obtained from \method. on the the Wikidata-2-hop dataset. "(In)accurate" represents (in)correct predictions.}
\small
\begin{tabular*}{\textwidth}{@{\extracolsep{\fill}}lcccccccc@{}}
\toprule
\textbf{Model} & \textbf{Task Accuracy} & \multicolumn{2}{c}{\textbf{Self-NLE}} & \multicolumn{2}{c}{\textbf{Latent Hop 1}} & \multicolumn{2}{c}{\textbf{Self-NLE}} \\
& & \multicolumn{2}{c}{\textbf{Correctness}} & \multicolumn{2}{c}{\textbf{Correctness}} & \multicolumn{2}{c}{\textbf{Faithfulness}} \\
\cmidrule(lr){3-4} \cmidrule(lr){5-6} \cmidrule(lr){7-8}
& & Accurate & Inaccurate & Accurate & Inaccurate & Accurate & Inaccurate \\
\midrule
\texttt{gemma-2-2b}  & 7.1\% & 58.0\% & 56.0\% & 48.3\% & 47.4\% & 48.4\% & 57.6\% \\
\texttt{gemma-2-9b}  & 16.6\% & 74.0\% & 55.4\% & 58.5\% & 44.3\% & 60.8\% & 55.0\% \\
\texttt{gemma-2-27b} & 22.3\% & 77.8\% & 55.6\% & 64.9\% & 46.9\% & 68.8\% & 60.2\% \\
\texttt{mistral-3-7b} & 21.2\% & 70.2\% & 37.9\% & 52.9\% & 29.6\% & 51.1\% & 32.6\% \\
\bottomrule
\end{tabular*}
\label{tab:results_2_hop_reasoning}
\end{table*}

\section{The Case of 2-Hop Reasoning}
\label{sec:2_hop}
In the previous section, we defined the high level core principle of \method. We now instantiate \method\ to 2-hop reasoning using probing-based faithfulness.

\subsection{\method\ Instantiation for 2-Hop Reasoning}
\label{sec:2_hop_desc}

\paragraph{Task Description.}
Multi-hop reasoning is a complex cognitive task that requires connecting a sequence of objects to reach a conclusion~\citep{mavi2024multi}. It consists of several single-hop operations~\citep{trivedi-etal-2022-musique}, which can individually be defined as triplets $(o_{i},r,o_{j})$ where $o_{i}$ is a source object, $r$ is a relation and $o_j$ is a target object. For example, the 2-hop reasoning statement "\textit{The country of origin of the movie maker that directed the movie Persona is Sweden}" requires sequentially solving the two single-hop operations: ($o_1= \texttt{personna}$, $r_1= \texttt{movie direction}$, $o_2=\texttt{ingmar bergman})$ and ($o_2 =  \texttt{ingmar bergman}$, $r_2 = \texttt{country}$, $o_3 = \texttt{sweden}$). An input text $x$ that requires performing 2-hop reasoning can be expressed as $x= (o_{1},r_1,\tiny{\blacktriangle{}},r_2,\bullet)$, where $\tiny{\blacktriangle{}}$ and $\bullet$ are placeholders that have to be associated with a bridge object $(\widehat{o_{2}})$ and the final object answer $(f(x) = \widehat{o_{3}})$. 

\paragraph{Concept Extraction.}
Following the notations introduced in Section~\ref{sec:framework}, given an input text $x \in \textbf{X}$, if $\widehat{o_{3}}=o_3$, the final answer is correct. Two reasoning chains are derived from $e(x)$: $(o_{1},r_1,\widehat{o_{2}})$ and $(\widehat{o_{2}},r_2,\widehat{o_{3}})$. In this instantiation, we do concept extraction from $e(x)$ by prompting an auxiliary LLM to get the bridge object $(\widehat{o_{2}})$ based on $o_1$ and $r_1$. This way, $\widehat{o_{2}}$ is our concept $c$ of interest, representing the critical intermediate step that connects the two reasoning operations. The next step consists in computing a post-hoc mechanistic interpretation to assess $c$ impact on the prediction.


\paragraph{Probing-based Concept Interpretation.} 
The presence of a single bridge object in 2-hop reasoning self-NLE makes probing-based interpretations particularly appropriate. For $f$ to correctly answer, it must either internally compute the bridge object during the first hop before executing the second hop or leverage shortcuts to directly answer based on $o_1$. Therefore, detecting the extracted bridge object $\widehat{o_{2}}=c$ in $f$ internal representations is a strong indication to assess whether $c$ is important for the prediction. We employ natural language interpretation methods such as \texttt{Selfie} and \texttt{Patchscopes}, rather than linear probes, as they provide higher accuracy and are unsupervised~\citep{patchscopes}. Recent work~\citep{hopping_too_late_multi_hop} has shown that the bridge object of 2-hop reasoning was resolved on early layers and can be detected on the last token representation of source object $o_{1}$ and the residual stream (\texttt{RS}). We leverage this information to define circuit $\Gamma$ and generate the natural language description $\tilde{h^\ell_k}$ of hidden state $h^\ell_k$. For concept (bridge object)~$c$, the probing-based concept attribution is defined as $\texttt{I}_{\Gamma}(c) = 1$  iff $\exists (k,\ell) \in \Gamma$ such that $c \in \tilde{h^\ell_k}$.

\paragraph{Self-NLE and Latent Reasoning Characterization.} Following the notations introduced in Section~\ref{sec:faith_measure}, $e(x)$ is assessed faithful if $c$ is detected in at least one hidden state within $\Gamma$ with the probe \texttt{I} ($\texttt{I}_{\Gamma}(c)$=1), implying that $F(x,e) \in \left\{ 0,1\right\}$ for 2-hop reasoning. Beyond faithfulness (\textbf{self-NLE faithfulness}), we can use the ground truth object ($o_2$) to characterize the correctness of $e(x)$ and the latent first hop respectively: 
$e(x)$ is said to be correct (\textbf{self-NLE correctness}) if its corresponding bridge object is the expected ground truth ($c=o_2$). Likewise, the model performs correct first-hop latent reasoning (\textbf{latent hop 1 correctness}) if $\exists (k,l) \in \Gamma$ such that $ o_{2} \in \tilde{h^\ell_k}$. This multi-dimensional analysis provides comprehensive characterization of explanation quality and internal reasoning alignment. We give a thorough taxonomy of self-NLE in the case of 2-hop reasoning in Appendix~\ref{sec::caract}, based on prediction correctness, self-NLE faithfulness and correctness and latent reasoning correctness.

\subsection{2-Hop Reasoning Experimental Analysis}
\paragraph{Experimental Setup.} 

We evaluate \method\ on two 2-hop reasoning datasets: Wikidata-2-hop~\citep{hopping_too_late_multi_hop} and the more challenging Socrates~\citep{yang2025large}. For Wikidata-2-hop, we test \texttt{Gemma-2} (2B, 9B, 27B)~\citep{team2024gemma}, and \texttt{Mistral-7B-Instruct-v0.3}~\citep{jiang2023mistral}; for Socrates, we restrict evaluation to \texttt{Gemma-2-27B} given its increased reasoning difficulty. We use \texttt{Qwen3-32B} to extract bridge objects from self-NLEs and \texttt{Patchscopes}~\cite{patchscopes} as interpreter. We report Latent Hop 1 Correctness, Self-NLE Faithfulness, and Self-NLE Correctness, stratified by prediction accuracy. Details on circuit $\Gamma$ and prompts to extract the bridge object and run \texttt{Patchscopes} are provided in Appendices~\ref{sec::method_CE} and~\ref{sec::method_MI}.

Since \method\ relies on \texttt{Qwen3-32B} to extract bridge objects, we evaluate its reliability for this task. When prompted to retrieve bridge objects from complete 2-hop reasoning chains ($o_1$, $r_1$, $o_2$) and ($o_2$, $r_2$, $o_3$), the latter achieves 99.5\% accuracy. We also report results using \texttt{Phi-4}~\citep{abdin2024phi} as an alternative extractor in Appendix\ref{sec::method_MI}, Table~\ref{tab:results_faithfulness_analysis_phi}, showing approximately 90\% correlation with \texttt{Qwen3-32B}. These findings confirm that \method\ is robust to the choice of auxiliary model and concepts are reliably extracted.

\paragraph{Key Findings.} Table~\ref{tab:results_2_hop_reasoning} shows the aggregated experimental results obtained by applying \method\ on Wikidata-2-hop. All the analyzed metrics are higher on average for accurate predictions and improve with model size. These results support that correct reasoning tend to produce more faithful and correct self-NLE. Key failure patterns emerge with \texttt{Gemma}: approximately 45\% of inaccurate predictions correctly resolve the first hop operation, indicating that failure often occurs in the second reasoning step ($o_2,r_2,o_3$). This suggests that models can identify correct bridge objects even when failing to complete the full reasoning chain. The difference between latent hop 1 correctness and self-NLE correctness shows that models sometimes identify correct bridge objects internally but fail to express them in their self-NLE. This gap decreases with model size, suggesting better alignment between internal and explicit reasoning in larger models. Moreover, self-NLE correctness and faithfulness are comparable between \texttt{Mistral} and \texttt{Gemma} for accurate predictions but substantially lower for \texttt{Mistral} on inaccurate ones.

\begin{figure}[t]{\centering}
\begin{center}
\includegraphics[width=0.7\linewidth]{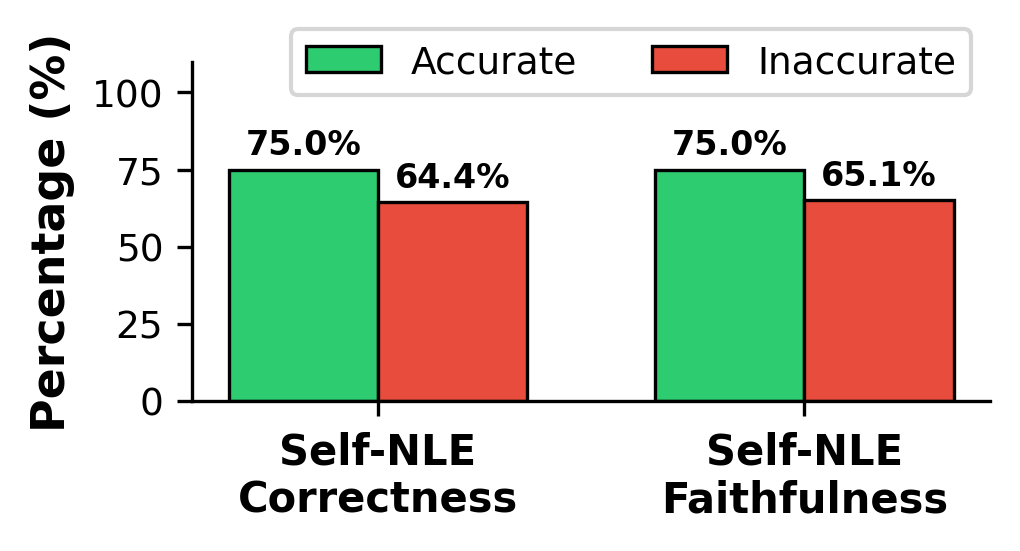}
    \caption{\method\ self-NLE correctness and faithfulness on Socrates on instances with accurate first hop resolution.}
    \label{fig:socrates_faithfulness}
\end{center}
\end{figure}

Remarkably, even when models produce correct final answers, they identify the correct bridge object in their internal representations only 48-65\% of the time. This suggests that models can arrive at correct answers through alternative reasoning pathways that bypass explicit bridge object computation. The models may be leveraging (1) direct associations between source and target entities (shortcut learning~\citep{geirhos2020shortcut}), or (2) alternative pathways making the model accurately answer for a reason different than the ground truth bridge object (see category 9 in Figure~\ref{fig:taxonomy}). Figure~\ref{fig:socrates_faithfulness} shows the results obtained by applying \method\ on Socrates, focusing on predictions with accurate latent first hop reasoning. Like the Wikidata-2-hop dataset, accurate predictions lead to more faithful and accurate self-NLE. Table~\ref{tab:results_faithfulness_analysis_socrates} in Appendix~\ref{sec::method_MI} shows the complete results obtained from the Socrates dataset. 

We additionally compare \method\ to two commonly used faithfulness measures: CI~\citep{atanasova_faithfulness_2023} and AA~\citep{wiegreffe_faithfulness}. For this comparison, we use an In-Context Editing evaluation protocol, extending the approach proposed by~\citet{zaman-srivastava-2025-causal}. Results in Appendix~\ref{sec:sota_comparison} (Table~\ref{tab:eval_overall_faithfulness_sota}) show that \method\ outperforms these baselines in 6 out of 9 settings. Finally, Appendix~\ref{sec::method_MI} presents a sensitivity analysis varying circuit size. Figures~\ref{fig:faithfulness_vs_circuit_size_2b}-\ref{fig:faithfulness_vs_circuit_size_27b} show that faithfulness scores remain consistent across circuit sizes, with only a 10\% gap between single-node and full circuits, demonstrating robustness to circuit specification.

\section{The Case of Classification}
\label{sec:classif}
This section instantiates \method\ for classification by employing concept importance methods as mechanistic interpreter \texttt{I}. Given an input text $x \in \textbf{X}$, we denote $f(x) = \hat{y} \in \mathcal{Y}$ with probability score $p_{\hat{y}}(x)$ where $\mathcal{Y}$ is the label space. We assume access to a predefined set of task-relevant concepts $\mathcal{C}$ with concept labels for input texts.

\subsection{\method\ Instantiation for Classification}

\paragraph{Concept Extraction.}
To enable mechanistic interpretation of each concept $c \in \mathcal{C}$, we compute Concept Activation Vectors (CAVs) that represent concepts as directions in $f$ representation space. Following established practices in concept-based interpretability, we employ the mean difference (\texttt{diff-mean})~\citep{steering_mean_embedding} approach for CAV computation due to its optimal balance between concept detection accuracy and computational efficiency~\citep{wuaxbench}. For a concept $c \in \mathcal{C}$, a token index $k$ and a layer $\ell$, the layer-wise CAV is defined as: $\overrightarrow{c_\ell} = \frac{1}{\left|\textbf{X}^{+}_{c}\right|} \sum_{x\in \textbf{X}^{+}_{c}}{h^{\ell}_{k}} - \frac{1}{\left|\textbf{X}^{-}_{c}\right|} \sum_{x\in \textbf{X}^{-}_{c}}{h^{\ell}_{k}}$, where $\textbf{X}^{+}_{c}$ and $\textbf{X}^{-}_{c}$ respectively represent the sets of texts from $\textbf{X}$ where the concept $c$ is present or absent and $h^{\ell}_{k}$ denotes the hidden state at layer $\ell$ and token position $k$ at the specified granularity $\mathcal{G}$. We set the token index to the final position of~$x$, as this location represents $f$'s complete computational state prior to next-token generation. These CAVs allow for computing the importance of each concept using Representation Engineering techniques. Following Section~\ref{sec:framework}, we use an auxiliary LLM to extract a set of concepts from $e(x)$, making sure that $\left\{c_{i}\right\}_{i=1}^{p} \subset \mathcal{C}$.

\paragraph{Importance-based Concept Interpretation.} For a concept $c \in \left\{c_{i}\right\}_{i=1}^{p}$, we approximate its causal influence on $f(x)$ through representation engineering and concept erasure~\citep{belrose2023leace}. We perform controlled interventions by erasing $c$ from hidden states during forward propagation: $h^{\ell}_{k}\leftarrow h^{\ell}_{k}-\lambda \times \overrightarrow{c_\ell}$, where $\lambda \in [0,1]$ represents intervention intensity and with $\lambda=1$ represents maximum intervention intensity to avoid $f$ collapse~\citep{steering_mean_embedding}. This approach provides strong causal evidence for concept importance without requiring computationally costly gradient computations such as \texttt{TCAV}. We define $\Gamma$ with granularity $\mathcal{G}$ set to residual stream (\texttt{RS}), token index as the final position (likewise CAVs), and layers are selected based on concept detectability (F1 score $>$ 60\% using layer-wise linear probes with \texttt{diff-mean}). Applying this intervention across circuit $\Gamma$, we measure the resulting probability score: $p_{\hat{y}}(x, \left\{ \text{do}(H^{\ell}_{k} =  h^{\ell}_{k}-\lambda \times \overrightarrow{c_\ell}) \right\}_{(k,\ell)\in \Gamma})$, where the $\text{do}(X = x)$ operator~\citep{pearl2009causality} represents an intervention that sets variable $X$ to value $x$. We model this relationship as linear: $p_{\hat{y}}(\cdot) = \beta_0 + \beta_1 \times \lambda$, where the concept importance score is $\texttt{I}_\Gamma(c) = \beta_1$ (i.e. the marginal effect of intervention on probability score). $\texttt{I}_\Gamma(c)$ is set to 0 when its significance $t$-test shows $p > 0.01$.  This process is repeated for each extracted concept from $e(x)$ to derive faithfulness scores as described in Section~\ref{sec:faith_measure}, with $F(x,e) \in \left [0,1 \right ]$ for classification tasks. Details and examples are provided in Appendices~\ref{sec::method_MI} and~\ref{sec::examples_classification}.

\begin{table*}[t]
\centering
\caption{Classification accuracy and self-NLE faithfulness from \method. "(In)accurate represents (in)correct predictions.}
\small
\begin{tabular*}{\textwidth}{@{\extracolsep{\fill}}lcccccc@{}}
\toprule
\textbf{Model} & \multicolumn{2}{c}{\textbf{Task Accuracy}} & \multicolumn{4}{c}{\textbf{Self-NLE Faithfulness}} \\
\cmidrule(lr){2-3} \cmidrule(lr){4-7}
& AGNews & Ledgar & \multicolumn{2}{c}{AGNews} & \multicolumn{2}{c}{Ledgar} \\
\cmidrule(lr){4-5} \cmidrule(lr){6-7}
& & & Accurate & Inaccurate & Accurate & Inaccurate \\
\midrule
\texttt{gemma-2-2B}  & 86.5\% & 40.3\% & 54.6\% & 65.1\% & 58.9\% & 40.5\% \\
\texttt{gemma-2-9B}  & 88.7\% & 59.7\% & 96.0\% & 92.3\% & 56.0\% & 58.3\% \\
\texttt{gemma-2-27B} & 88.9\% & 58.1\% & 74.0\% & 72.4\% & 53.0\% & 31.6\% \\
\texttt{mistral-3-7B} & 80.6\% & 82.9\% & 86.6\% & 84.1\% & 94.8\% & 84.4\% \\
\bottomrule
\end{tabular*}
\label{tab:results_classification}
\end{table*}

\subsection{Classification Experimental Analysis}

\paragraph{Experimental Setup.} We evaluate \method's classification instantiation on two datasets with varying complexity: AGNews~\citep{AG_news}, a newspaper article classification and Ledgar~\citep{tuggener2020ledgar}, a more challenging critical domain legal document classification dataset. We apply \method\ to \texttt{Gemma-2} (2B, 9B, 27B)~\citep{team2024gemma} and \texttt{Mistral-7B-Instruct-v0.3}~\citep{jiang2023mistral}. We use the concept set $\mathcal{C}$ and labels of AGNews and Ledgar from \citet{bhan2025towards} to compute the CAVs. We use \texttt{Qwen3-32B} to extract concepts from the self-NLE, ensuring extracted concepts belong to $\mathcal{C}$ for each dataset. We focus on instances solely containing concepts that are linearly detectable in at least one layer.  We evaluate \textbf{Self-NLE Faithfulness} according to the status of the prediction preceding the self-NLE (accurate vs. inaccurate).

\paragraph{Key Findings.}
Table~\ref{tab:results_classification} shows experimental results for AGNews and Ledgar datasets. Higher average self-NLE faithfulness is observed with \texttt{Gemma} on AGNews and \texttt{Mistral} on Ledgar, supporting that explanation faithfulness might correlate with task accuracy. Accurate predictions generally yield more faithful explanations, suggesting that correct predictions might rely on relevant concepts. Notably, \texttt{gemma-2-9B} and \texttt{Mistral-3-7B} respectively achieve the highest faithfulness scores on AGNews and Ledgar, going against the idea of a monotonic relationship between model scale and explanation faithfulness. Figures~\ref{fig:concept_frequency_business_unfaithful_2b_agnews} and~\ref{fig:concept_frequency_world_unfaithful_2b_agnews} in Appendix~\ref{sec::appendix_2_hop_detailed_results} present examples of concepts frequently associated with unfaithful self-NLEs, revealing confusion where the model invokes concepts related to "business" when predicting the "world" class, and vice versa.

Comparing findings across classification and 2-hop reasoning reveals that accurate predictions tend to produce more faithful explanations, whereas scaling effects vary between tasks and models, as shown in prior work~\citep{atanasova_faithfulness_2023, parcalabescu_faithfulness, walk_the_talk_faithfulness}.

\section{Linearly Detecting and Improving Self-NLE Faithfulness}
\label{sec:linear_detection}
Recent work has demonstrated that various safety behaviors naturally associated with faithfulness are encoded as linear directions in LLM representation spaces~\citep{MI_survey}, such as hallucination~\citep{steering_mean_embedding} or deceptiveness~\citep{goldowskydetecting}. We demonstrate that \method faithfulness also exhibits linear structure for (1) accurate detection in representation space and (2) manipulation to improve faithfulness through activation steering.

\subsection{Linear Detection of Faithfulness}


\begin{figure}[t]{\centering}
\begin{center}
\includegraphics[width=\linewidth]{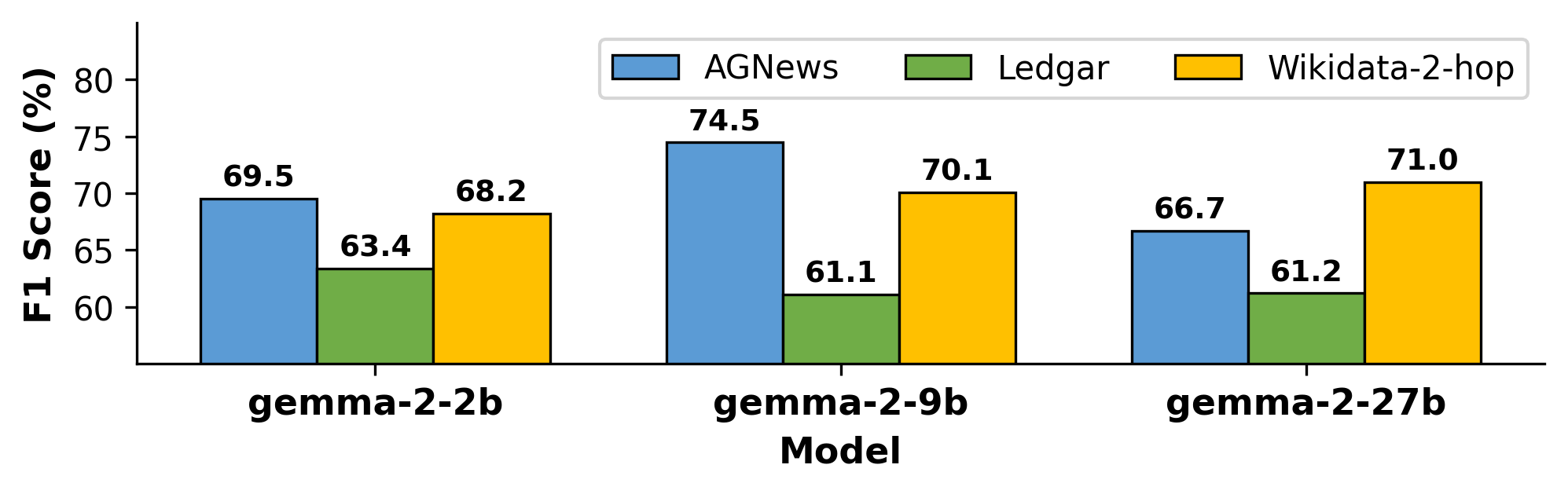}
    \caption{Faithfulness linear probe performance per model/ dataset.}
    \label{fig:linear_faithfulness}
\end{center}
\end{figure}

\begin{table*}[t]
\centering
\caption{Proportion of initially unfaithful self-NLE made faithful through steering interventions.}
\small
\begin{tabular}{lccccccccc}
\toprule
\textbf{Model} & \multicolumn{3}{c}{\textbf{Faithfulness} {\tiny $\lambda = 1$}} & \multicolumn{3}{c}{\textbf{Hallucination} {\tiny $\lambda = -1$}} & \multicolumn{3}{c}{\textbf{Deceptiveness} {\tiny $\lambda = -1$}} \\
\cmidrule(lr){2-4} \cmidrule(lr){5-7} \cmidrule(lr){8-10}
& \textbf{accurate} & \textbf{inaccurate} & \textbf{overall} & \textbf{accurate} & \textbf{inaccurate} & \textbf{overall} & \textbf{accurate} & \textbf{inaccurate} & \textbf{overall} \\
\midrule
\texttt{gemma-2-2b}  & 11.3\% & 9.3\% & 11.1\% & 10.0\% & 10.1\% & 10.0\% & 4.1\% & 5.7\% & 4.2\% \\
\texttt{gemma-2-9b}  & 8.3\% & 10.8\% & 8.7\% & 8.0\% & 11.2\% & 8.5\% & 4.9\% & 2.5\% & 4.5\% \\
\texttt{gemma-2-27b} & 10.9\% & 11.3\% & 11.0\% & 9.2\% & 9.1\% & 9.2\% & 4.1\% & 3.2\% & 3.9\% \\
\bottomrule
\end{tabular}
\label{tab:steering_results}
\end{table*}

\paragraph{Faithfulness Linear Representation Computation.} We construct datasets of faithful and unfaithful self-NLE pairs using \method\ scores. Unlike 2-hop reasoning tasks, classification self-NLE faithfulness yields continuous scores between 0 and 1. To ensure clear class separation, we focus on polarized cases where $F(x,e) \in \left\{0,1 \right\}$ for clear separation. We construct input sequences as $x_{nle} = [x, f(x), e(x)]$ and extract hidden states from final token positions across all layers. We train \texttt{diff-mean} linear probes on these states, yielding layer-wise faithfulness representations $\overrightarrow{F_{\ell}}$. For classification tasks, we use class-averaged faithfulness vectors to isolate faithfulness-specific representations independent of class confounders. Based on these faithfulness vectors, we compute a majority vote across layer-wise linear probes beyond layer 5.  Figure~\ref{fig:linear_faithfulness} shows faithfulness can be reliably detected across all evaluated tasks and models, with F1 scores ranging from 61.1\% to 74.5\%. This consistency suggests that \method\ captures a systematic aspect of model behavior that manifests similarly across different tasks. Details about class-averaged faithfulness vectors and layer-wise classification score are in Appendix~\ref{sec::linear_appendix}.


\begin{table}[t]
\centering
\caption{Linear vectors max. cosine similarity on \texttt{gemma-2-27b} with related layer. * indicates statistical significance, with $p<1\%$}
\small
\begin{tabular}{lcccc}
\toprule
& \textbf{AGNews} & \textbf{Ledgar} & \textbf{2 hop} & \textbf{Halluc.} \\
\midrule
\textbf{Ledgar} & \cellcolor{orange!60}0.49* {\tiny\textit{(22)}} & & & \\
\textbf{2 hop} & \cellcolor{yellow!40}0.28* {\tiny\textit{(43)}} & \cellcolor{blue!20}-0.37* {\tiny\textit{(18)}} & & \\
\textbf{Halluc.} & \cellcolor{blue!25}-0.32* {\tiny\textit{(22)}} & \cellcolor{blue!30}-0.38* {\tiny\textit{(19)}} & \cellcolor{blue!20}-0.28* {\tiny\textit{(22)}} & \\
\textbf{Decep.} & \cellcolor{blue!15}-0.19* {\tiny\textit{(28)}} & \cellcolor{yellow!30}0.21* {\tiny\textit{(18)}} & \cellcolor{blue!15}-0.20* {\tiny\textit{(40)}} & \cellcolor{orange!45}0.34* {\tiny\textit{(19)}} \\
\bottomrule
\end{tabular}
\label{tab:similarity_matrix_gemma_27_b}
\end{table}



\paragraph{Linear Representation Similarity Analysis.} To investigate the hypothesized relationships between faithfulness and related safety behaviors, we analyze the similarity between \method\ faithfulness vectors and established linear representations of hallucination and deceptiveness from ~\citet{steering_mean_embedding} and~\citet{goldowskydetecting}. Given that faithful self-NLE should reflect model reasoning while hallucination and deceptiveness involve misrepresentation of information, we expect their neural representations to exhibit inverse correlations. We focus on layers with F1 $>$ 60\% where faithfulness is properly linearly represented. Table~\ref{tab:similarity_matrix_gemma_27_b} shows maximum cosine similarity and corresponding layers for \texttt{gemma-2-27B} (additional results in Appendix~\ref{sec::linear_appendix}). Faithfulness vectors are in general negatively correlated with hallucination and deceptiveness ones (up to -0.38), aligning with expectations about the opposition between faithfulness and both deceptiveness and hallucination. Cross-task faithfulness vectors are overall positively correlated (up to 0.49), suggesting shared underlying representations of faithful across tasks. These correlations validate that \method\ captures LLMs behavioral aspects aligned with established safety behaviors. These results are corroborated by Table~\ref{tab:transfer_faithfulness} in Appendix~\ref{sec::linear_appendix}, showing cross-task transfer with \texttt{gemma-2-27b}: faithfulness vectors from 2-hop reasoning can detect unfaithful self-NLE in AGNews, and AGNews faithfulness vectors can detect unfaithful self-NLE in Ledgar.

\subsection{Faithfulness Enhancement Through Steering on 2-hop reasoning}
Having established the linear structure of \method\ faithfulness representations and their negative correlation with hallucination and deceptiveness vectors, we investigate whether linear steering can improve self-NLE faithfulness during inference. We implement activation steering by modifying hidden states during inference: $h^{\ell}_{k}\leftarrow h^{\ell}_{k} + \lambda \times \overrightarrow{SV_\ell}$, where $\overrightarrow{SV_\ell}$ is the steering vector and $\lambda$ controls intervention intensity. Our objective is here to demonstrate immediate practical value for improving self-NLE faithfulness during inference without model modification. More sophisticated steering methods~\citep{hedstromsteer, vogels2025distribution} would likely yield superior results. We evaluate three approaches: \textit{faithfulness amplification} ($\overrightarrow{SV_\ell} = \overrightarrow{F_\ell}$, $\lambda = 1$) and \textit{hallucination/deceptiveness inhibition} ($\overrightarrow{SV_\ell} = \overrightarrow{H_\ell}$ or $\overrightarrow{D_\ell}$, $\lambda = -1$), where $\overrightarrow{H_\ell}$ and $\overrightarrow{D_\ell}$ are hallucination and deceptiveness linear vectors obtained from~\cite{steering_mean_embedding} and~\cite{goldowskydetecting}. Faithfulness intervention targets only layers with F1 $>$ 60\%, ensuring modifications occur where faithfulness is reliably encoded.



Table~\ref{tab:steering_results} shows steering interventions successfully convert 8-11\% of unfaithful self-NLE into faithful explanations across all models. Direct faithfulness amplification slightly outperforms hallucination inhibition and substantially outperforms deceptiveness inhibition. Steering impact on faithfulness remains consistent regardless of prediction accuracy, demonstrating practical utility of \method\ for real-time faithfulness enhancement without model modification. Figure~\ref{fig:example} shows how a faithfulness linear probe identifies bridge objects as key tokens determining self-NLE faithfulness, comparing an unfaithful example with its steered faithful version. Additional analyses include comprehensive faithfulness status changes and other examples of converted self-NLE (see Appendix~\ref{sec::appendix_2_hop_detailed_results} and~\ref{sec::examples}).

\section{Conclusion}
\label{conclusion}
This work introduces \method, a framework that measures self-NLE faithfulness by measuring alignment between explanations and mechanistic analysis of model internals. \method\ aligns more closely than existing approaches with the common acceptance of faithfulness. Our findings suggest that faithfulness exhibits linear structure in LLM representation space for 2-hop reasoning and classification tasks, with emergent similarity with other established safety behaviors. Faithfulness linear structure enables both fast detection and enhancement through steering interventions, opening new avenues of research on faithfulness. In this paper, \method\ has been applied on circuit either previously defined or along the last token, based on probe accuracy. It would be valuable to run \method\ based on circuit discovery methods. Our analysis focuses on \textit{predict-then-explain} scenarios, extending to \textit{explain-then-predict} settings could reveal how explanation faithfulness relates to performance gains~\citep{bhan-etal-2024-self}. Extending \method\ to chain-of-thought reasoning could also provide valuable comparisons with existing CoT faithfulness studies~\citep{lanham_faithfulness,turpin_faithfuless}.

\newpage
\section*{Impact Statements}

This work aims to advance the trustworthiness of AI systems by developing methods to evaluate whether LLM self-explanations actually reflect their internal reasoning processes. Faithful self-explanations are critical for deploying LLMs in high-stakes domains such as healthcare or legal decision-making, where users must understand why a model reached a specific outcome. \method\ provides a principled tool for auditing explanation quality, potentially reducing over-reliance on plausible but misleading justifications. However any potential \method\ user must consider faithfulness score with caution, avoiding over-reliance on a single metric to take high-stake decisions. We especially higlight the importance of human validation during critical \method\ steps such as concept extraction. 

\bibliography{faithfulness}

@article{geirhos2020shortcut,
  title={Shortcut learning in deep neural networks},
  author={Geirhos, Robert and Jacobsen, J{\"o}rn-Henrik and Michaelis, Claudio and Zemel, Richard and Brendel, Wieland and Bethge, Matthias and Wichmann, Felix A},
  journal={Nature Machine Intelligence},
  volume={2},
  number={11},
  pages={665--673},
  year={2020},
  publisher={Nature Publishing Group UK London}
}

@book{grabisch2009aggregation,
	title={Aggregation functions},
	author={Grabisch, M. and Marichal, J.-L. and Mesiar, R. and Pap, E.},
	volume={127},
	year={2009},
	publisher={Cambridge University Press}
}

@article{paszke2019pytorch,
  title={Pytorch: An imperative style, high-performance deep learning library},
  author={Paszke, Adam and Gross, Sam and Massa, Francisco and Lerer, Adam and Bradbury, James and Chanan, Gregory and Killeen, Trevor and Lin, Zeming and Gimelshein, Natalia and Antiga, Luca and others},
  journal={Advances in neural information processing systems},
  OPTvolume={32},
  year={2019},
url    = {https://dl.acm.org/doi/10.5555/3454287.3455008}
}

@inproceedings{wolf2020transformers,
  title={Transformers: State-of-the-art natural language processing},
  author={Wolf, Thomas and Debut, Lysandre and Sanh, Victor and Chaumond, Julien and Delangue, Clement and Moi, Anthony and Cistac, Pierric and Rault, Tim and Louf, R{\'e}mi and Funtowicz, Morgan and others},
  booktitle={Proc. of the Conf. on Empirical Methods in Natural language Processing: system demonstrations, EMNLP},
  pages={38--45},
  year={2020},
  url = {https://aclanthology.org/2020.emnlp-demos.6/}
}

@article{pedregosa2011scikit,
  title={Scikit-learn: Machine learning in Python},
  author={Pedregosa, Fabian and Varoquaux, Ga{\"e}l and Gramfort, Alexandre and Michel, Vincent and Thirion, Bertrand and Grisel, Olivier and Blondel, Mathieu and Prettenhofer, Peter and Weiss, Ron and Dubourg, Vincent and others},
  journal={Journal of Machine Learning Research},
  volume={12},
  pages={2825--2830},
  year={2011},
  publisher={JMLR. org},
  url = {https://www.jmlr.org/papers/v12/pedregosa11a.html}
}

@inproceedings{biecek2024position,
  title={Position: explain to question not to justify},
  author={Biecek, Przemyslaw and Samek, Wojciech},
  booktitle={Proceedings of the 41st International Conference on Machine Learning},
  pages={3996--4006},
  year={2024}
}

@article{farah2023assessment,
  title={Assessment of Performance, Interpretability, and Explainability in Artificial Intelligence--Based Health Technologies: What Healthcare Stakeholders Need to Know},
  author={Farah, Line and Murris, Juliette M and Borget, Isabelle and Guilloux, Agathe and Martelli, Nicolas M and Katsahian, Sandrine IM},
  journal={Mayo Clinic Proceedings: Digital Health},
  volume={1},
  number={2},
  pages={120--138},
  year={2023},
  publisher={Elsevier}
}

@inproceedings{ribeiro2016should,
  title={" {{W}hy should {I} trust you?" Explaining the predictions of any classifier}},
  author={Ribeiro, Marco Tulio and Singh, Sameer and Guestrin, Carlos},
  booktitle={Proc. of the 22nd ACM SIGKDD {I}nt. {C}onference on {K}nowledge {D}iscovery and {D}ata {M}ining},
  year={2016}
}

@inproceedings{lundberg_unified_2017,
	title = {A {Unified} {Approach} to {Interpreting} {Model} {Predictions}},
	booktitle = {Advances in {Neural} {Information} {Processing} {Systems}},
	publisher = {NeurIPS},
	author = {Lundberg, Scott M and Lee, Su-In},
	year = {2017},
}

@inproceedings{sundararajan_axiomatic_2017,
	OPTaddress = {Sydney, NSW, Australia},
	series = {{ICML}'17},
	title = {Axiomatic attribution for deep networks},
	urldate = {2023-05-03},
	booktitle = {Proc. of the 34th {Int.} {Conf.} on {Machine} {Learning}, ICML},
    volume = 70,
    url = {https://proceedings.mlr.press/v70/sundararajan17a/sundararajan17a.pdf},
	publisher = {JMLR.org},
	author = {Sundararajan, Mukund and Taly, Ankur and Yan, Qiqi},
	month = aug,
	year = {2017},
	pages = {3319--3328}
}

@inproceedings{bhan2025towards,
    title = "Towards Achieving Concept Completeness for Textual Concept Bottleneck Models",
    author = {Bhan, Milan and Choho, Yann and Moreau, Pierre and Vittaut, Jean-Noel and Chesneau, Nicolas and Lesot, Marie-Jeanne},
    booktitle = "Findings of the 2025 Conference on Empirical Methods in Natural Language Processing",
    month = nov,
    year = "2025",
    publisher = "Association for Computational Linguistics",
}

@article{belrose2023leace,
  title={Leace: Perfect linear concept erasure in closed form},
  author={Belrose, Nora and Schneider-Joseph, David and Ravfogel, Shauli and Cotterell, Ryan and Raff, Edward and Biderman, Stella},
  journal={Advances in Neural Information Processing Systems},
  volume={36},
  pages={66044--66063},
  year={2023}
}

@inproceedings{tcav,
  title={Interpretability beyond feature attribution: Quantitative testing with concept activation vectors (tcav)},
  author={Kim, Been and Wattenberg, Martin and Gilmer, Justin and Cai, Carrie and Wexler, James and Viegas, Fernanda and others},
  booktitle={International conference on machine learning},
  pages={2668--2677},
  year={2018},
  organization={PMLR}
}

@article{elhage_linear_representation,
   title={Toy Models of Superposition},
   author={Elhage, Nelson and Hume, Tristan and Olsson, Catherine and Schiefer, Nicholas and Henighan, Tom and Kravec, Shauna and Hatfield-Dodds, Zac and Lasenby, Robert and Drain, Dawn and Chen, Carol and Grosse, Roger and McCandlish, Sam and Kaplan, Jared and Amodei, Dario and Wattenberg, Martin and Olah, Christopher},
   year={2022},
   journal={Transformer Circuits Thread},
   note={https://transformer-circuits.pub/2022/toy\_model/index.html}
}

@inproceedings{linear_rep_hypothesis,
  title={The Linear Representation Hypothesis and the Geometry of Large Language Models},
  author={Park, Kiho and Choe, Yo Joong and Veitch, Victor},
  booktitle={Forty-first International Conference on Machine Learning},
year={2024}
}

@article{ruiz2024theoretical,
  title={A theoretical design of concept sets: improving the predictability of concept bottleneck models},
  author={Ruiz Luyten, Max and van der Schaar, Mihaela},
  journal={Advances in Neural Information Processing Systems},
  volume={37},
  pages={100160--100195},
  year={2024}
}

@inproceedings{what_is_a_concept,
  title={What is a concept?},
  author={Goguen, Joseph},
  booktitle={Proceedings of the 13th international conference on Conceptual Structures: common Semantics for Sharing Knowledge},
  pages={52--77},
  year={2005}
}

@article{elhage2021mathematical,
   title={A Mathematical Framework for Transformer Circuits},
   author={Elhage, Nelson and Nanda, Neel and Olsson, Catherine and Henighan, Tom and Joseph, Nicholas and Mann, Ben and Askell, Amanda and Bai, Yuntao and Chen, Anna and Conerly, Tom and DasSarma, Nova and Drain, Dawn and Ganguli, Deep and Hatfield-Dodds, Zac and Hernandez, Danny and Jones, Andy and Kernion, Jackson and Lovitt, Liane and Ndousse, Kamal and Amodei, Dario and Brown, Tom and Clark, Jack and Kaplan, Jared and McCandlish, Sam and Olah, Chris},
   year={2021},
   journal={Transformer Circuits Thread},
   note={https://transformer-circuits.pub/2021/framework/index.html}
}

@misc{huang_can_2023,
	title = {Can {Large} {Language} {Models} {Explain} {Themselves}? {A} {Study} of {LLM}-{Generated} {Self}-{Explanations}},
	shorttitle = {Can {Large} {Language} {Models} {Explain} {Themselves}?},
	url = {http://arxiv.org/abs/2310.11207},
	urldate = {2023-12-04},
	publisher = {arXiv},
	author = {Huang, Shiyuan and Mamidanna, Siddarth and Jangam, Shreedhar and Zhou, Yilun and Gilpin, Leilani H.},
	month = oct,
	year = {2023},
	note = {arXiv:2310.11207 [cs]},
}

@inproceedings{self_consistency_madsen,
  title={Are self-explanations from large language models faithful},
  author={Madsen, Andreas and Chandar, Sarath and Reddy, Siva},
  booktitle = {Findings of the {Association} for {Computational} {Linguistics}: {ACL} 2024},
  year={2024}
}

@inproceedings{teach_me_explain,
  title={Teach Me to Explain: A Review of Datasets for Explainable Natural Language Processing},
  author={Wiegreffe, Sarah and Marasovic, Ana},
  booktitle={Thirty-fifth Conference on Neural Information Processing Systems Datasets and Benchmarks Track (Round 1)},
  year = {2021}
}

@inproceedings{yang2025large,
  title={Do Large Language Models Perform Latent Multi-Hop Reasoning without Exploiting Shortcuts?},
  author={Yang, Sohee and Kassner, Nora and Gribovskaya, Elena and Riedel, Sebastian and Geva, Mor},
  booktitle={Findings of the Association for Computational Linguistics: ACL 2025},
  pages={3971--3992},
  year={2025}
}

@article{abdin2024phi,
  title={Phi-4 Technical Report},
  author={Abdin, Marah and Aneja, Jyoti and Behl, Harkirat and Bubeck, S{\'e}bastien and Eldan, Ronen and Gunasekar, Suriya and Harrison, Michael and Hewett, Russell J and Javaheripi, Mojan and Kauffmann, Piero and others},
  journal={arXiv preprint arXiv:2412.08905},
  year={2024}
}

@inproceedings{bhan-etal-2024-self,
    title = "Self-{AMPLIFY}: Improving Small Language Models with Self Post Hoc Explanations",
    author = {Bhan, Milan  and
      Vittaut, Jean-No{\"e}l  and
      Chesneau, Nicolas  and
      Lesot, Marie-Jeanne},
    editor = "Al-Onaizan, Yaser  and
      Bansal, Mohit  and
      Chen, Yun-Nung",
    booktitle = "Proceedings of the 2024 Conference on Empirical Methods in Natural Language Processing",
    month = nov,
    year = "2024",
    address = "Miami, Florida, USA",
    publisher = "Association for Computational Linguistics",
    url = "https://aclanthology.org/2024.emnlp-main.615/",
    doi = "10.18653/v1/2024.emnlp-main.615",
    pages = "10974--10991"
}

@article{trivedi-etal-2022-musique,   
title = "{M}u{S}i{Q}ue: Multihop Questions via Single-hop Question Composition",
    author = "Trivedi, Harsh  and
      Balasubramanian, Niranjan  and
      Khot, Tushar  and
      Sabharwal, Ashish",
    editor = "Roark, Brian  and
      Nenkova, Ani",
    journal = "Transactions of the Association for Computational Linguistics",
    volume = "10",
    year = "2022",
    address = "Cambridge, MA",
    publisher = "MIT Press",
    url = "https://aclanthology.org/2022.tacl-1.31/",
    doi = "10.1162/tacl_a_00475",
    pages = "539--554"
}

@article{mavi2024multi,
  title={Multi-hop Question Answering},
  author={Mavi, Vaibhav and Jangra, Anubhav and Jatowt, Adam},
  journal={Foundations and Trends{\textregistered} in Information Retrieval},
  volume={17},
  number={5},
  pages={457--586},
  year={2024},
  publisher={Now Publishers Boston—Delft}
}

@article{team2024gemma,
  title={Gemma: Open models based on gemini research and technology},
  author={Gemma Team, Gemma},
  journal={arXiv preprint arXiv:2403.08295},
  year={2024}
}

@article{jiang2023mistral,
  title={Mistral 7B},
  author={Jiang, Albert Q and Sablayrolles, Alexandre and Mensch, Arthur and Bamford, Chris and Chaplot, Devendra Singh and Casas, Diego de las and Bressand, Florian and Lengyel, Gianna and Lample, Guillaume and Saulnier, Lucile and others},
  journal={arXiv preprint arXiv:2310.06825},
  year={2023},
url= {https://arxiv.org/abs/2310.06825}
}

@article{AG_news,
  title={Ag’s corpus of news articles},
  author={Gulli, Antonio},
  journal={Dipartimento di Informatica, University of Pisa, Nov},
  year={2005}
}

@inproceedings{tuggener2020ledgar,
  title={LEDGAR: a large-scale multi-label corpus for text classification of legal provisions in contracts},
  author={Tuggener, Don and Von D{\"a}niken, Pius and Peetz, Thomas and Cieliebak, Mark},
  booktitle={Proceedings of the twelfth language resources and evaluation conference},
  pages={1235--1241},
  year={2020}
}

@article{gu2024survey,
  title={A survey on llm-as-a-judge},
  author={Gu, Jiawei and Jiang, Xuhui and Shi, Zhichao and Tan, Hexiang and Zhai, Xuehao and Xu, Chengjin and Li, Wei and Shen, Yinghan and Ma, Shengjie and Liu, Honghao and others},
  journal={arXiv preprint arXiv:2411.15594},
  year={2024}
}

@inproceedings{hopping_too_late_multi_hop,
  title={Hopping Too Late: Exploring the Limitations of Large Language Models on Multi-Hop Queries},
  author={Biran, Eden and Gottesman, Daniela and Yang, Sohee and Geva, Mor and Globerson, Amir},
  booktitle={Proceedings of the 2024 Conference on Empirical Methods in Natural Language Processing},
  pages={14113--14130},
  year={2024}
}

@inproceedings{zaman-srivastava-2025-causal,
    title = "A Causal Lens for Evaluating Faithfulness Metrics",
    author = "Zaman, Kerem  and
      Srivastava, Shashank",
    editor = "Christodoulopoulos, Christos  and
      Chakraborty, Tanmoy  and
      Rose, Carolyn  and
      Peng, Violet",
    booktitle = "Proceedings of the 2025 Conference on Empirical Methods in Natural Language Processing",
    month = nov,
    year = "2025",
    address = "Suzhou, China",
    publisher = "Association for Computational Linguistics",
    url = "https://aclanthology.org/2025.emnlp-main.1496/",
    doi = "10.18653/v1/2025.emnlp-main.1496",
    pages = "29425--29449",
    ISBN = "979-8-89176-332-6"
}

@inproceedings{wiegreffe_faithfulness,
  title={Measuring Association Between Labels and Free-Text Rationales},
  author={Wiegreffe, Sarah and Marasovi{\'c}, Ana and Smith, Noah A},
  booktitle={Proceedings of the 2021 Conference on Empirical Methods in Natural Language Processing},
  pages={10266--10284},
  year={2021}
}

@inproceedings{atanasova_faithfulness_2023,
	address = {Toronto, Canada},
	title = {Faithfulness {Tests} for {Natural} {Language} {Explanations}},
	url = {https://aclanthology.org/2023.acl-short.25},
	doi = {10.18653/v1/2023.acl-short.25},
	urldate = {2024-01-31},
	booktitle = {Proc. of the 61st {Annual} {Meeting} of the {Association} for {Computational} {Linguistics} ({Volume} 2: {Short} {Papers})},
	publisher = {Association for Computational Linguistics},
	author = {Atanasova, Pepa and Camburu, Oana-Maria and Lioma, Christina and Lukasiewicz, Thomas and Simonsen, Jakob Grue and Augenstein, Isabelle},
	editor = {Rogers, Anna and Boyd-Graber, Jordan and Okazaki, Naoaki},
	month = jul,
	year = {2023},
	pages = {283--294},
}

@inproceedings{jacovi2020towards,
  title={Towards Faithfully Interpretable NLP Systems: How Should We Define and Evaluate Faithfulness?},
  author={Jacovi, Alon and Goldberg, Yoav},
  booktitle={Proceedings of the 58th Annual Meeting of the Association for Computational Linguistics},
  year={2020},
  organization={Association for Computational Linguistics}
}

@article{turpin_faithfuless,
  title={Language models don't always say what they think: Unfaithful explanations in chain-of-thought prompting},
  author={Turpin, Miles and Michael, Julian and Perez, Ethan and Bowman, Samuel},
  journal={Advances in Neural Information Processing Systems},
  volume={36},
  pages={74952--74965},
  year={2023}
}

@article{lanham_faithfulness,
  title={Measuring faithfulness in chain-of-thought reasoning},
  author={Lanham, Tamera and Chen, Anna and Radhakrishnan, Ansh and Steiner, Benoit and Denison, Carson and Hernandez, Danny and Li, Dustin and Durmus, Esin and Hubinger, Evan and Kernion, Jackson and others},
  journal={arXiv preprint arXiv:2307.13702},
  year={2023}
}

@article{XAI_nlp_survey,
  title={Towards faithful model explanation in nlp: A survey},
  author={Lyu, Qing and Apidianaki, Marianna and Callison-Burch, Chris},
  journal={Computational Linguistics},
  pages={1--67},
  year={2024},
  publisher={MIT Press 255 Main Street, 9th Floor, Cambridge, Massachusetts 02142, USA~…}
}

@article{trust_faithfulness,
  title={Is ignorance bliss? the role of post hoc explanation faithfulness and alignment in model trust in laypeople and domain experts},
  author={Han, Tessa and Ektefaie, Yasha and Farhat, Maha and Zitnik, Marinka and Lakkaraju, Himabindu},
  journal={arXiv preprint arXiv:2312.05690},
  year={2023}
}

@article{faithfulness_plausibility,
  title={Faithfulness vs. plausibility: On the (un) reliability of explanations from large language models},
  author={Agarwal, Chirag and Tanneru, Sree Harsha and Lakkaraju, Himabindu},
  journal={arXiv preprint arXiv:2402.04614},
  year={2024}
}

@inproceedings{luo2022evaluating,
  title={Evaluating Explanation Correctness in Legal Decision Making.},
  author={Luo, Chu Fei and Bhambhoria, Rohan and Dahan, Samuel and Zhu, Xiaodan},
  booktitle={Canadian AI},
  year={2022}
}

@inproceedings{kayser2024fool,
  title={Fool Me Once? Contrasting Textual and Visual Explanations in a Clinical Decision-Support Setting},
  author={Kayser, Maxime and Menzat, Bayar and Emde, Cornelius and Bercean, Bogdan and Novak, Alex and Morgado, Abdal{\'a} and Papiez, Bartlomiej and Gaube, Susanne and Lukasiewicz, Thomas and Camburu, Oana-Maria},
  booktitle={Proceedings of the 2024 Conference on Empirical Methods in Natural Language Processing},
  pages={18891--18919},
  year={2024}
}

@inproceedings{siegel_faithfulness,
    title = "The Probabilities Also Matter: A More Faithful Metric for Faithfulness of Free-Text Explanations in Large Language Models",
    author = "Siegel, Noah  and
      Camburu, Oana-Maria  and
      Heess, Nicolas  and
      Perez-Ortiz, Maria",
    editor = "Ku, Lun-Wei  and
      Martins, Andre  and
      Srikumar, Vivek",
    booktitle = "Proceedings of the 62nd Annual Meeting of the Association for Computational Linguistics (Volume 2: Short Papers)",
    month = aug,
    year = "2024",
    address = "Bangkok, Thailand",
    publisher = "Association for Computational Linguistics",
    url = "https://aclanthology.org/2024.acl-short.49/",
    doi = "10.18653/v1/2024.acl-short.49",
    pages = "530--546"
}

@article{siegel_faithfulness_V2,
  title={Faithfulness of LLM Self-Explanations for Commonsense Tasks: Larger Is Better, and Instruction-Tuning Allows Trade-Offs but Not Pareto Dominance},
  author={Siegel, Noah Y and Heess, Nicolas and Perez-Ortiz, Maria and Camburu, Oana-Maria},
  journal={arXiv preprint arXiv:2503.13445},
  year={2025}
}

@inproceedings{walk_the_talk_faithfulness,
  title={Walk the Talk? Measuring the Faithfulness of Large Language Model Explanations},
  author={Matton, Katie and Ness, Robert and Kiciman, Emre},
  booktitle={The Fourteenth International Conference on Learning Representations},
year={2025}}

@inproceedings{parcalabescu_faithfulness,
  title={On Measuring Faithfulness or Self-consistency of Natural Language Explanations},
  author={Parcalabescu, Letitia and Frank, Anette},
  booktitle={Proceedings of the 62nd Annual Meeting of the Association for Computational Linguistics (Volume 1: Long Papers)},
  pages={6048--6089},
  year={2024}
}

@inproceedings{yeo_faithfulness,
  title={Towards faithful natural language explanations: A study using activation patching in large language models},
  author={Yeo, Wei Jie and Satapathy, Ranjan and Cambria, Erik},
  booktitle={Proceedings of the 2025 Conference on Empirical Methods in Natural Language Processing},
  pages={10436--10458},
  year={2025}
}

@inproceedings{tutek_faithfulness,
  title={Measuring chain of thought faithfulness by unlearning reasoning steps},
  author={Tutek, Martin and Chaleshtori, Fateme Hashemi and Marasovi{\'c}, Ana and Belinkov, Yonatan},
  booktitle={Proceedings of the 2025 Conference on Empirical Methods in Natural Language Processing},
  pages={9946--9971},
  year={2025}
}

@article{liu2025rethinking,
  title={Rethinking machine unlearning for large language models},
  author={Liu, Sijia and Yao, Yuanshun and Jia, Jinghan and Casper, Stephen and Baracaldo, Nathalie and Hase, Peter and Yao, Yuguang and Liu, Chris Yuhao and Xu, Xiaojun and Li, Hang and others},
  journal={Nature Machine Intelligence},
  pages={1--14},
  year={2025},
  publisher={Nature Publishing Group UK London}
}

@article{geiger2025causal,
  title={Causal abstraction: A theoretical foundation for mechanistic interpretability},
  author={Geiger, Atticus and Ibeling, Duligur and Zur, Amir and Chaudhary, Maheep and Chauhan, Sonakshi and Huang, Jing and Arora, Aryaman and Wu, Zhengxuan and Goodman, Noah and Potts, Christopher and others},
  journal={Journal of Machine Learning Research},
  volume={26},
  number={83},
  pages={1--64},
  year={2025}
}

@inproceedings{rauker2023toward,
  title={Toward transparent ai: A survey on interpreting the inner structures of deep neural networks},
  author={R{\"a}uker, Tilman and Ho, Anson and Casper, Stephen and Hadfield-Menell, Dylan},
  booktitle={2023 ieee conference on secure and trustworthy machine learning (satml)},
  pages={464--483},
  year={2023},
  organization={IEEE}
}

@article{rai2024practical,
  title={A practical review of mechanistic interpretability for transformer-based language models},
  author={Rai, Daking and Zhou, Yilun and Feng, Shi and Saparov, Abulhair and Yao, Ziyu},
  journal={arXiv preprint arXiv:2407.02646},
  year={2024}
}

@inproceedings{wanginterpretability,
  title={Interpretability in the Wild: a Circuit for Indirect Object Identification in GPT-2 Small},
  author={Wang, Kevin Ro and Variengien, Alexandre and Conmy, Arthur and Shlegeris, Buck and Steinhardt, Jacob},
  booktitle={The Eleventh International Conference on Learning Representations},
  year={2023}
}

@article{meng2022locating,
  title={Locating and editing factual associations in gpt},
  author={Meng, Kevin and Bau, David and Andonian, Alex and Belinkov, Yonatan},
  journal={Advances in neural information processing systems},
  volume={35},
  pages={17359--17372},
  year={2022}
}

@inproceedings{ferrando2024information,
  title={Information Flow Routes: Automatically Interpreting Language Models at Scale},
  author={Ferrando, Javier and Voita, Elena},
  booktitle={Proceedings of the 2024 Conference on Empirical Methods in Natural Language Processing},
  pages={17432--17445},
  year={2024}
}

@inproceedings{dunefskytranscoders,
  title={Transcoders find interpretable LLM feature circuits},
  author={Dunefsky, Jacob and Chlenski, Philippe and Nanda, Neel},
  booktitle={The Thirty-eighth Annual Conference on Neural Information Processing Systems},
  year={2024}
  }

@inproceedings{selfie,
  title={SelfIE: Self-Interpretation of Large Language Model Embeddings},
  author={Chen, Haozhe and Vondrick, Carl and Mao, Chengzhi},
  booktitle={Forty-first International Conference on Machine Learning},
  year={2024}
}

@inproceedings{patchscopes,
  title={Patchscopes: A Unifying Framework for Inspecting Hidden Representations of Language Models},
  author={Ghandeharioun, Asma and Caciularu, Avi and Pearce, Adam and Dixon, Lucas and Geva, Mor},
  booktitle={Forty-first International Conference on Machine Learning},
  year={2024}
}

@article{belinkov2022probing,
  title={Probing Classifiers: Promises, Shortcomings, and Advances},
  author={Belinkov, Yonatan},
  journal={Computational Linguistics},
  volume={48},
  number={1},
  pages={207--219},
  year={2022}
}

@article{zou2023representation,
  title={Representation engineering: A top-down approach to ai transparency},
  author={Zou, Andy and Phan, Long and Chen, Sarah and Campbell, James and Guo, Phillip and Ren, Richard and Pan, Alexander and Yin, Xuwang and Mazeika, Mantas and Dombrowski, Ann-Kathrin and others},
  journal={arXiv preprint arXiv:2310.01405},
  year={2023}
}

@inproceedings{wuaxbench,
  title={AxBench: Steering LLMs? Even Simple Baselines Outperform Sparse Autoencoders},
  author={Wu, Zhengxuan and Arora, Aryaman and Geiger, Atticus and Wang, Zheng and Huang, Jing and Jurafsky, Dan and Manning, Christopher D and Potts, Christopher},
    year = {2025},
  booktitle={Forty-second International Conference on Machine Learning}
}

@article{acdc,
  title={Towards automated circuit discovery for mechanistic interpretability},
  author={Conmy, Arthur and Mavor-Parker, Augustine and Lynch, Aengus and Heimersheim, Stefan and Garriga-Alonso, Adri{\`a}},
  journal={Advances in Neural Information Processing Systems},
  volume={36},
  pages={16318--16352},
  year={2023}
}

@article{vogels2025distribution,
  title={In-Distribution Steering: Balancing Control and Coherence in Language Model Generation},
  author={Vogels, Arthur and Wong, Benjamin and Choho, Yann and Blangero, Annabelle and Bhan, Milan},
  journal={arXiv preprint arXiv:2510.13285},
  year={2025}
}

@article{MI_survey,
  title={Mechanistic Interpretability for AI Safety-A Review},
  author={Bereska, Leonard and Gavves, Stratis},
  journal={Transactions on Machine Learning Research},
  year={2024}
}

@inproceedings{steering_mean_embedding,
    title = "Steering Llama 2 via Contrastive Activation Addition",
    author = "Rimsky, Nina  and
      Gabrieli, Nick  and
      Schulz, Julian  and
      Tong, Meg  and
      Hubinger, Evan  and
      Turner, Alexander",
    editor = "Ku, Lun-Wei  and
      Martins, Andre  and
      Srikumar, Vivek",
    booktitle = "Proceedings of the 62nd Annual Meeting of the Association for Computational Linguistics (Volume 1: Long Papers)",
    month = aug,
    year = "2024",
    address = "Bangkok, Thailand",
    publisher = "Association for Computational Linguistics",
    url = "https://aclanthology.org/2024.acl-long.828/",
    doi = "10.18653/v1/2024.acl-long.828",
    pages = "15504--15522"
}

@inproceedings{sahoo2024comprehensive,
  title={A Comprehensive Survey of Hallucination in Large Language, Image, Video and Audio Foundation Models},
  author={Sahoo, Pranab and Meharia, Prabhash and Ghosh, Akash and Saha, Sriparna and Jain, Vinija and Chadha, Aman},
  booktitle={Findings of the Association for Computational Linguistics: EMNLP 2024},
  pages={11709--11724},
  year={2024}
}

@inproceedings{goldowskydetecting,
  title={Detecting Strategic Deception with Linear Probes},
  author={Goldowsky-Dill, Nicholas and Chughtai, Bilal and Heimersheim, Stefan and Hobbhahn, Marius},
  booktitle={Forty-second International Conference on Machine Learning},
year={2025}
}

@inproceedings{hedstromsteer,
  title={To Steer or Not to Steer? Mechanistic Error Reduction with Abstention for Language Models},
  author={Hedstr{\"o}m, Anna and Amoukou, Salim I and Bewley, Tom and Mishra, Saumitra and Veloso, Manuela},
year={2025},
  booktitle={Forty-second International Conference on Machine Learning}
}

@book{pearl2009causality,
  title={Causality},
  author={Pearl, Judea},
  year={2009},
  publisher={Cambridge university press}
}
\bibliographystyle{icml2026}

\newpage
\onecolumn
\appendix
\section*{Appendix}


\printappendixcontents

\renewcommand{\thesubsection}{\Alph{subsection}}

\subsection{Scientific Libraries}
\label{sec:appendix_scientific_library}
We used several open-source libraries in this work: pytorch~\citep{paszke2019pytorch}, HuggingFace transformers~\citep{wolf2020transformers} and sklearn~\cite{pedregosa2011scikit}.

\subsection{LLM Implementation Details}
\label{sec:appendix_llm_implementation_details}
\paragraph{Backbone and Special Tokens.} The library used to import the pretrained autoregressive language models is Hugging-Face. In particular, the backbones version of Gemma-2-2B, Gemma-2-9B and Gemma-2-27B are \texttt{gemma-2-2B-it}, \texttt{gemma-2-9B-it}, \texttt{gemma-2-27B-it} respectively. The models were imported with the \texttt{Bfloat16} computational format. The following special tokens used were used for instruction prompting:
\begin{itemize}
        \item \texttt{user\_token= '<start\_of\_turn>user'}
        \item \texttt{assistant\_token= '<start\_of\_turn>model'}
        \item \texttt{stop\_token= '<eos>'}
    \end{itemize}
    
\paragraph{Text Generation.} Text generation was performed using the native functions of the Hugging Face library: \texttt{generate}. The \texttt{generate} function was used with the following parameters:
\begin{itemize}
    \item \texttt{do\_sample = True}
    \item \texttt{num\_beams = 2}
    \item \texttt{no\_repeat\_ngram\_size = 2}
    \item \texttt{repetition\_penalty =1.2}
    \item \texttt{early\_stopping = True}
    \item \texttt{temperature = 0.05}
\end{itemize}

\paragraph{Self-NLE Generation.} The prompt used to get self-NLE was as follows: 
\begin{itemize}
    \item \textbf{$<$user$>$}
    \item \textit{Question}
    \item \textbf{$<$Assistant$>$}
    \item \textit{Answer}
    \item \textbf{$<$user$>$}
    \item \textit{"\texttt{Give me a simple explanation of your answer.}"} 
    \item \textbf{$<$Assistant$>$}
\end{itemize}

\subsection{Datasets}
\label{sec:datasets}
Here we provide detailed information about the analyzed datasets. For 2-hop reasoning, we run \method\ on the Wikidata-2-hop dataset~\citep{hopping_too_late_multi_hop} and Socrates~\citep{yang2025large}. We first compute task accuracy on the whole dataset, and then sample 1500 accurate and inaccurate prediction for Wikidata-2-hop. To sample these 1500 instances, we filter out 2-hop reasoning questions where relations are subjective (e.g. "\textit{the most notable work of}") or equivocal, potentially leading to numerous possible answers (e.g. "\textit{the work that features}" ). We also sample by setting a maximum number of occurrences for generated answers (15), to foster diversity in the input questions. This filter enables to avoid having too many questions where the answer is a country (e.g. USA). We end up with a dataset made of 3000 samples with 1500 accurate and 1500 inaccurate predictions.

For classification, we run \method\ on AGNews~\citep{AG_news}, a newspaper article classification and Ledgar~\citep{tuggener2020ledgar}, a more challenging critical domain legal document classification dataset. We retrieve the enriched versions from~\cite{bhan2025towards} with labeled concepts for CAV computation. For AGNews, the classes to predict are 'world', "sport", "business" and "science \& technology". For Ledgar, the classes to predict are "Amendments", "Survival", "Terminations" and "Terms". Each dataset is made of 4000 samples.

\subsection{\method\ Implementation Details}
\label{sec::appendix_implementation_detail}
\subsubsection{Concept Extraction}
\label{sec::method_CE}

Here we provide the prompts used to give instructions to \texttt{Qwen-3-32b} to extract relevant concepts (\method\ step 1). For 2-hop reasoning, concept extraction consists in retrieving the bridge object from the self-NLE. Since the input text as the following structures : $x= (o_{1},r_1,\tiny{\blacktriangle{}},r_2,\bullet)$, we directly prompt the model to resolve $(o_{1},r_1,\tiny{\blacktriangle{}}$) by grounding its response on the self-NLE only. We structure our prompt following an in-context learning template with two examples differing from the dataset of interest:

\begin{tcolorbox}[title = Concept Extraction Prompt for 2-hop Reasoning, width=\linewidth]
    \textbf{user}\\
    \texttt{preprompt + preprompt\_example\_1} \\
    \textbf{assistant} \\
    \textit{Emmanuel Macron} \\
    \textbf{user} \\
    \texttt{preprompt + preprompt\_example\_2} \\
    \textbf{assistant} \\
    \textit{Ingmar Bergman} \\
    \textbf{user} \\
    \texttt{preprompt + } $(o_{1},r_1,\tiny{\blacktriangle{}}$)
\end{tcolorbox}

with \texttt{preprompt} = "\textit{Answer briefly and only according to the provided text. If there is no clear answer, say **no bridge object**}", \texttt{preprompt\_example\_1} = "\textit{Emmanuel Macron is the president of Italy, and the capital city of Italy is Rome.** the president of Italy is}" and \texttt{preprompt\_example\_2} = "\textit{The movie Persona is a movie happening in the Faro island and has been directed by Ingmar Bergman, who is from Sweden.**: 'The director of Persona is}"

For classification, we assess having access to a set of relevant concepts $\mathcal{C}$ and labels related to the task of interest. Given a certain concept $c$, we only prompt the model to assess if the concept is present in the self-NLE if the concept was initially present in the input text, making the concept extraction process computationally less expensive. We structure our prompt following an in-context learning template with three examples differing from the dataset of interest:

\begin{tcolorbox}[title = Concept Extraction Prompt for Classification, width=\linewidth]
    \textbf{user}\\
    \texttt{preprompt + preprompt\_example\_1} \\
    \textbf{assistant} \\
    \textit{yes} \\
    \textbf{user} \\
    \texttt{preprompt + preprompt\_example\_2} \\
    \textbf{assistant} \\
    \textit{yes} \\
    \textbf{user} \\
    \texttt{preprompt + preprompt\_example\_3} \\
    \textbf{assistant} \\
    \textit{no} \\
    \textbf{user} \\
    \texttt{preprompt + context\_extraction\_prompt}($ \hat{y}, e(x), c$)
\end{tcolorbox}

with \texttt{preprompt} = "\textit{Analyze whether a given concept has a meaningful impact on predicting a specific category from the provided text explanation. Instructions: (1) Answer with exactly "YES" if the concept is clearly mentioned in the given text and relevant to the category prediction, (2) Answer with exactly "NO" if the concept is neither mentioned nor relevant in the given text (3) Consider the logical connection between the concept and the category in the given text}". 

For AGNews: \begin{itemize}
    \item \texttt{preprompt\_example\_1} = "\textit{Text explanation: 'The article says that OECD countries became richer in the 20th century. This falls under the category of world'. Question: According to the previous text, does the concept "economic trends" have a meaningful impact on predicting the "world" category?}"
    \item \texttt{preprompt\_example\_2} = "\textit{Text explanation: 'The abstract underlines that the French soccer striker is a good player. It is relevant to sport'. Question: According to the previous text, does the concept "jargon specific to the sport" have a meaningful impact on predicting the "sport" category?}".
    \item \texttt{preprompt\_example\_3} = "\textit{Text explanation: 'The research paper discusses quantum computing algorithms and their complexity. This falls under the category of technology'. Question: According to the previous text, does the concept "cooking techniques" have a meaningful impact on predicting the "technology" category?}".
\end{itemize}   

For Ledgar: \begin{itemize}
    \item \texttt{preprompt\_example\_1} = "\textit{Text explanation: 'This clause defines what happens to your stock options if your employment ends *before* they are fully vested.  It's part of the core **terms** of your employment agreement that outlines how these shares work.' Question: According to the previous text, does the legal  concept "Minimum commitment periods" have a meaningful impact on predicting the "terms" category?}"
    \item \texttt{preprompt\_example\_2} = "\textit{Text explanation: 'This clause defines what happens to your stock options if your employment ends *before* they are fully vested.  It's part of the core **terms** of your employment agreement that outlines how these shares work.' Question: According to the previous text, does the legal  concept "Effect of termination" have a meaningful impact on predicting the "terms" category?}".
    \item \texttt{preprompt\_example\_3} = "\textit{Text explanation: 'The clause specifically talks about how changes can be made to the agreement:"terminated, amended, modified or supplemented". These are all words that mean changing the original terms of the contract. Since it focuses on how the contract itself can be altered, the relevant category is **Amendments**' Question: According to the previous text, does the legal  concept "Amendment procedures" have a meaningful impact on predicting the "Amendments" category?}". 
\end{itemize}  

Finally, given a prediction $\hat{y}$, and self NLE $e(x)$ and a concept $c$, \texttt{context\_extraction\_prompt}($ \hat{y}, e(x), c$) = "\textit{Text explanation: $e(x)$. Question: According to the previous text, does the concept $c$ have a meaningful impact on predicting the $\hat{y}$ category?}".

\subsubsection{Mechanistic Interpretation}
\label{sec::method_MI}

Here we provide implementation details about mechanistic interpretations of extracted concepts (\method\ step 2).

\paragraph{Circuit Granularity.} To generate an mechanistic interpretation of an LLM hidden state, the first step is to define the specific neural architectural component and granularity $\mathcal{G}$.  Given an index $k$ and a layer $\ell$, a granularity $\mathcal{G}$ refers to a specific point within the model's computational subgraph where neural activations can be analyzed and from which a local neural interpretation $i^{\ell}_{k}(x)$ is going to be generated. Transformer-based generative LLM can be viewed as a stack of decoder computational blocks~\citep{elhage_linear_representation}. The information flow through a single layer and from a layer to another can be described with the following equations: 
\begin{align}
h_k^\ell &= h_k^{\ell-1} + a_k^\ell + m_k^\ell \nonumber \\
a_k^\ell &= \text{MHA}^\ell \left( h_1^{\ell-1}, h_2^{\ell-1}, \ldots, h_k^{\ell-1} \right) \nonumber \\
m_k^\ell &= \text{MLP}^\ell(a_k^\ell + h_k^{\ell-1})
\label{eq:transformer}
\end{align}
where $h_k^\ell$ denotes the residual stream at index $k$ and layer $\ell$, $a_k^\ell$ 
the attention output of the multi-head attention operation $\text{MHA}^\ell$ and $m_k^\ell$ 
the output from the multi-layer perceptron $\text{MLP}^\ell$.

This characterization of the information flow through a Transformer block naturally highlights three possible granularities, each offering different perspectives on the model's information processing: the residual stream (\texttt{RS}) focusing on $h_k^\ell$, which serves as the main information pathway through the network; the multi-head attention (\texttt{MHA}) focusing on $a_k^\ell$, which captures token-to-token interactions; and the multi-layer perceptron (\texttt{MLP}) focusing on $m_k^\ell$ and performing non-linear transformations of the information~\citep{elhage2021mathematical,rai2024practical}. These computational blocks offer a relevant 
granularity to decode the internal activity of an LLM, since focusing on neurons alone can rarely be done due to their polysemanticity~\citep{elhage_linear_representation}. Thus, a hidden state granularity is defined such as $\mathcal{G} \in  \left\{\texttt{RS, MHA, MLP}\right\}$.

\paragraph{Interpreter.} For the 2-hop reasoning instantiation, the interpreter used to decode $f$ hidden states $\left\{ h_k^{\ell} \right\}$ is \texttt{Patchscopes}~\cite{patchscopes}. We use the prompt \textit{"What is the following? Answer briefly [X,X]}" to generate the interpretation where X is a token placeholder to be replaced in the latent space by the hidden state to be interpreted. We replace the placeholder tokens at layers 3 and 4 to get two interpretations per hidden state to decode. The layers to be interpreted by \texttt{Patchscopes} vary depending on the assessed model. We set the index token $k$ as the last one related to $o_{1}$ and focus on the late early layers as in~\cite{hopping_too_late_multi_hop}. This way, $\ell \in \left\{ {5,6,7,8,9,10,11}\right\}$ for \texttt{gemma-2-2B}, $\ell \in \left\{ {8,9,10,11,12,13,14}\right\}$ for \texttt{gemma-2-9B} and $\ell \in \left\{ {11,12,13,14,15,16,17}\right\}$ for \texttt{gemma-2-27B}.

For the classification instantiation and given a concept $c$, concept importance $\texttt{I}_{\Gamma}(c)$ is based on the linear representation of the concept called CAV and denoted $\overrightarrow{c}$. Concepts are selected based on the identifiability of each concept on $f$ representation space based on $\texttt{I}_{\Gamma}(c)$. Below are examples of selected concepts from Ledgar and AGNews for \texttt{gemma-2-27b} with F1 score $>60\%$:

\begin{table}[H]
\centering
\begin{tabular}{lcc}
\toprule
\textbf{Concept} & \textbf{Layer} & \textbf{F1 Score (\%)} \\
\midrule
Players & 36 & 0.758 \\
Political developments & 43 & 0.842 \\
Scores & 10 & 0.626 \\
Financial markets & 35 & 0.758 \\
Companies & 36 & 0.805 \\
Industry-specific terminology and jargon & 34 & 0.745 \\
Global issues & 45 & 0.812 \\
Sports events & 45 & 0.854 \\
Industry analysis & 38 & 0.610 \\
Economic trends & 36 & 0.798 \\
Industries & 36 & 0.776 \\
International events & 36 & 0.847 \\
Global politics & 36 & 0.809 \\
International relations & 36 & 0.776 \\
Foreign affairs & 43 & 0.795 \\
News about wars, conflicts & 43 & 0.694 \\
Athletic competitions & 45 & 0.855 \\
Teams & 45 & 0.679 \\
Game summaries & 34 & 0.780 \\
Jargon specific to the sport & 45 & 0.854 \\
Charts, graphs, and financial data & 38 & 0.658 \\
Advancements in computing & 41 & 0.607 \\
Technological trends & 43 & 0.769 \\
\bottomrule
\end{tabular}
\caption{AGNews concept max. F1 scores ($>60\%$) with related layer for \texttt{gemma-2-27b}.}
\label{tab:topic_performance_agnews}
\end{table}

\begin{table}[H]
\centering
\begin{tabular}{lcc}
\toprule
\textbf{Concept} & \textbf{Layer} & \textbf{F1 Score (\%)} \\
\midrule
Modification rights & 34 & 0.838 \\
Amendment procedures & 39 & 0.803 \\
Notice requirements & 20 & 0.743 \\
Approval mechanisms & 28 & 0.816 \\
Integration with original agreement & 23 & 0.729 \\
Format requirements & 26 & 0.768 \\
Severability of amendments & 37 & 0.812 \\
Retroactive application & 27 & 0.638 \\
Waiver limitations & 30 & 0.793 \\
Amendment thresholds & 42 & 0.753 \\
Amendment restrictions & 36 & 0.730 \\
Prior versions validity & 45 & 0.759 \\
Amendment documentation & 24 & 0.759 \\
Version control mechanisms & 37 & 0.738 \\
Material change provisions & 33 & 0.754 \\
Post-termination obligations & 28 & 0.798 \\
Duration of surviving terms & 28 & 0.885 \\
Identification of specific clauses & 11 & 0.785 \\
Indemnification continuation & 40 & 0.862 \\
Payment obligations survival & 21 & 0.776 \\
Representations/warranties survival & 21 & 0.767 \\
Remedies availability post-termination & 42 & 0.845 \\
Perpetual rights & 40 & 0.815 \\
Legal compliance requirements & 34 & 0.721 \\
Duration specifications & 42 & 0.885 \\
Commencement date & 36 & 0.638 \\
Expiration conditions & 37 & 0.910 \\
Renewal mechanisms & 44 & 0.654 \\
Term length & 26 & 0.798 \\
Condition precedents & 28 & 0.798 \\
Milestone-based periods & 36 & 0.739 \\
Initial term vs. renewal term distinctions & 44 & 0.789 \\
Evergreen provisions & 13 & 0.768 \\
Term modification triggers & 28 & 0.728 \\
Minimum commitment periods & 34 & 0.777 \\
Maximum term limitations & 36 & 0.785 \\
Regulatory term constraints & 41 & 0.769 \\
Term acceleration provisions & 30 & 0.738 \\
Rolling term provisions & 29 & 0.823 \\
Termination rights & 28 & 0.888 \\
Notice periods & 45 & 0.726 \\
Termination for convenience & 44 & 0.677 \\
Effect of termination & 28 & 0.912 \\
Wind-down procedures & 28 & 0.865 \\
Mutual termination provisions & 27 & 0.854 \\
Partial termination rights & 37 & 0.833 \\
Change of control provisions & 29 & 0.729 \\
Performance-based termination & 44 & 0.719 \\
Regulatory/legal change termination & 15 & 0.776 \\
Termination certification requirements & 27 & 0.774 \\
Post-termination restrictions & 13 & 0.747 \\
Transition obligations & 28 & 0.856 \\
\bottomrule
\end{tabular}
\caption{Ledgar concept max. F1 scores ($>60\%$) with related layer for \texttt{gemma-2-27b}.}
\label{tab:topic_performance_ledgar}
\end{table}

The concept importance $\texttt{I}_{\Gamma}(c)$ can be either computed through representation engineering and concept erasure or gradient-based approaches such as TCAV. We perform concept erasure for $\lambda \in [0,1]$ with a step size of 0.1. Table~\ref{tab:topic_performance_agnews} and~\ref{tab:topic_performance_ledgar} show the F1 scores of the properly detected concepts in \texttt{gemma-2-27b} representation space. We estimate the attribution of $c$ given circuit $\Gamma$ as the $\beta_1$ parameter of the following regression: $p_{\hat{y}}(x, \left\{ \text{do}(H^{\ell}_{k} =  h^{\ell}_{k}-\lambda \times \overrightarrow{c_\ell}) \right\}_{(k,\ell)\in \Gamma}) = \beta_0 + \beta_1 \times \lambda$. The t-test for $\beta_1$ significancy is expressed as follows: $t = \frac{\hat{\beta}_1}{SE(\hat{\beta}_1)} = \frac{\hat{\beta}_1}{\sqrt{\frac{MSE}{\sum_{i=1}^{n}(\lambda_i - \bar{\lambda})^2}}}$, where $\lambda_i$ is the $i$ realization of $\lambda$ and $\bar{\lambda}$ is the average $\lambda$ on the analyzed sample. 

The concept importance $\texttt{I}_{\Gamma}(c)$ can also be computed with gradients-based approaches such as \texttt{TCAV}. It can be formally expressed based on the previously computed CAV $\overrightarrow{c}$.  $\texttt{I}_{\Gamma}(c)$ is calculated by aggregating the local importance measures related to the hidden states along circuit $\Gamma$: 

\begin{equation}
    \texttt{I}_{\Gamma}(c) = \underset{(k,\ell)\in \Gamma}{\bigodot}  \langle \overrightarrow{c}, \nabla f^{\ell}_{\hat{y},k}(h^\ell_k)\rangle
\end{equation} where $f^{\ell}_{\hat{y},k}$ is the sub-function from $f$ taking $h^\ell_k$ as input and generating the output $p_{\hat{y}}$ and $\bigodot$ an aggregation operator chosen according to the expected desired level of strictness to measure faithfulness.  Among the many aggregation operators (see e.g. \citet{grabisch2009aggregation}), $\bigodot$ can be conjunctive (e.g. defined as the  $\min$ function), disjunctive (e.g. the $\max$ function) or the \texttt{average} measure.

\begin{figure}[H]{\centering}
\centering
\includegraphics[scale=0.5]{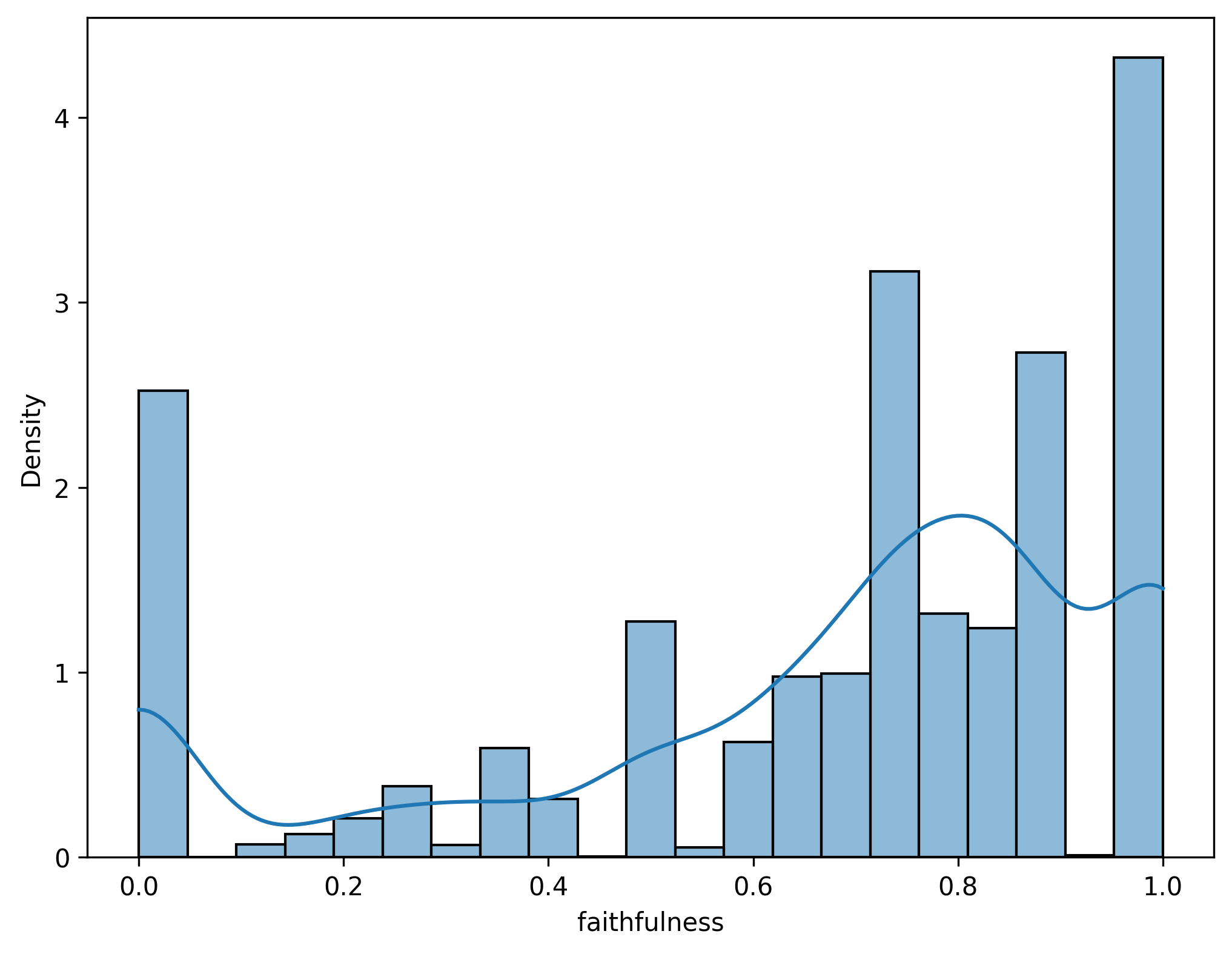}
\caption{\method\ faithfulness distribution for AGNews for \texttt{gemma-2-27b}.}  
\label{fig:faithfulness_distrib_agnews}
\end{figure}

\begin{figure}[H]{\centering}
\centering
\includegraphics[scale=0.5]{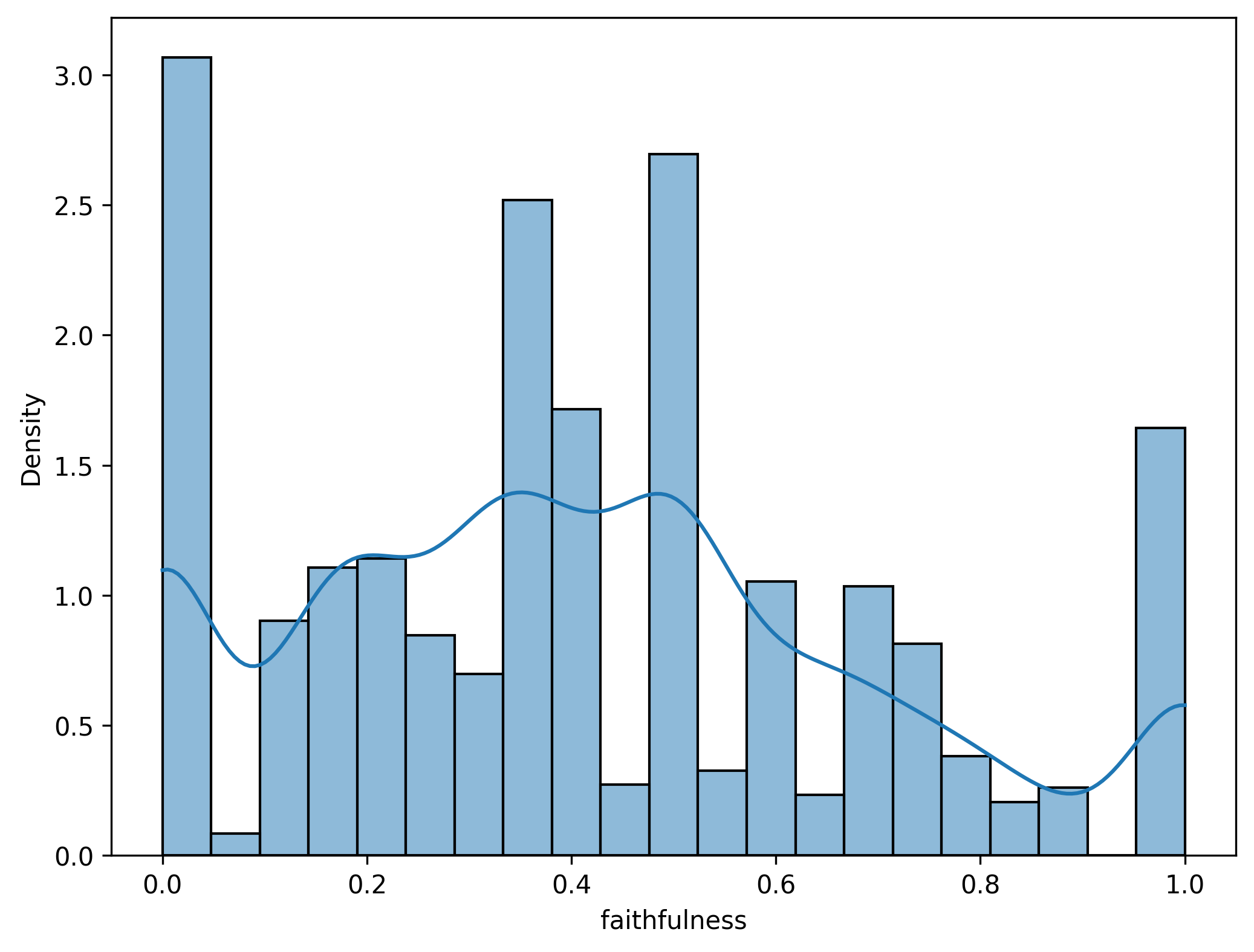}
\caption{\method\ faithfulness distribution for Ledgar for \texttt{gemma-2-27b}.}  
\label{fig:faithfulness_distrib_ledgar}
\end{figure}

Faithfulness is finally calculated based on concept importance as follows: $F(x,e) = \frac{1}{p} \sum_{i=1}^{p} \mathbf{1}_{\texttt{I}_\Gamma(c_i) > 0}$. Figures~\ref{fig:faithfulness_distrib_agnews} and~\ref{fig:faithfulness_distrib_ledgar} show the faithfulness distributions of \method\ for AGNews and Ledgar for \texttt{gemma-2-27b}.

\begin{figure}[H]{\centering}
\centering
\includegraphics[width=0.75\linewidth]{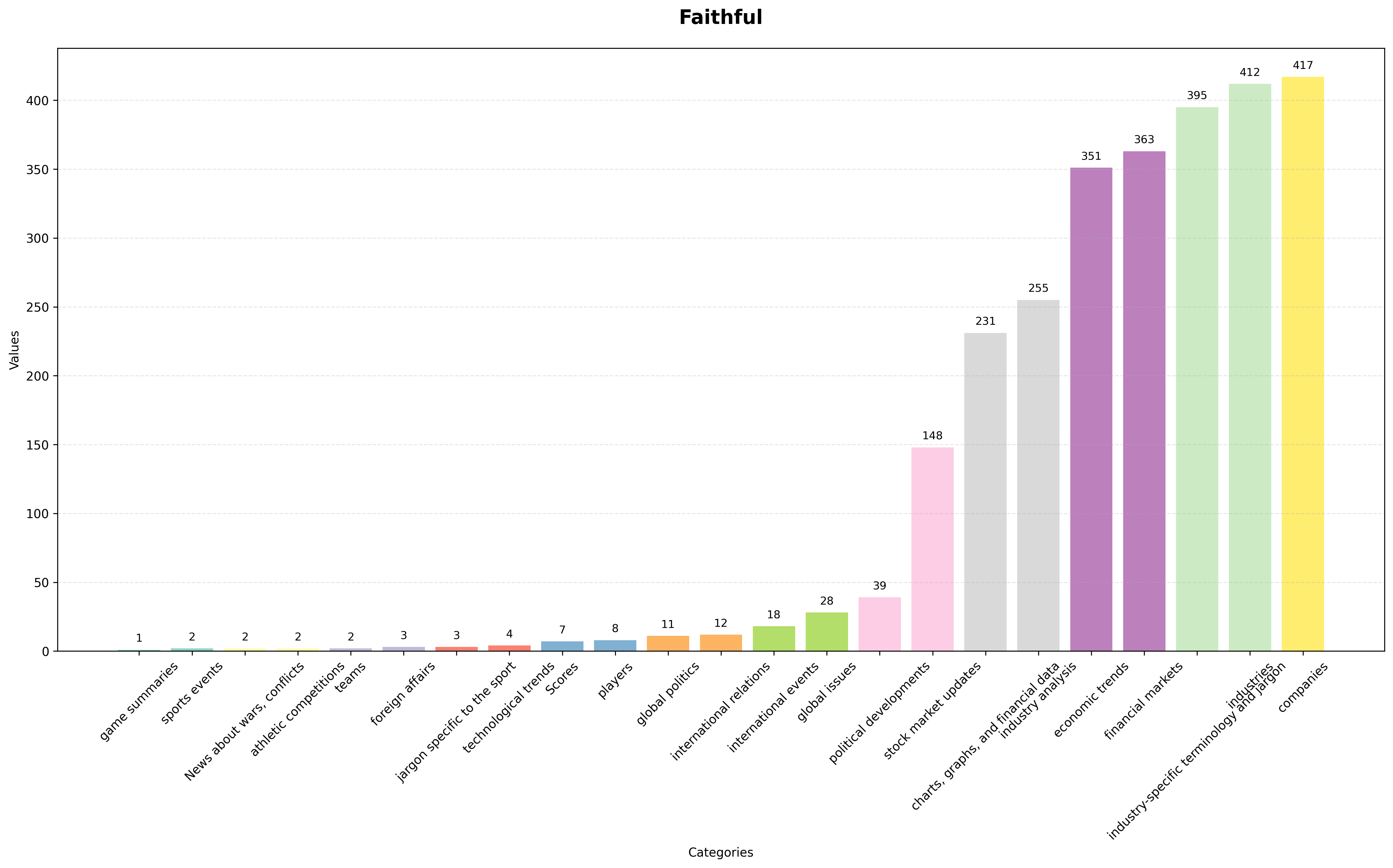}
\caption{Concepts related to faithful self-NLE and the prediction "business", sorted by frequency for AGNews for \texttt{gemma-2-2b}.}  
\label{fig:concept_frequency_business_faithful_2b_agnews}
\end{figure}

\begin{figure}[H]{\centering}
\centering
\includegraphics[width=0.75\linewidth]{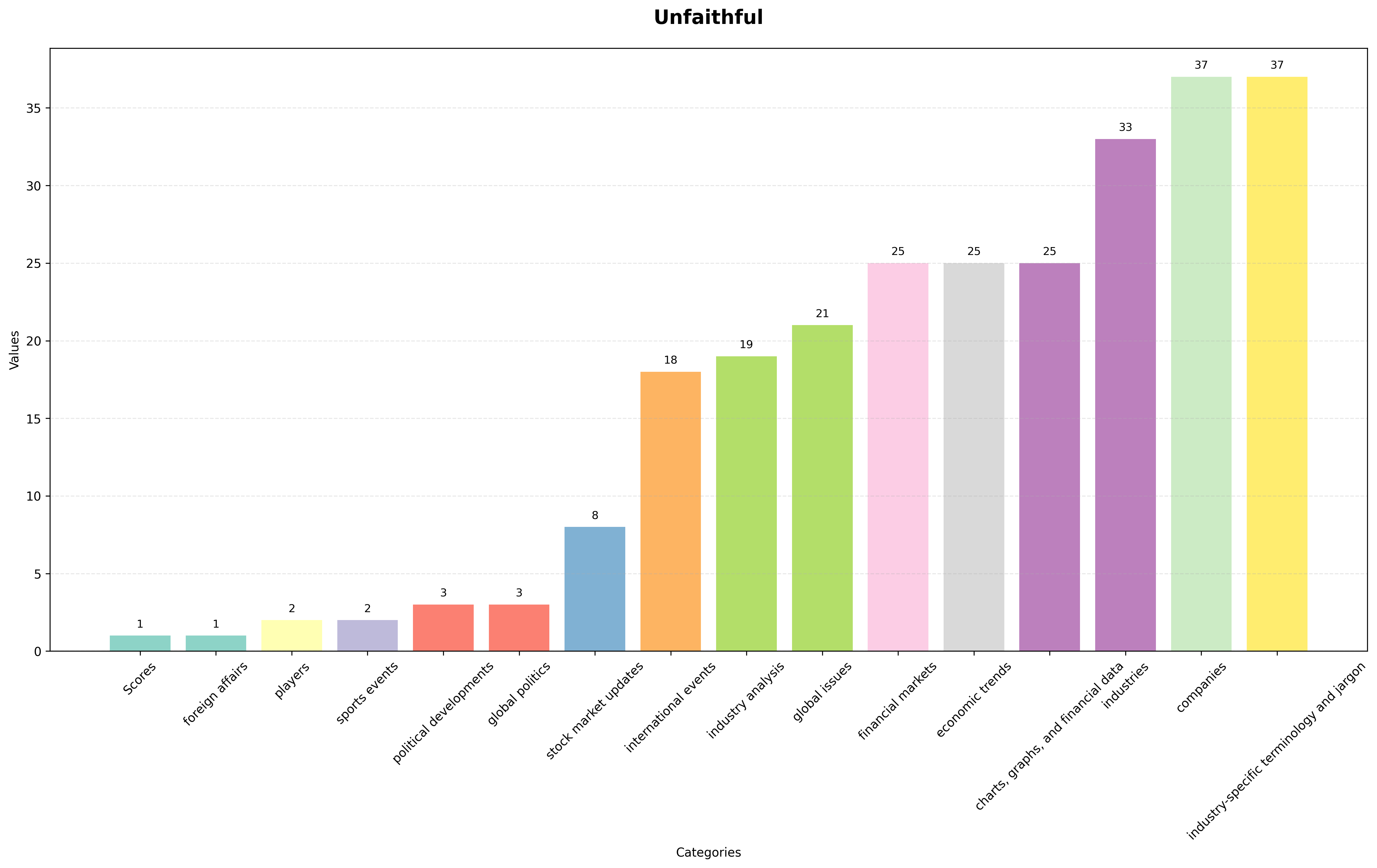}
\caption{Concepts related to unfaithful self-NLE and the prediction "business", sorted by frequency for AGNews for \texttt{gemma-2-2b}.}  
\label{fig:concept_frequency_business_unfaithful_2b_agnews}
\end{figure}

\begin{figure}[H]{\centering}
\centering
\includegraphics[width=0.75\linewidth]{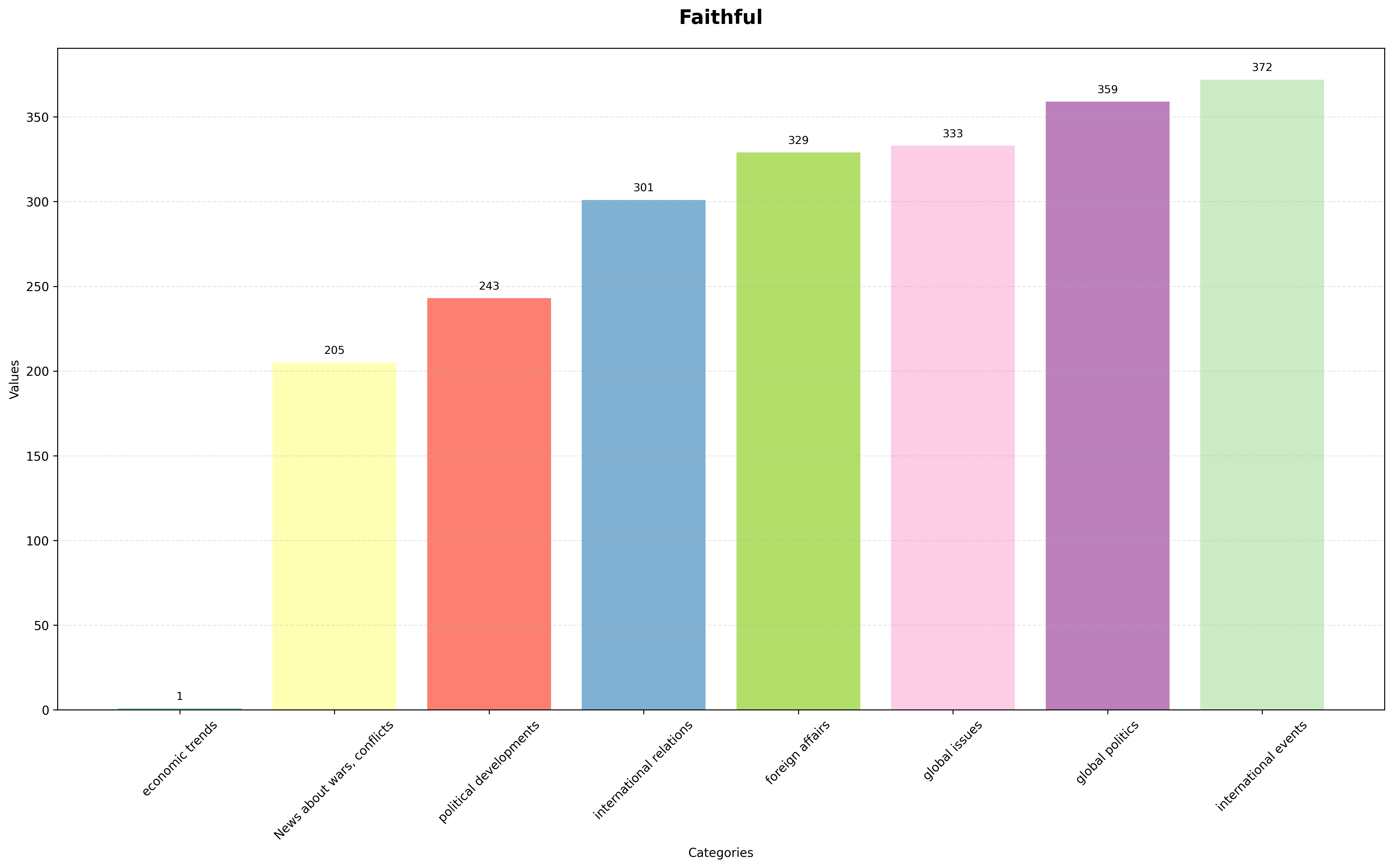}
\caption{Concepts related to faithful self-NLE and the prediction "world", sorted by frequency for AGNews for \texttt{gemma-2-2b}.}  
\label{fig:concept_frequency_world_faithful_2b_agnews}
\end{figure}

\begin{figure}[H]{\centering}
\centering
\includegraphics[width=0.75\linewidth]{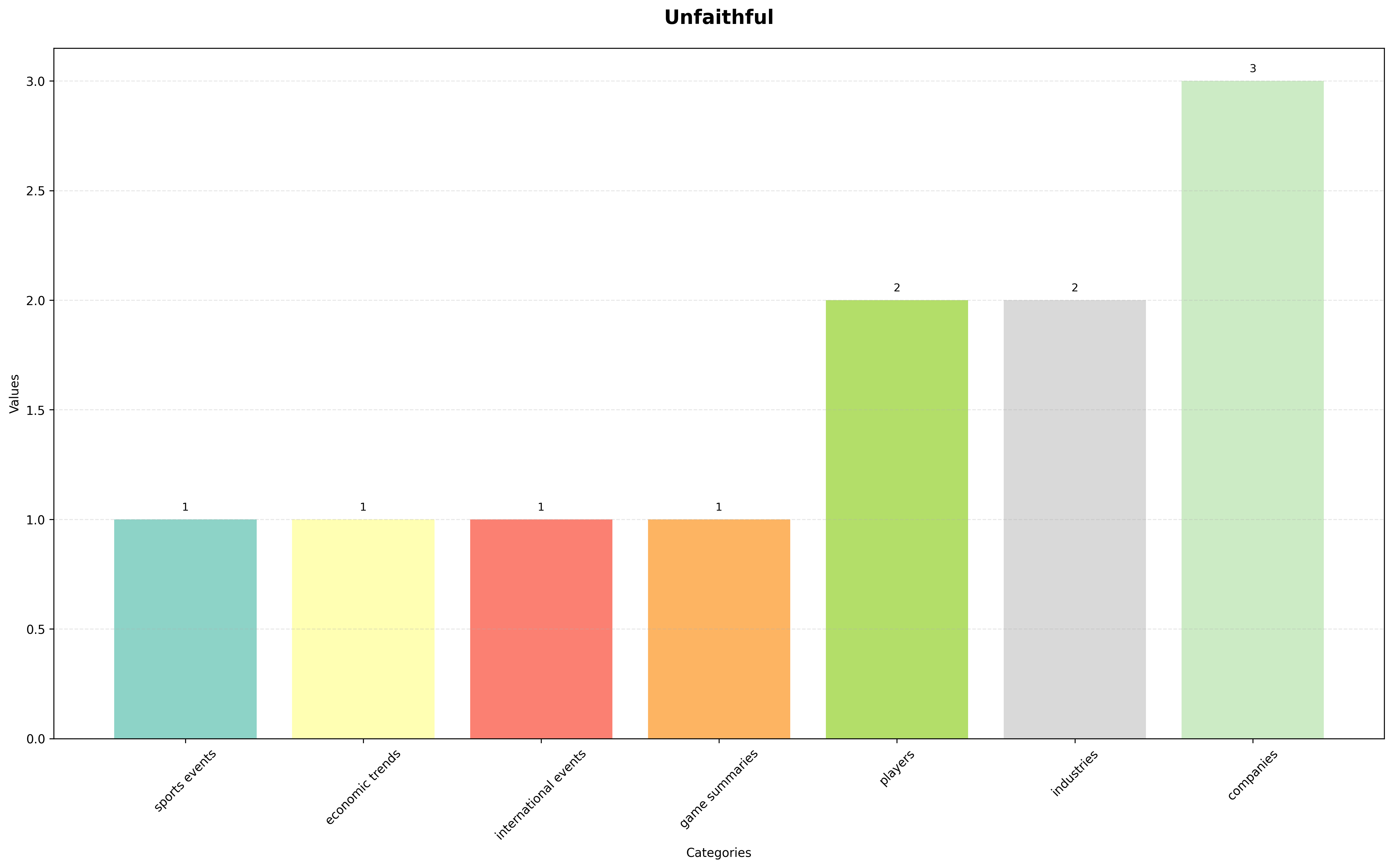}
\caption{Concepts related to unfaithful self-NLE and the prediction "world", sorted by frequency for AGNews for \texttt{gemma-2-2b}.}  
\label{fig:concept_frequency_world_unfaithful_2b_agnews}
\end{figure}

\begin{figure}[H]{\centering}
\centering
\includegraphics[width=0.75\linewidth]{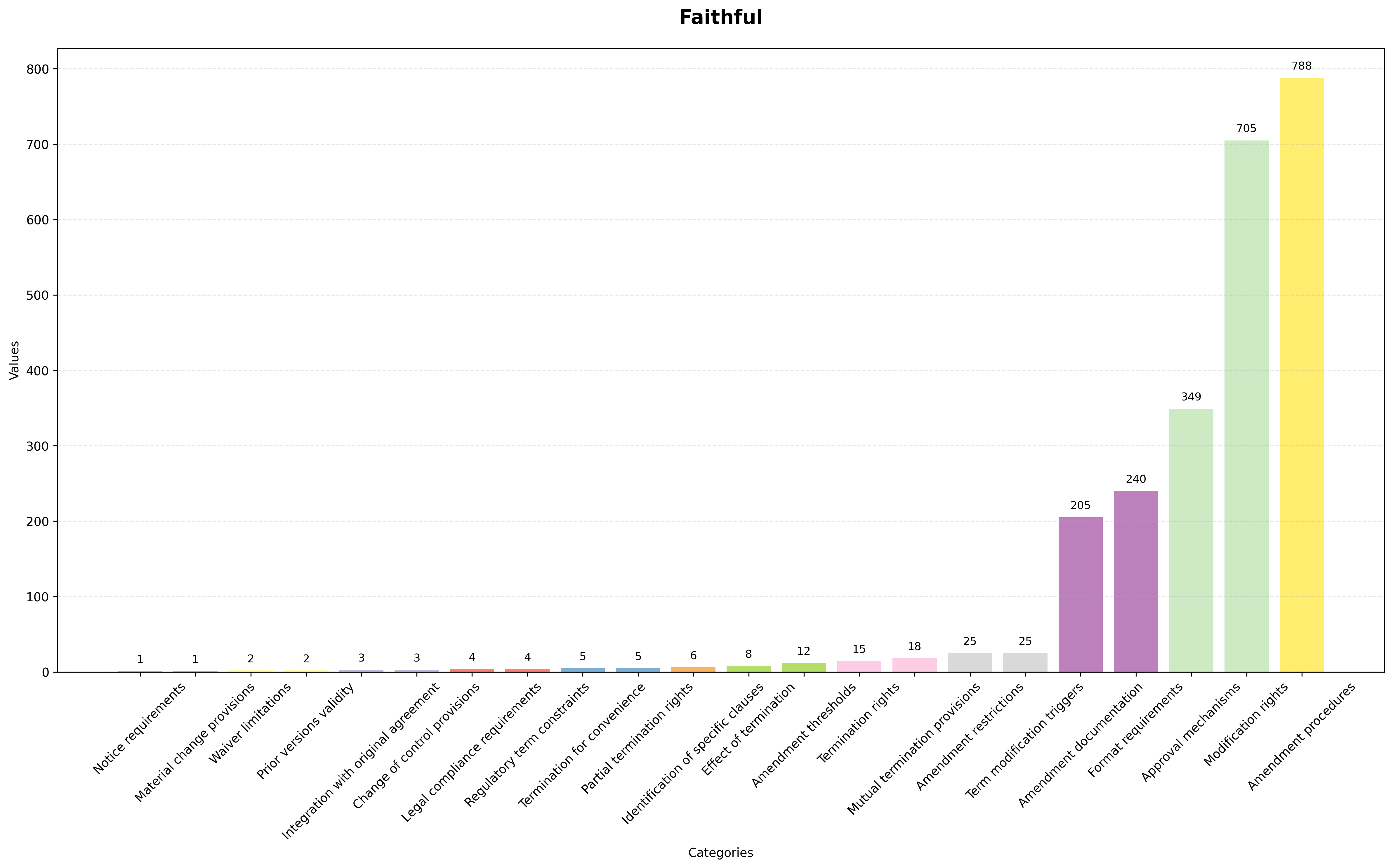}
\caption{Concepts related to faithful self-NLE and the prediction "amendments", sorted by frequency for Ledgar for \texttt{gemma-2-9b}.}  
\label{fig:concept_frequency_amendments_faithful_9b_agnews}
\end{figure}

\begin{figure}[H]{\centering}
\centering
\includegraphics[width=0.75\linewidth]{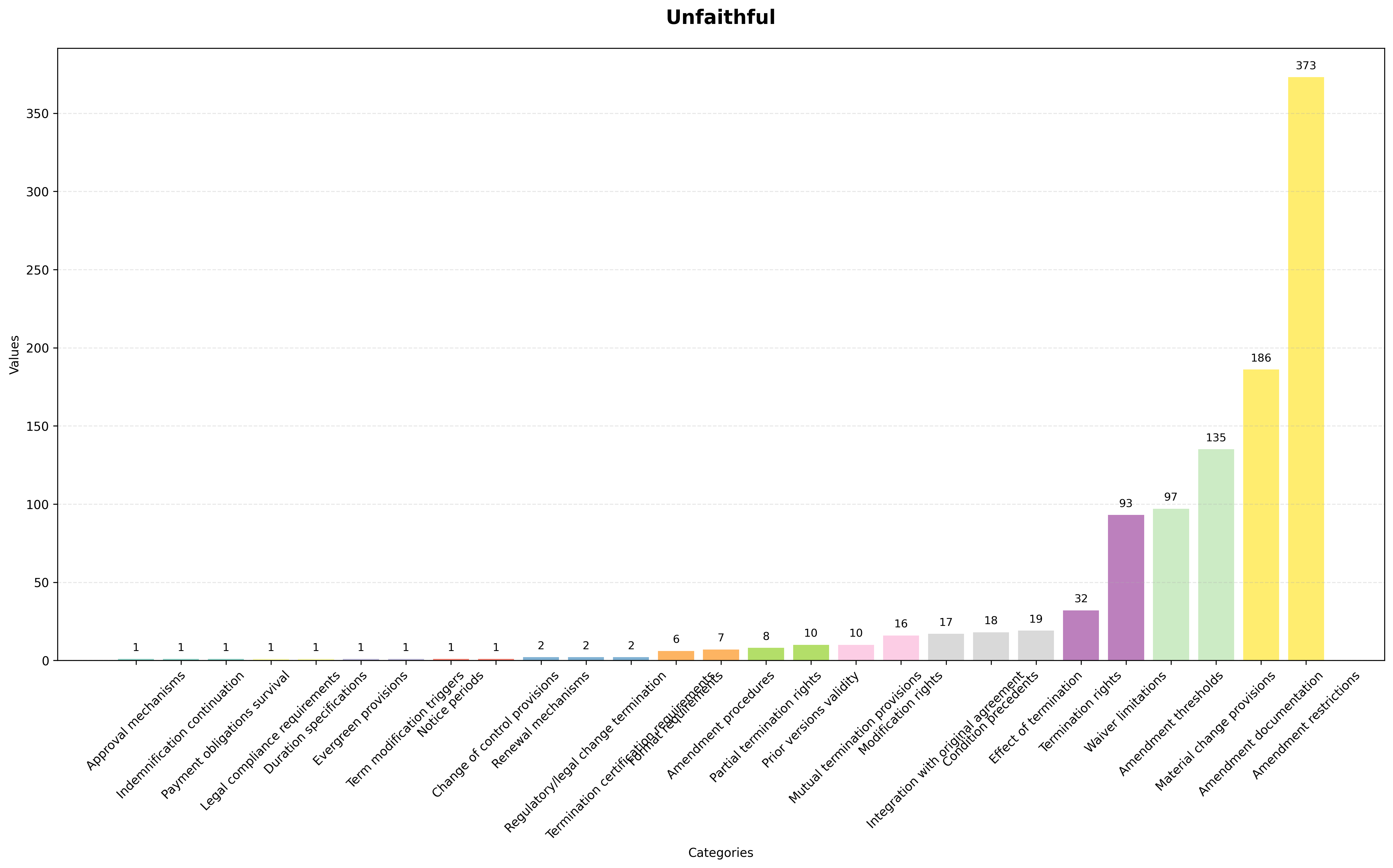}
\caption{Concepts related to unfaithful self-NLE and the prediction "amendments", sorted by frequency for Ledgar for \texttt{gemma-2-9b}.}  
\label{fig:concept_frequency_amendments_unfaithful_9b_agnews}
\end{figure}

In Figure~\ref{fig:concept_frequency_business_faithful_2b_agnews}-\ref{fig:concept_frequency_amendments_unfaithful_9b_agnews} we plot the concepts sorted by frequency for faithful and unfaithful self-NLE for AGNews and Ledgar for \texttt{gemma-2-2b} and \texttt{gemma-2-9b} for several class predictions. These figures reveal which concepts are associated with faithful versus unfaithful self-NLEs. For instance, Figure~\ref{fig:concept_frequency_world_unfaithful_2b_agnews} shows that \method\ correctly identifies counterintuitive associations (e.g., the concept "companies" linked to the class "world") as unfaithful, demonstrating its reliability.

We also plot the density of faithfulness for AGNews and Ledgar for \texttt{gemma-2-27b} in Figure~\ref{fig:faithfulness_distrib_agnews} and~\ref{fig:faithfulness_distrib_ledgar}.

\paragraph{2-hop Reasoning Additional Results.}

We show in Table~\ref{tab:results_faithfulness_analysis_phi} the results obtained by applying \method\ in the case of 2-hop reasoning with \texttt{Phi-4} as bridge object extractor. The results are highly similar to the ones obtained by using \texttt{Qwen-3-32B} as bridge object extractor (see column "\textit{Faithfulness Corr w/ qwen}").

\begin{table*}[]
\centering
\small
\begin{tabular*}{\textwidth}{@{\extracolsep{\fill}}lcccccccc@{}}
\toprule
\textbf{Model} & \textbf{Judge} & \multicolumn{2}{c}{\textbf{Self-NLE}} & \multicolumn{2}{c}{\textbf{Latent Hop 1}} & \multicolumn{2}{c}{\textbf{Self-NLE}} & \textbf{Faithfulness} \\
& & \multicolumn{2}{c}{\textbf{Correctness}} & \multicolumn{2}{c}{\textbf{Correctness}} & \multicolumn{2}{c}{\textbf{Faithfulness}} & \textbf{Corr w/ qwen} \\
\cmidrule(lr){3-4} \cmidrule(lr){5-6} \cmidrule(lr){7-8}
& & Accurate & Inaccurate & Accurate & Inaccurate & Accurate & Inaccurate & \\
\midrule
\texttt{gemma-2-2b}  & Phi-4 & 57.80\% & 56.50\% & 47.50\% & 48.30\% & 48.00\% & 56.50\% & 86.50\% \\
\texttt{gemma-2-9b}  & Phi-4 & 73.70\% & 55.10\% & 58.50\% & 44.30\% & 61.70\% & 54.80\% & 91.20\% \\
\texttt{gemma-2-27b} & Phi-4 & 77.80\% & 55.70\% & 64.90\% & 46.90\% & 69.10\% & 59.80\% & 90.90\% \\
\bottomrule
\end{tabular*}
\caption{Self-NLE correctness, first hop latent reasoning correctness, and self-NLE faithfulness across models, evaluated using Phi-4 as judge. "(In)accurate" represents the set of predictions initially (in)correct. "Faithfulness Corr w/ qwen" shows faithfulness correlation between faithfulness scores obtained based on \texttt{Phi-4} and \texttt{Qwen-3-32B}.}
\label{tab:results_faithfulness_analysis_phi}
\end{table*}

\begin{table*}[]
\centering
\small
\begin{tabular*}{\textwidth}{@{\extracolsep{\fill}}lccccccc@{}}
\toprule
\textbf{Model} & \textbf{Task Acc.} & \multicolumn{2}{c}{\textbf{Self-NLE}} & \multicolumn{2}{c}{\textbf{Latent Hop 1}} & \multicolumn{2}{c}{\textbf{Self-NLE}} \\
& & \multicolumn{2}{c}{\textbf{Correctness}} & \multicolumn{2}{c}{\textbf{Correctness}} & \multicolumn{2}{c}{\textbf{Faithfulness}} \\
\cmidrule(lr){3-4} \cmidrule(lr){5-6} \cmidrule(lr){7-8}
& & Accurate & Inaccurate & Accurate & Inaccurate & Accurate & Inaccurate \\
\midrule
\texttt{gemma-2-27B} & 3.4\% & 72.8\% & 62.1\% & 36.4\% & 8.1\% & 27.3\% & 7.6\% \\
\bottomrule
\end{tabular*}
\caption{Self-NLE correctness, first hop latent reasoning correctness, and self-NLE faithfulness for \texttt{gemma-2-27B} on the Socrates dataset.}
\label{tab:results_faithfulness_analysis_socrates}
\end{table*}

\begin{figure}[H]{\centering}
\centering
\includegraphics[scale=0.4]{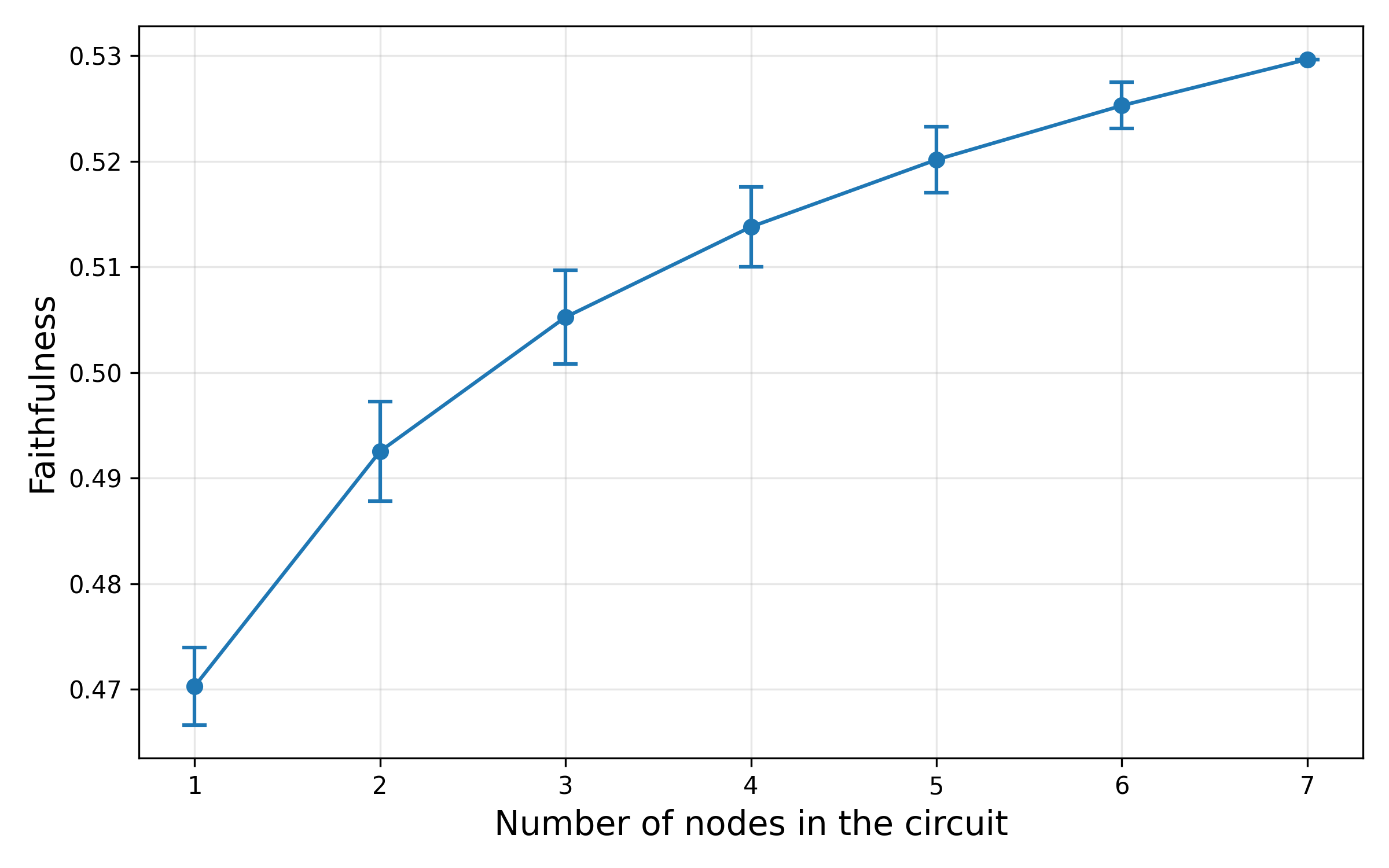}
\caption{\method\ faithfulness sensitivity analysis with regards to circuit size, \texttt{gemma-2-2B}.}  
\label{fig:faithfulness_vs_circuit_size_2b}
\end{figure}

\begin{figure}[H]{\centering}
\centering
\includegraphics[scale=0.4]{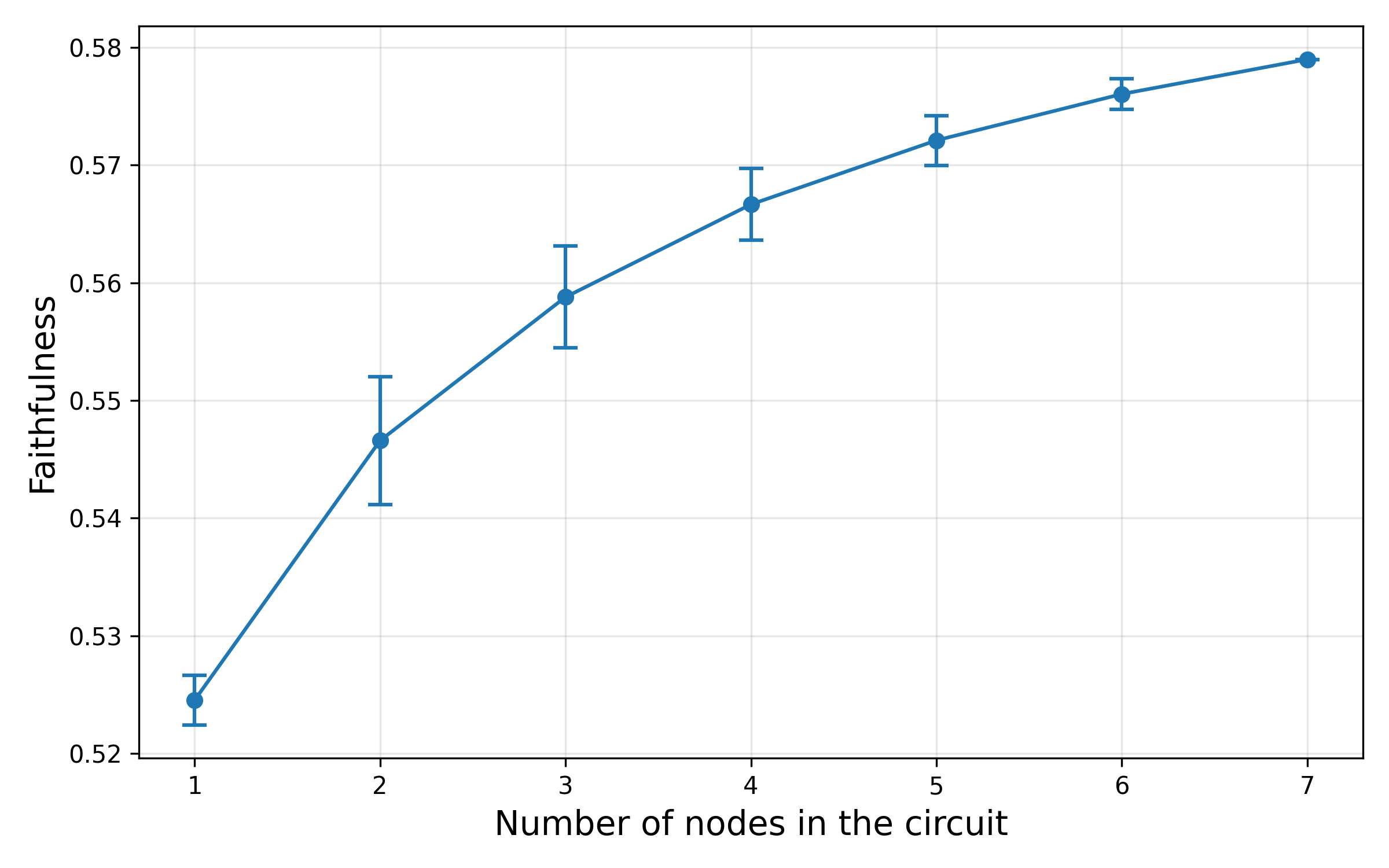}
\caption{\method\ faithfulness sensitivity analysis with regards to circuit size, \texttt{gemma-2-9B}.}  
\label{fig:faithfulness_vs_circuit_size_9b}
\end{figure}

\begin{figure}[H]{\centering}
\centering
\includegraphics[scale=0.4]{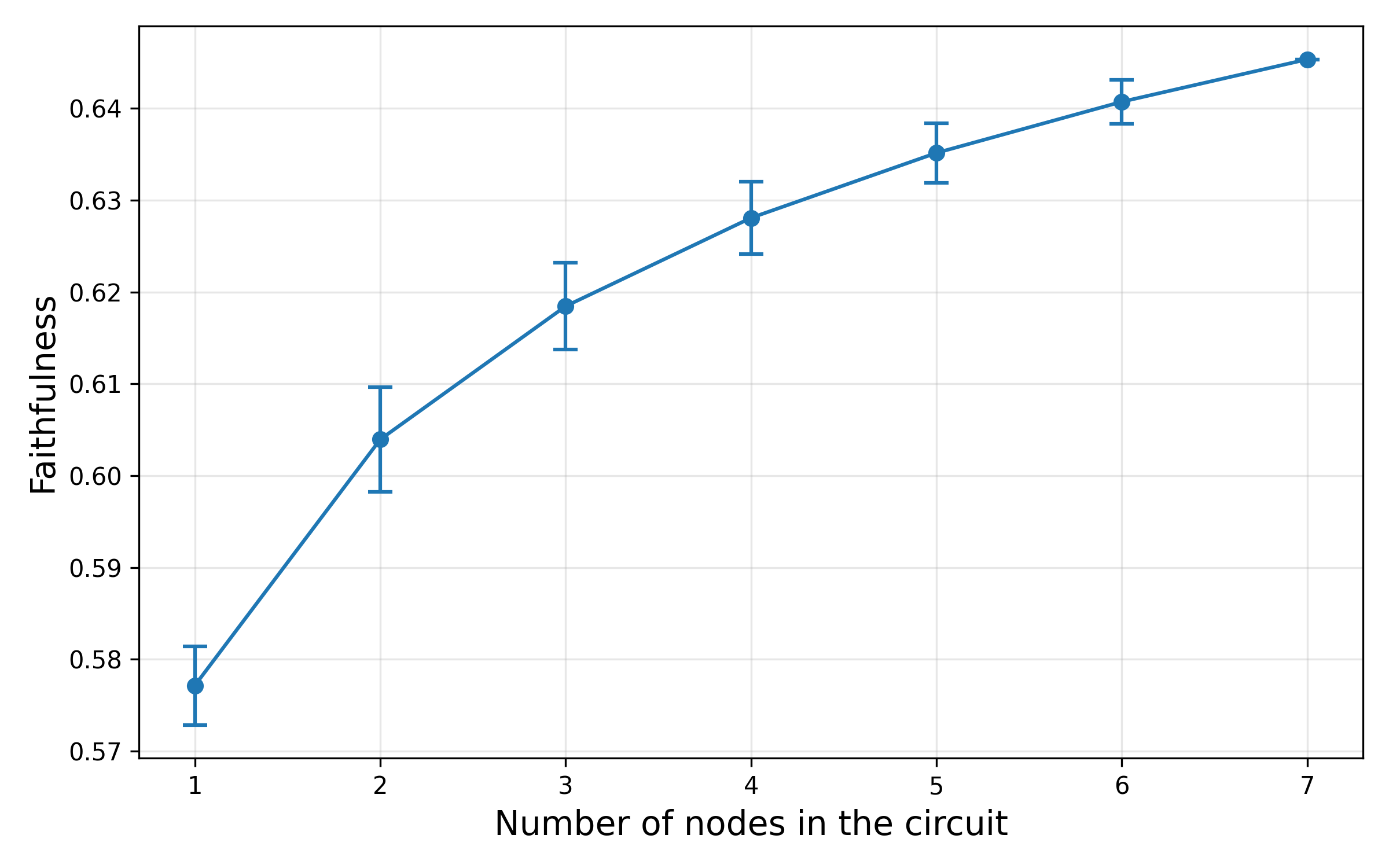}
\caption{\method\ faithfulness sensitivity analysis with regards to circuit size, \texttt{gemma-2-27B}.}  
\label{fig:faithfulness_vs_circuit_size_27b}
\end{figure}

\subsubsection{Linear Latent Faithfulness Detection.}
\label{sec::linear_appendix}
\begin{figure}[H]{\centering}
\centering
\includegraphics[scale=0.3]{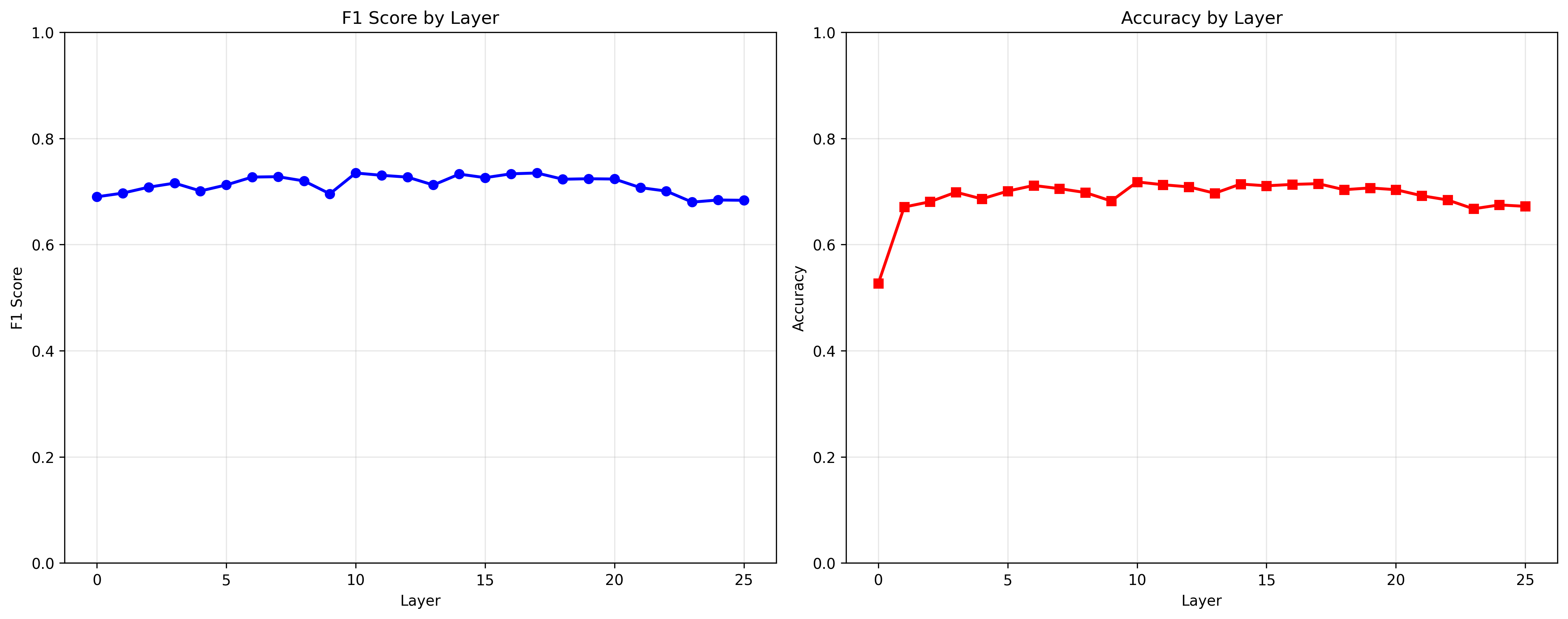}
\caption{Linear faithfulness probe classification performance for 2-hop reasoning, \texttt{gemma-2-2B}.}  
\label{fig:2hop_probe_2b}
\end{figure}

\begin{figure}[H]{\centering}
\centering
\includegraphics[scale=0.3]{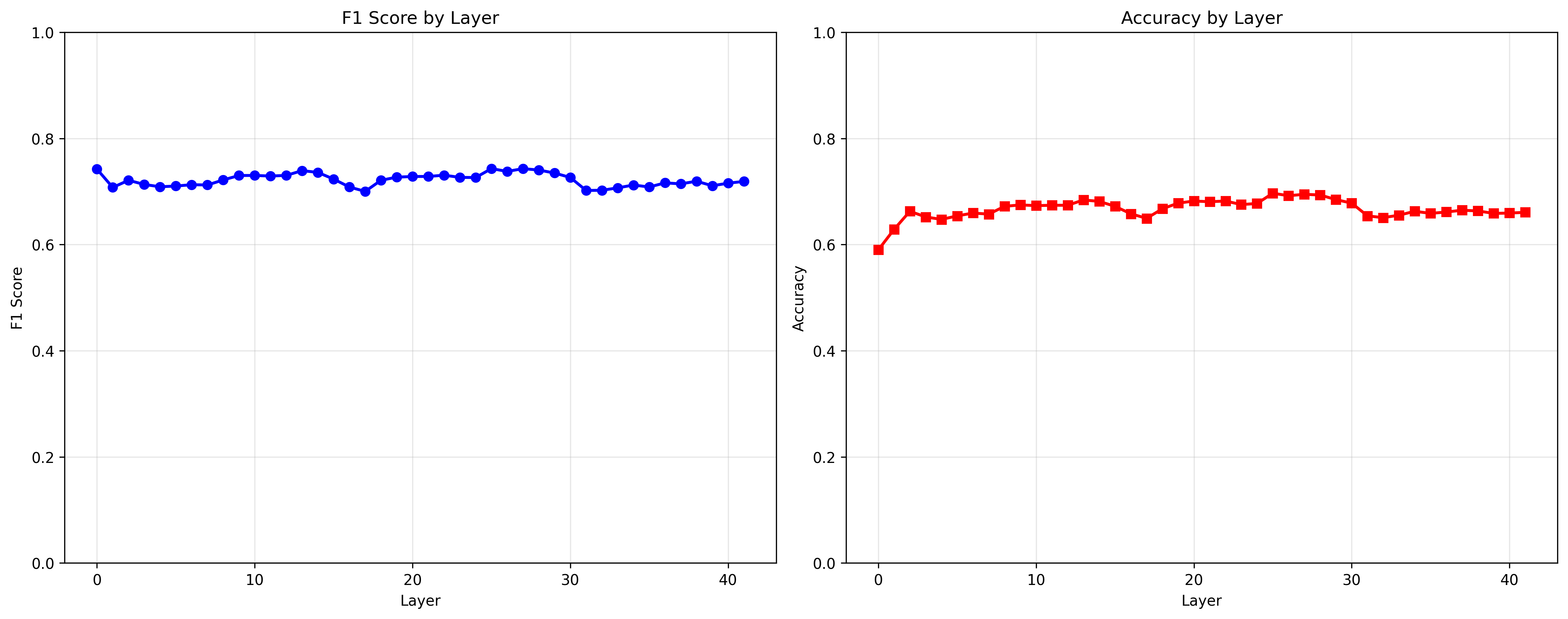}
\caption{Linear faithfulness probe classification performance for 2-hop reasoning, \texttt{gemma-2-9B}.}  
\label{fig:2hop_probe_9b}
\end{figure}

\begin{figure}[H]{\centering}
\centering
\includegraphics[scale=0.3]{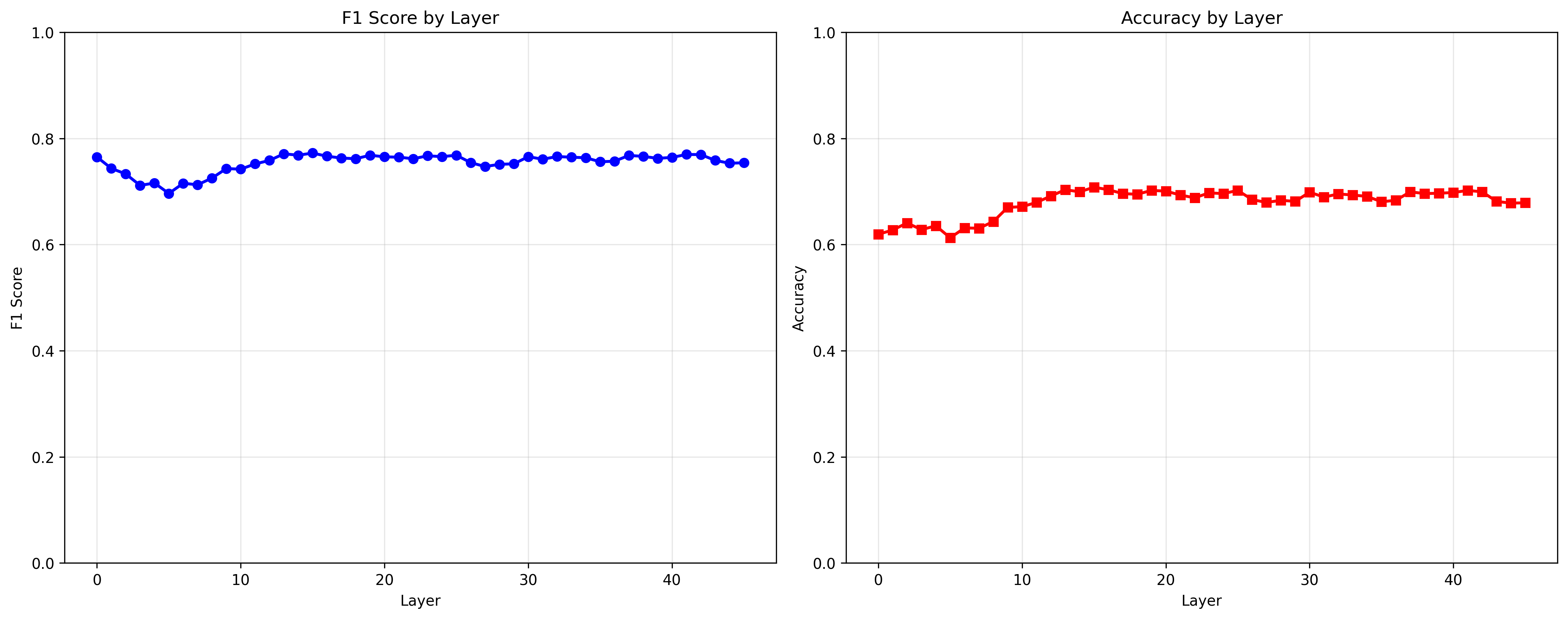}
\caption{Linear faithfulness probe classification performance for 2-hop reasoning, \texttt{gemma-2-27B}.}  
\label{fig:2hop_probe_27b}
\end{figure}

\begin{figure}[H]{\centering}
\centering
\includegraphics[scale=0.3]{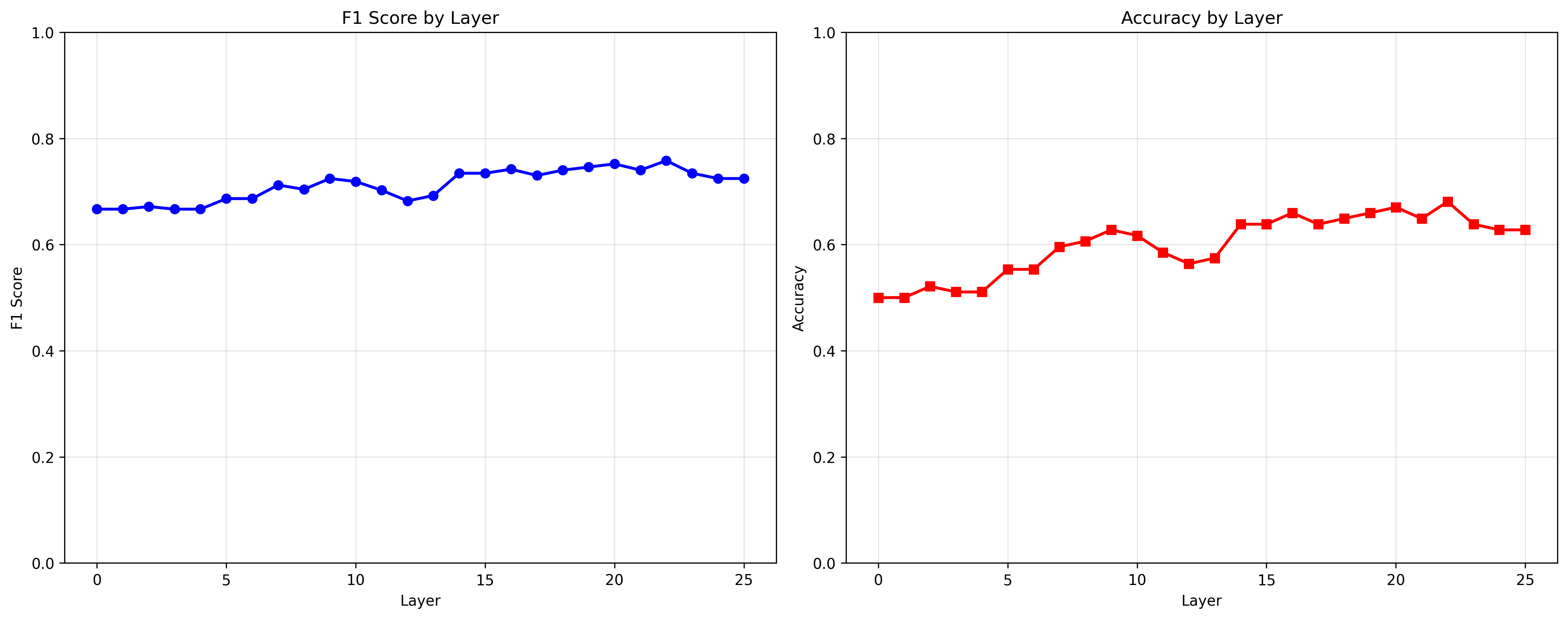}
\caption{Linear faithfulness probe classification performance for AGNews classification, \texttt{gemma-2-2B}.}  
\label{fig:agnews_probe_2b}
\end{figure}

\begin{figure}[H]{\centering}
\centering
\includegraphics[scale=0.3]{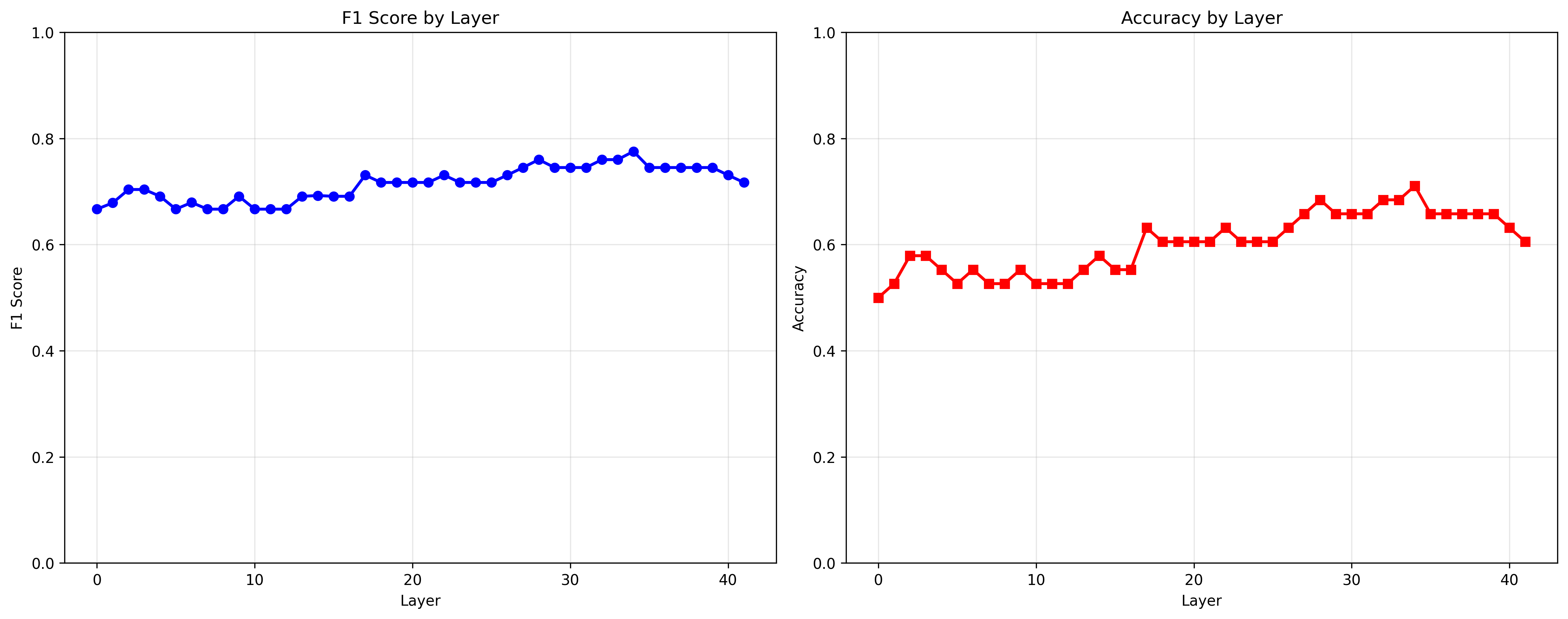}
\caption{Linear faithfulness probe classification performance for AGNews classification, \texttt{gemma-2-9B}.}  
\label{fig:agnews_probe_9b}
\end{figure}

\begin{figure}[H]{\centering}
\centering
\includegraphics[scale=0.3]{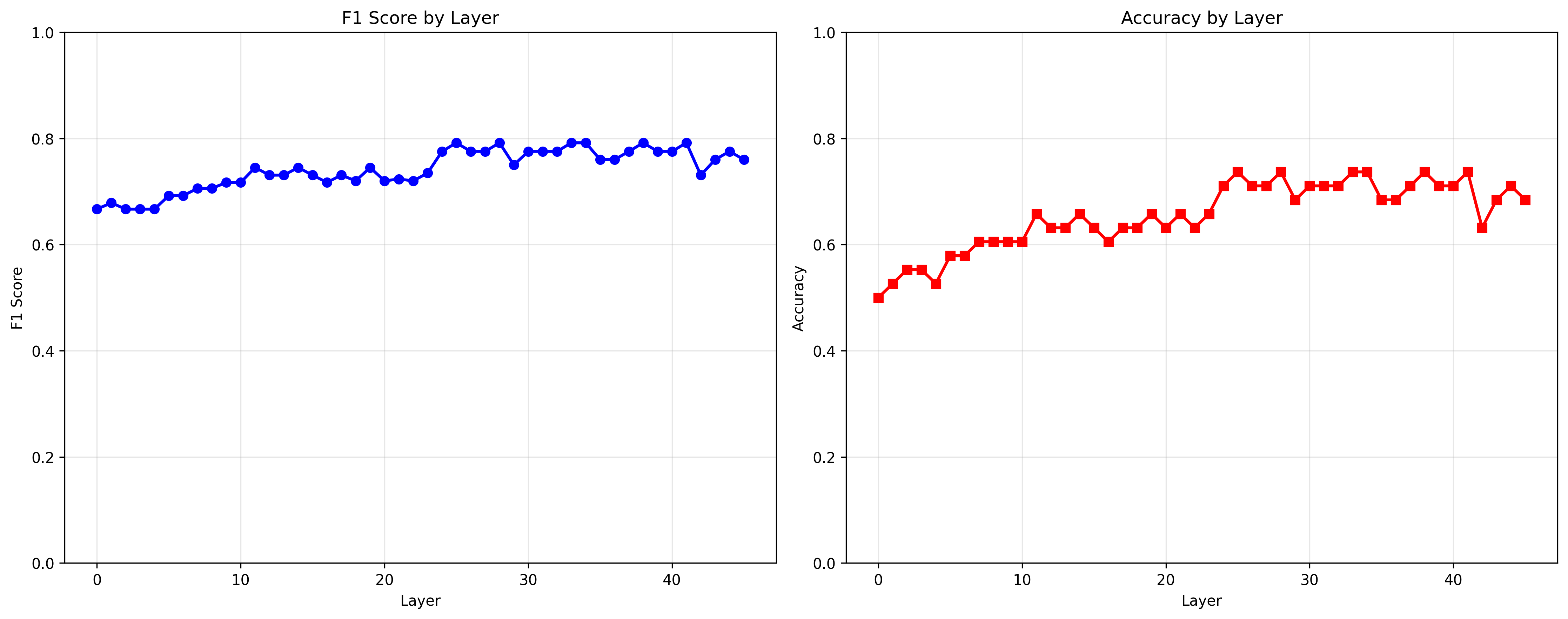}
\caption{Linear faithfulness probe classification performance for AGNews classification, \texttt{gemma-2-27B}.}  
\label{fig:agnews_probe_27b}
\end{figure}

\begin{figure}[H]{\centering}
\centering
\includegraphics[scale=0.3]{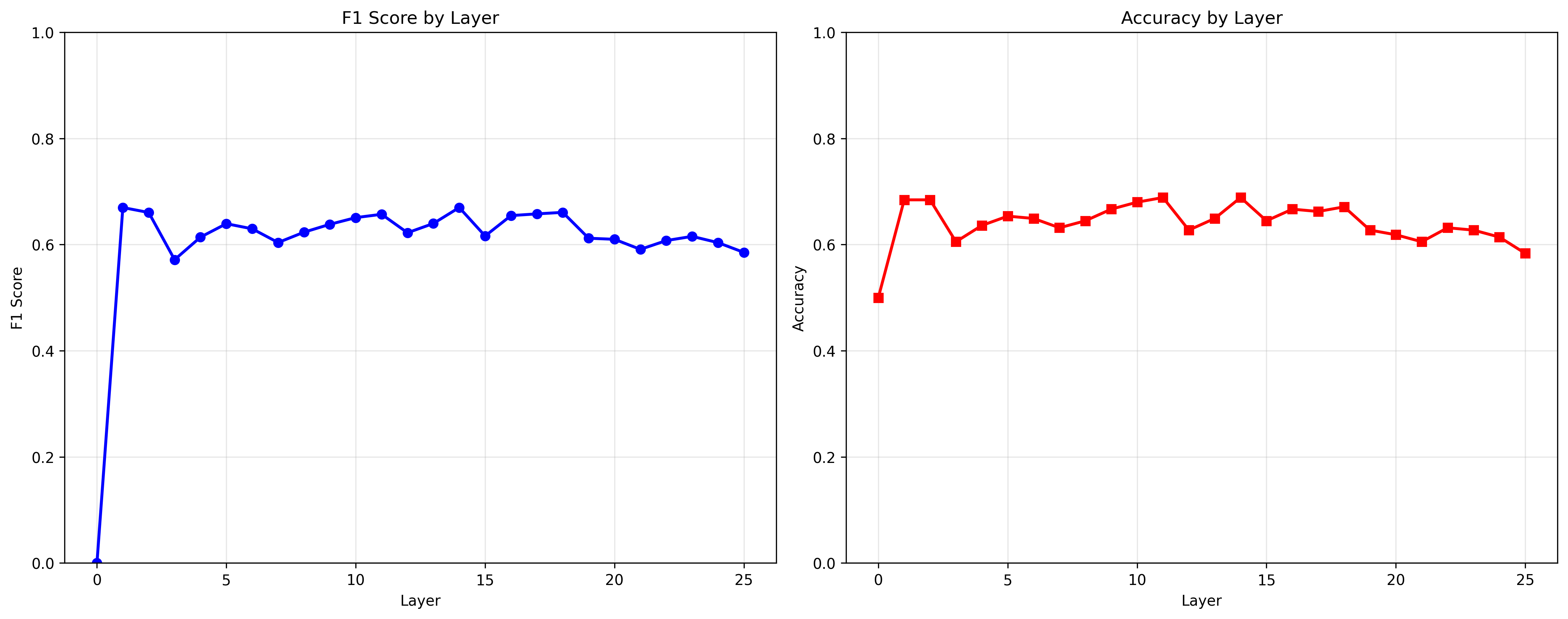}
\caption{Linear faithfulness probe classification performance for Ledgar classification, \texttt{gemma-2-2B}.}  
\label{fig:ledgar_probe_2b}
\end{figure}

\begin{figure}[H]{\centering}
\centering
\includegraphics[scale=0.3]{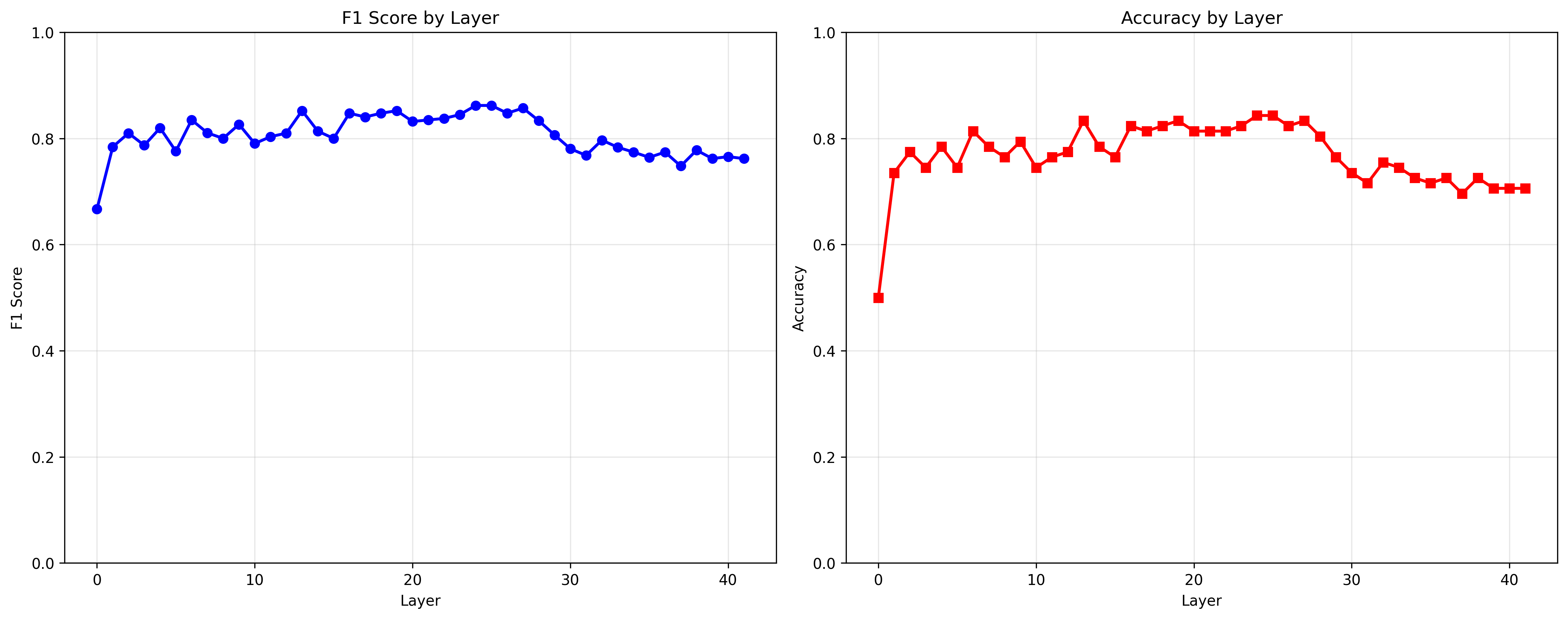}
\caption{Linear faithfulness probe classification performance for Ledgar classification, \texttt{gemma-2-9B}.}  
\label{fig:ledgar_probe_9b}
\end{figure}

\begin{figure}[H]{\centering}
\centering
\includegraphics[scale=0.3]{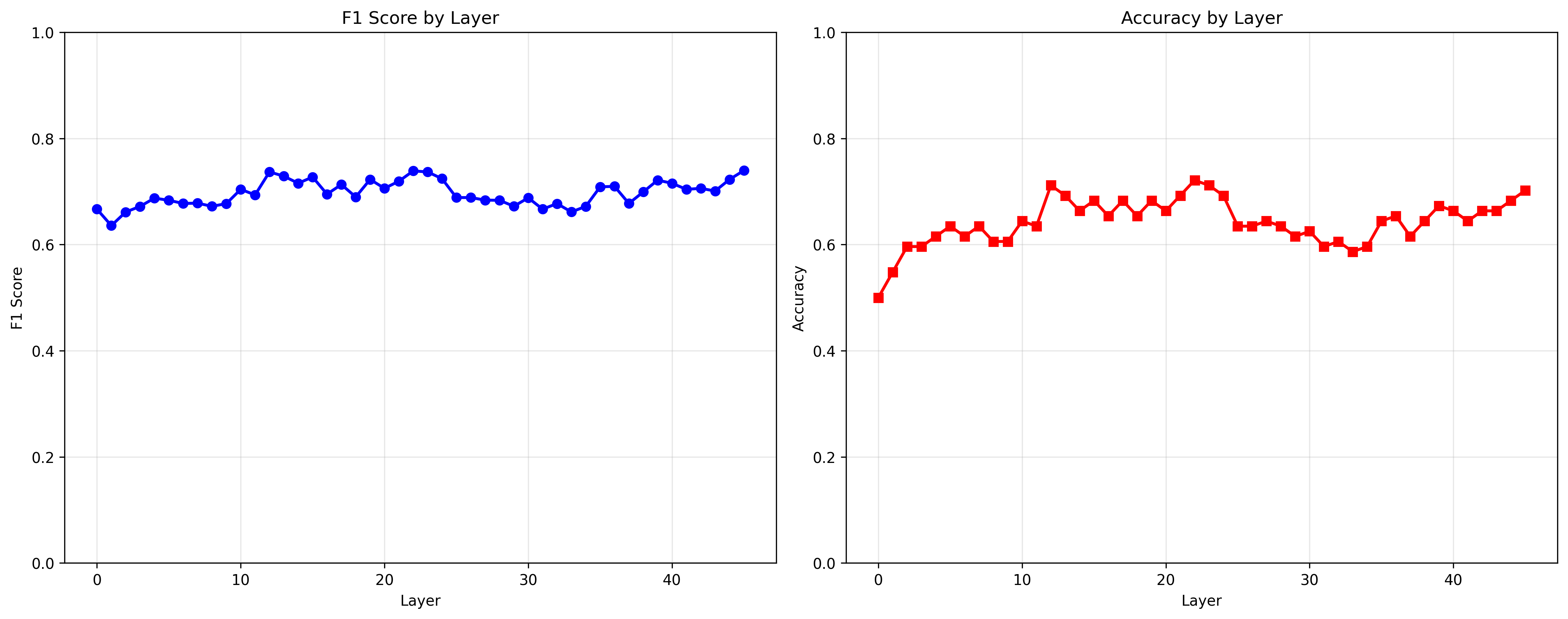}
\caption{Linear faithfulness probe classification performance for Ledgar classification, \texttt{gemma-2-27B}.}  
\label{fig:ledgar_probe_27b}
\end{figure}

\begin{figure}[H]
    \centering
    \begin{subfigure}{0.45\textwidth}
        \centering
        \includegraphics[width=\linewidth]{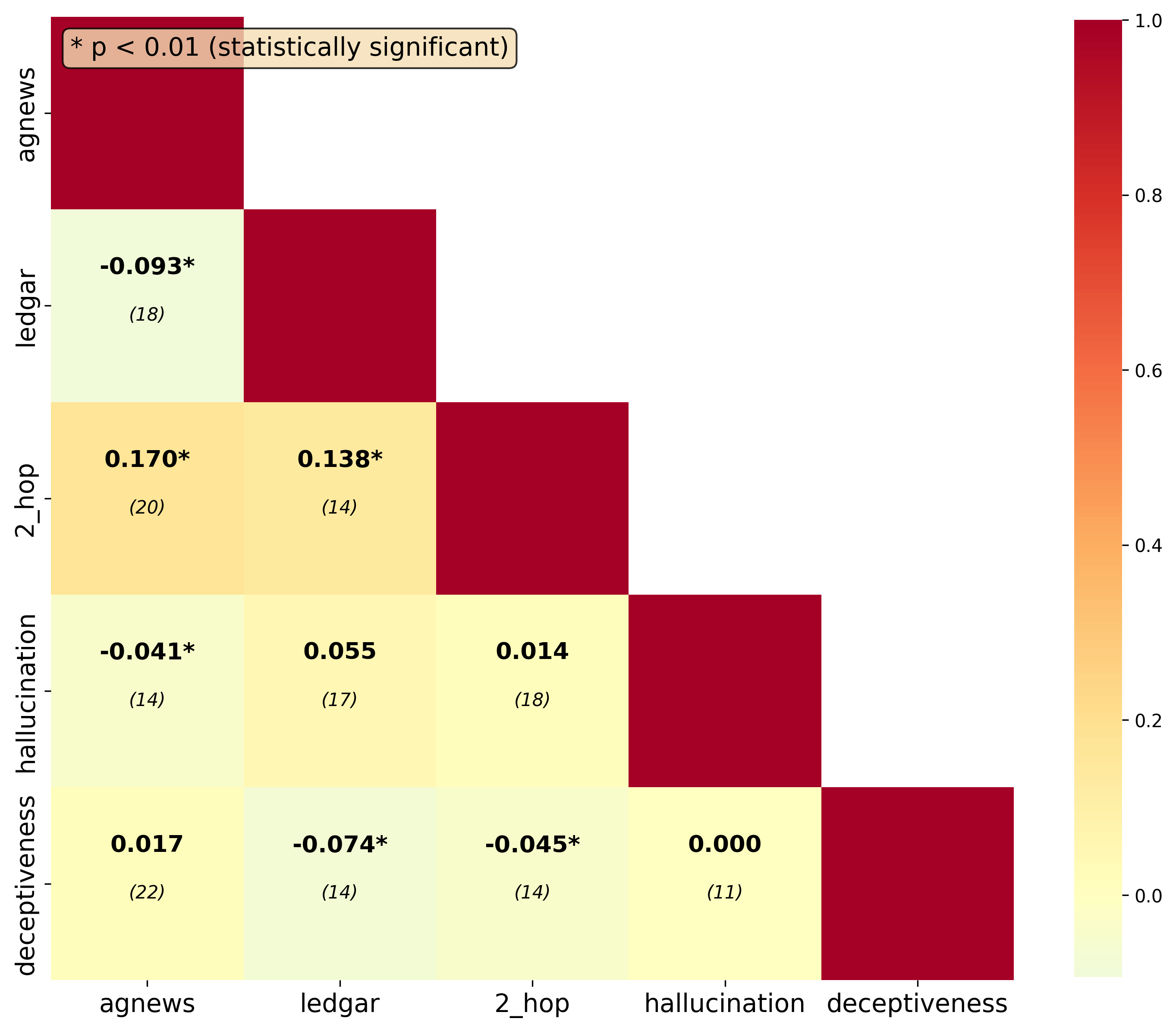}
        \caption{Mean operator}
        \label{fig:chart1}
    \end{subfigure}
    \hfill
    \begin{subfigure}{0.45\textwidth}
        \centering
        \includegraphics[width=\linewidth]{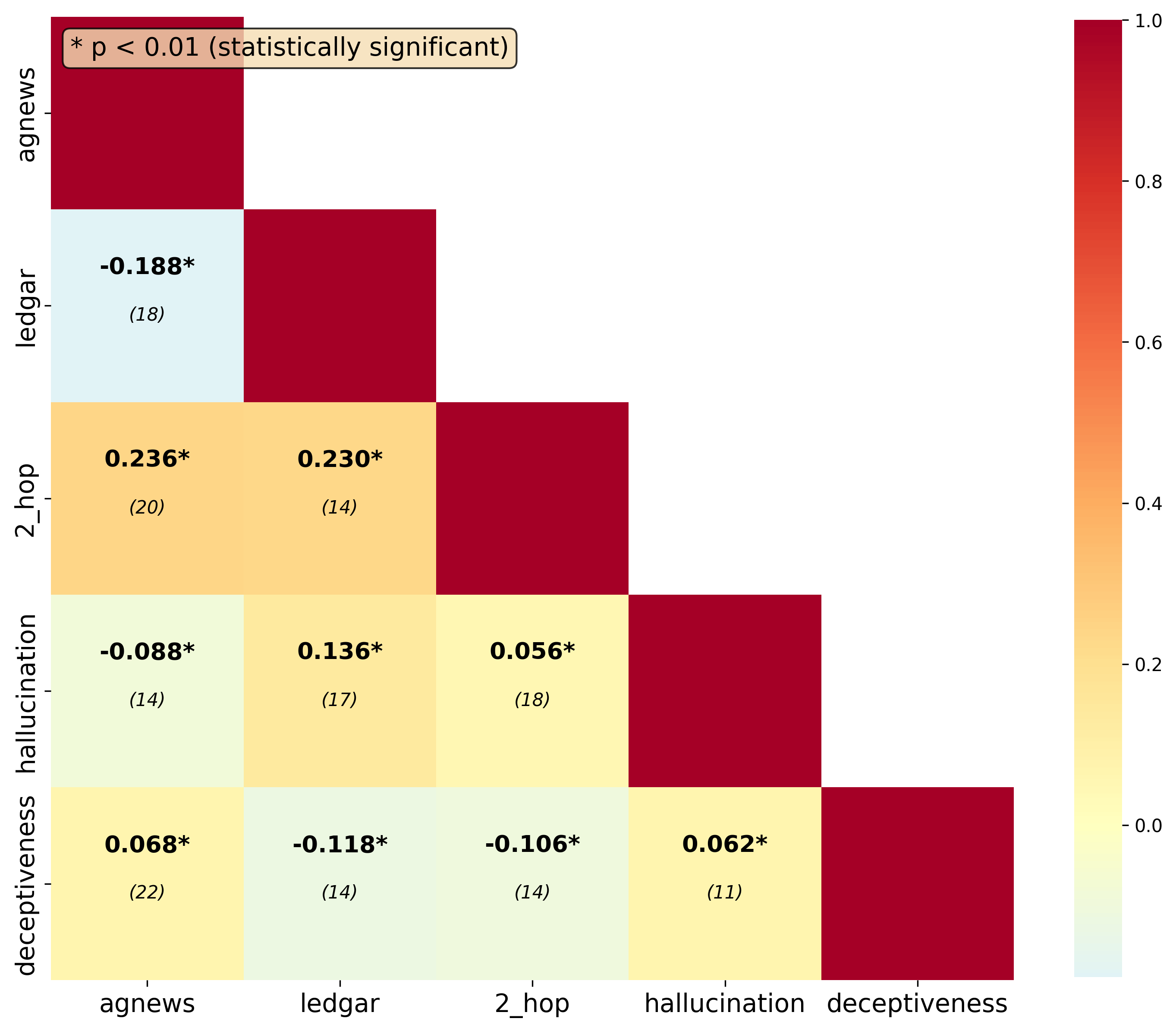}
        \caption{Max operator}
        \label{fig:chart2}
    \end{subfigure}
    \caption{Linear vectors cosine similarity analysis on \texttt{gemma-2-2b}}
    \label{fig:cosine_similatiy_2b}
\end{figure}

\begin{figure}[H]
    \centering
    \begin{subfigure}{0.45\textwidth}
        \centering
        \includegraphics[width=\linewidth]{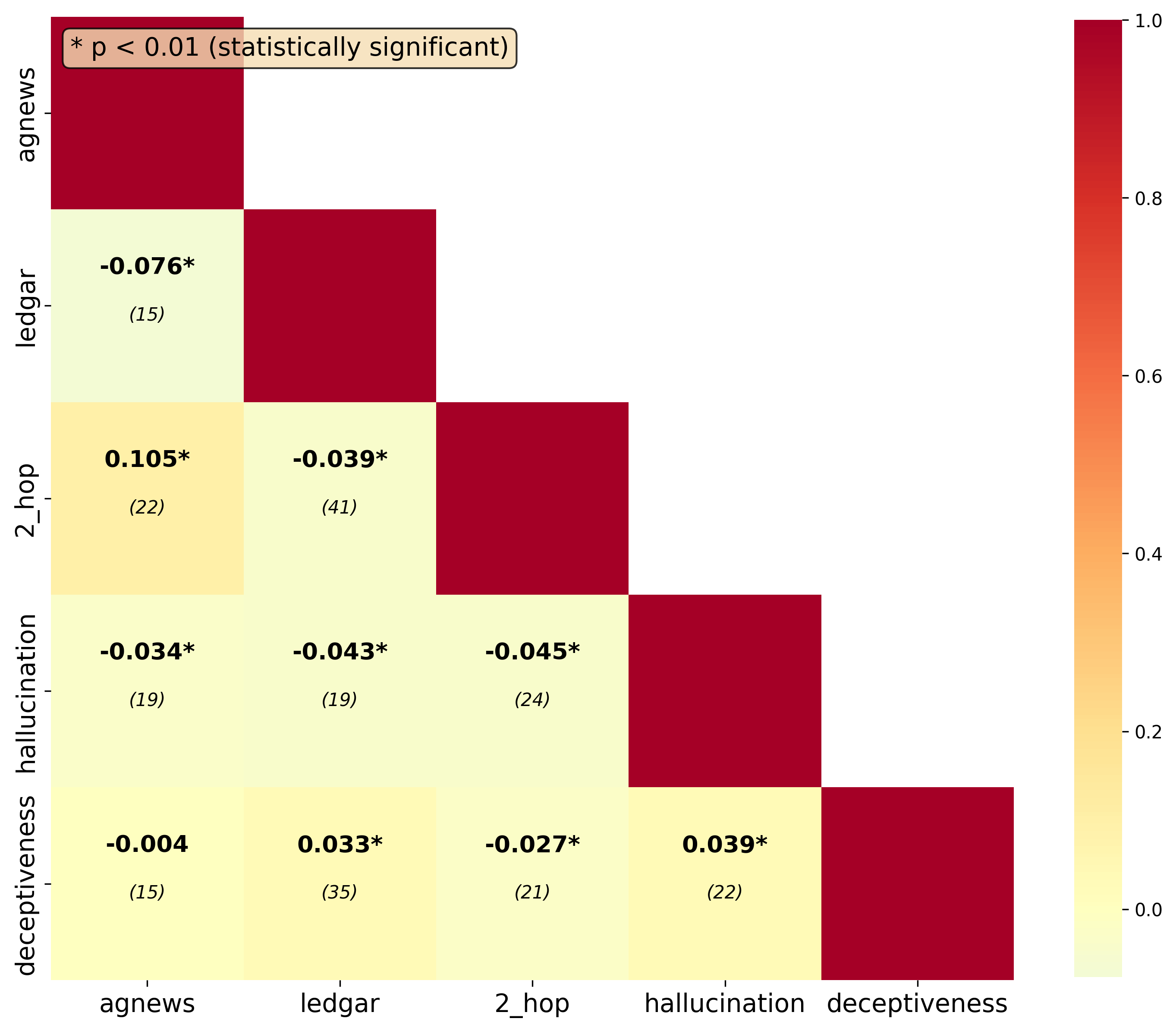}
        \caption{Mean operator}
        \label{fig:chart1}
    \end{subfigure}
    \hfill
    \begin{subfigure}{0.45\textwidth}
        \centering
        \includegraphics[width=\linewidth]{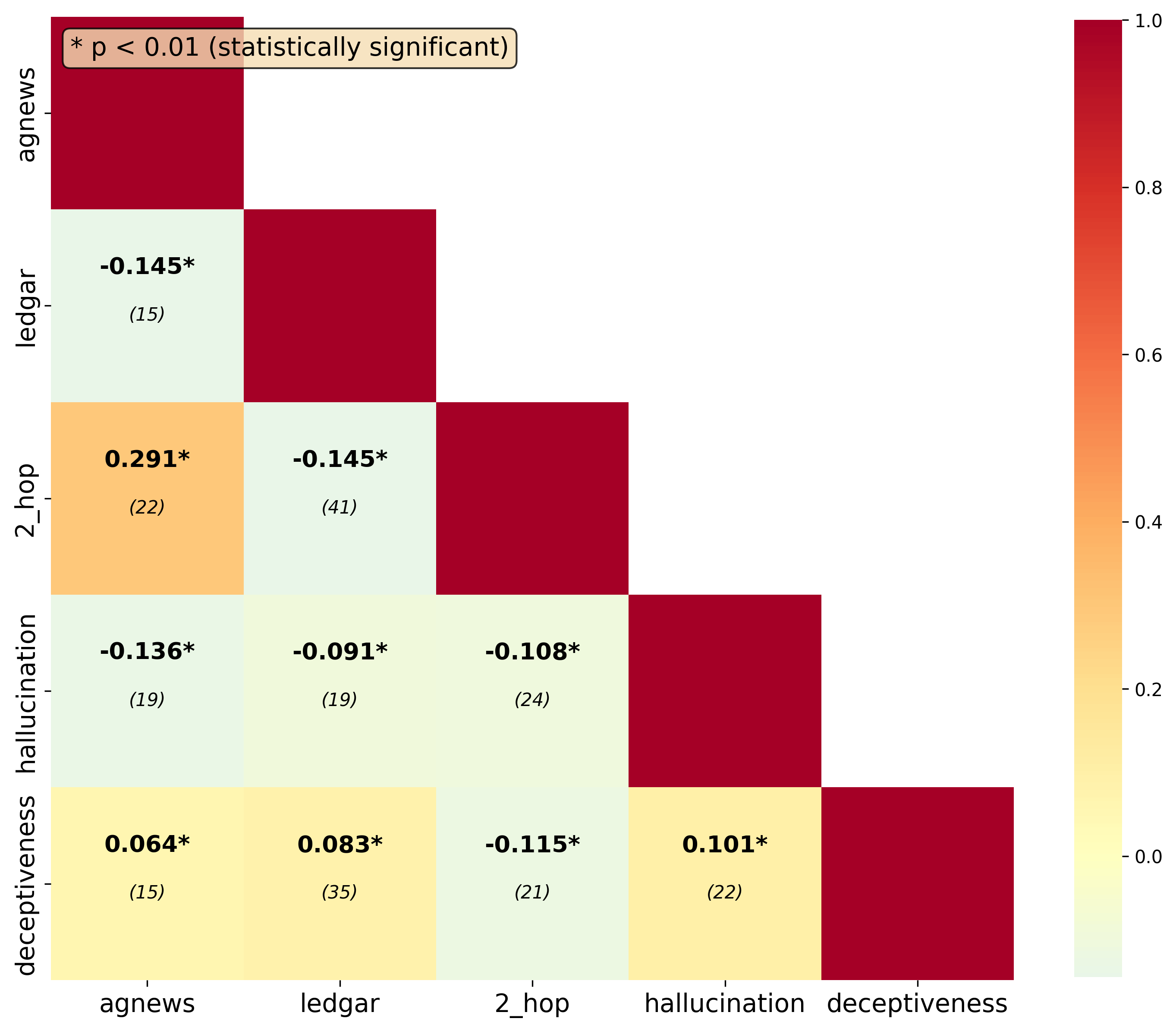}
        \caption{Max operator}
        \label{fig:chart2}
    \end{subfigure}
    \caption{Linear vectors cosine similarity analysis on \texttt{gemma-2-9b}}
    \label{fig:cosine_similatiy_9b}
\end{figure}

\begin{figure}[H]
    \centering
    \begin{subfigure}{0.45\textwidth}
        \centering
        \includegraphics[width=\linewidth]{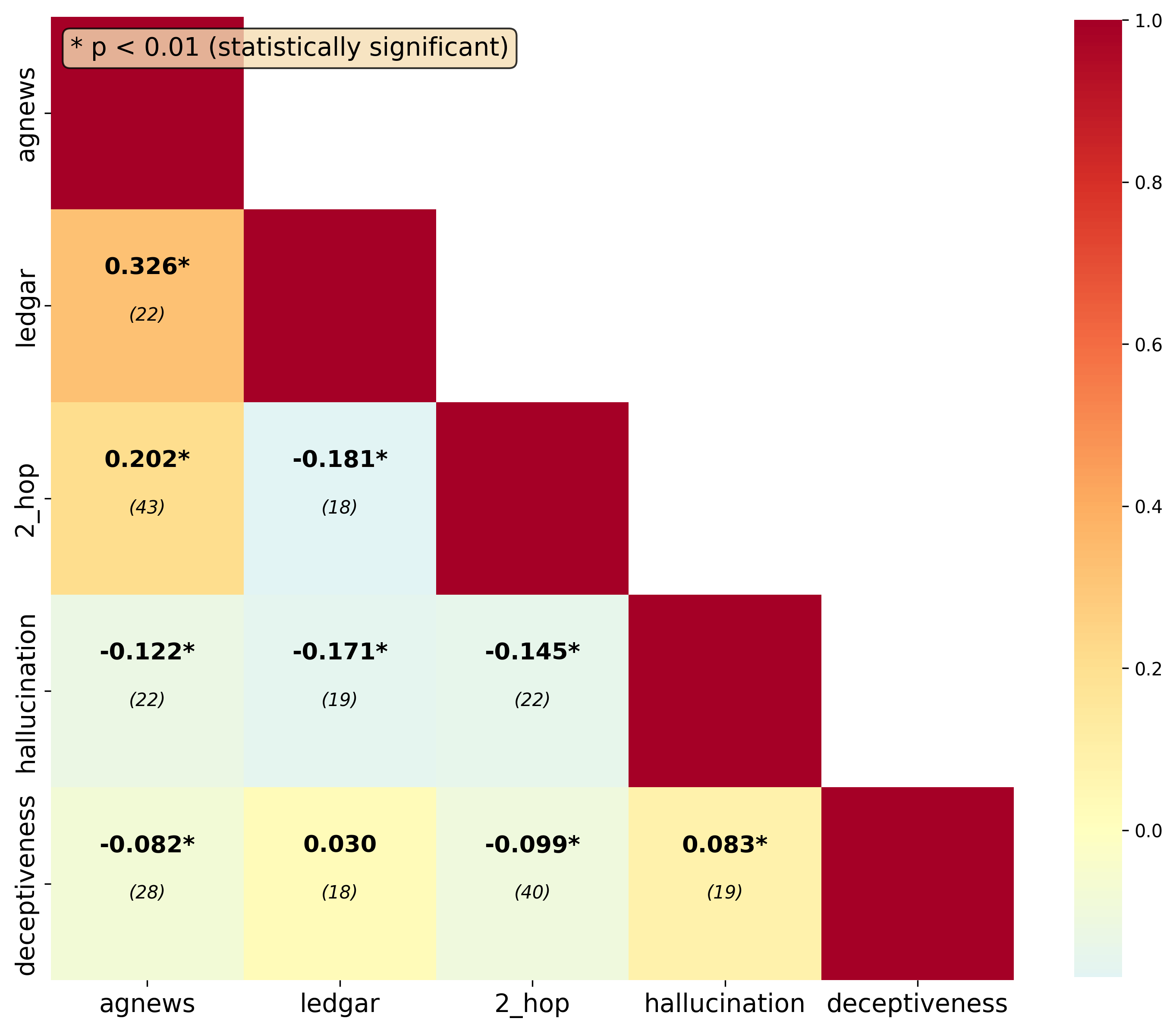}
        \caption{Mean operator}
        \label{fig:chart1}
    \end{subfigure}
    \hfill
    \begin{subfigure}{0.45\textwidth}
        \centering
        \includegraphics[width=\linewidth]{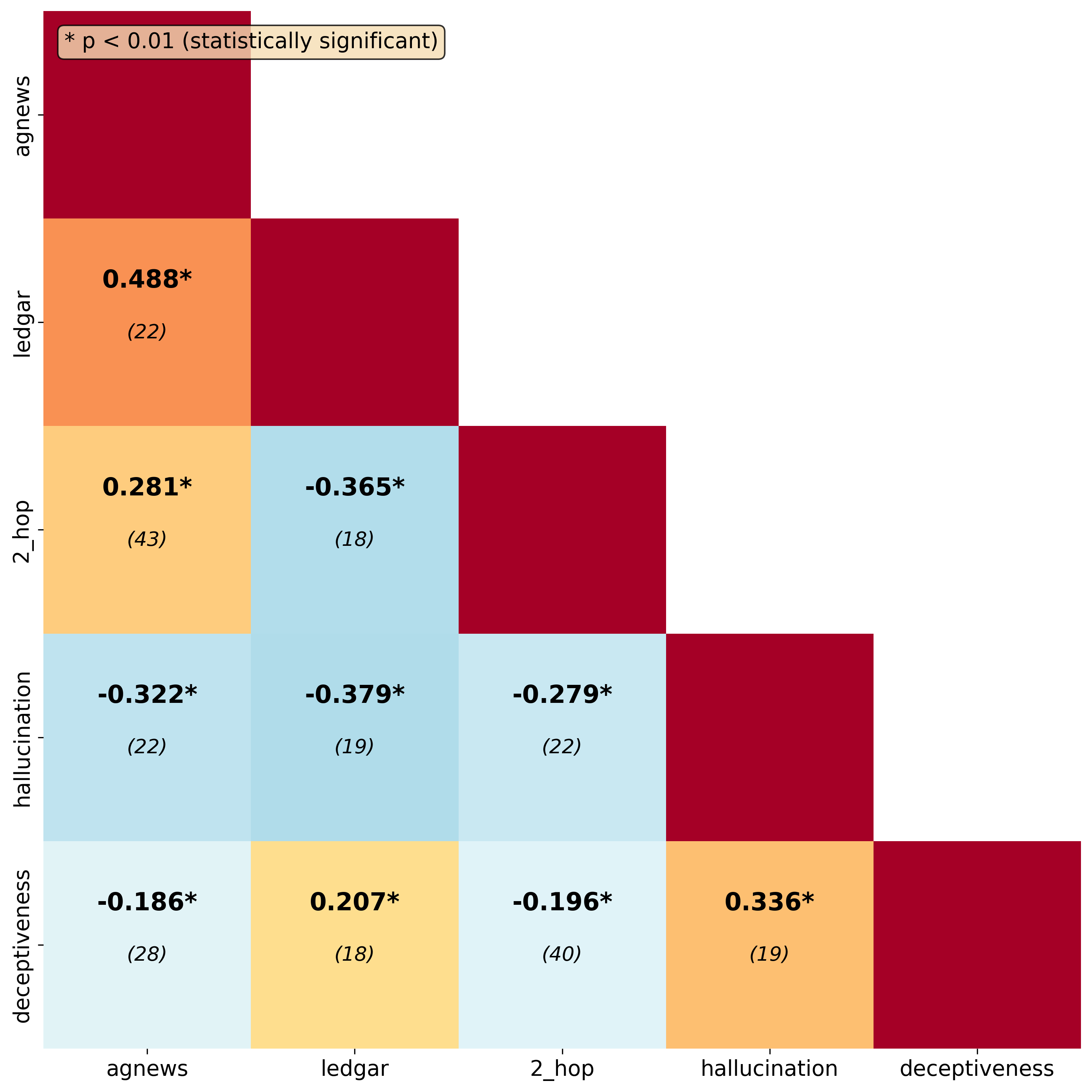}
        \caption{Max operator}
        \label{fig:chart2}
    \end{subfigure}
    \caption{Linear vectors cosine similarity analysis on \texttt{gemma-2-27b}}
    \label{fig:cosine_similatiy_27b}
\end{figure}

Here we provide additional information about layer-wise linear probe performance and similarity between faithfulness vectors and with other AI safety linear vectors. Figures~\ref{fig:2hop_probe_2b}-\ref{fig:ledgar_probe_27b} show the layer-wise performance of the faithfulness linear probes for 2-hop reasoning, AGNews and Ledgar.  As shown in Figure~\ref{fig:cosine_similatiy_2b}, \ref{fig:cosine_similatiy_9b} and~\ref{fig:cosine_similatiy_27b}, cosine similarity between task-specific linear faithfulness and AI safety behaviors vectors becomes more pronounced with the size of the model. 

Table~\ref{tab:transfer_faithfulness} shows the F1 score for faithfulness linear detection by using the linear representation from one task (e.g. 2-hop reasoning) to predict self-NLE faithfulness from other tasks (e.g. Ledgar and AGNews). These results corroborate the cosine similarity analysis, highlighting that the 2-hop reasoning/AGNews and AGNews/Ledgar pairs are moderately correlated with \texttt{gemma-2-27b}, enabling to properly detect faithfulness (approximately 62\%). This phenomenon does not appear on smaller models.  

\begin{table}[H]
\centering
\small
\begin{tabular*}{\textwidth}{@{\extracolsep{\fill}}lcccccc@{}}
\toprule
 & \multicolumn{2}{c}{\texttt{gemma-2b}} & \multicolumn{2}{c}{\texttt{gemma-9b}} & \multicolumn{2}{c}{\texttt{gemma-27b}} \\
\cmidrule(lr){2-3} \cmidrule(lr){4-5} \cmidrule(lr){6-7}
\textbf{Base/Target} & \textbf{Agnews} & \textbf{Ledgar} & \textbf{Agnews} & \textbf{Ledgar} & \textbf{Agnews} & \textbf{Ledgar} \\
\midrule
2-hop reasoning & 52.2\% & 56.1\% & 47.2\% & 40.0\% & 61.3\% & 36.1\% \\
AGNews & -- & 45.3\% & -- & 42.2\% & -- & 63.1\% \\
\bottomrule
\end{tabular*}
\caption{Faithfulness detection by transfer, from a base to a target faithfulness linear vector.}
\label{tab:transfer_faithfulness}
\end{table}

\subsection{Detailed Taxonomy of Self-NLE in Two-hop Reasoning}
\label{sec::caract}

\paragraph{Taxonomy Definition.}
The correctness of the prediction combined with both the faithfulness and the correctness of the self-NLE and the correctness of the first hop of the latent reasoning enables to precisely characterize $e(x)$. In this subsection, we focus on ten disjoint cases of interest to characterize the behavior of $f$ with respect to $e(x)$. They are illustrated in Figure~\ref{fig:taxonomy}. Given a ground truth 2-hop reasoning trace $(o_{1},r_1,o_{2},r_2,o_{3})$ and the actual reasoning trace obtained from both the model answer and self-NLE: $(o_{1},r_1, \widehat{o_{2}},r_2,\widehat{o_{3}})$:

\begin{figure}[H]{\centering}
\begin{center}
\includegraphics[scale=0.48]{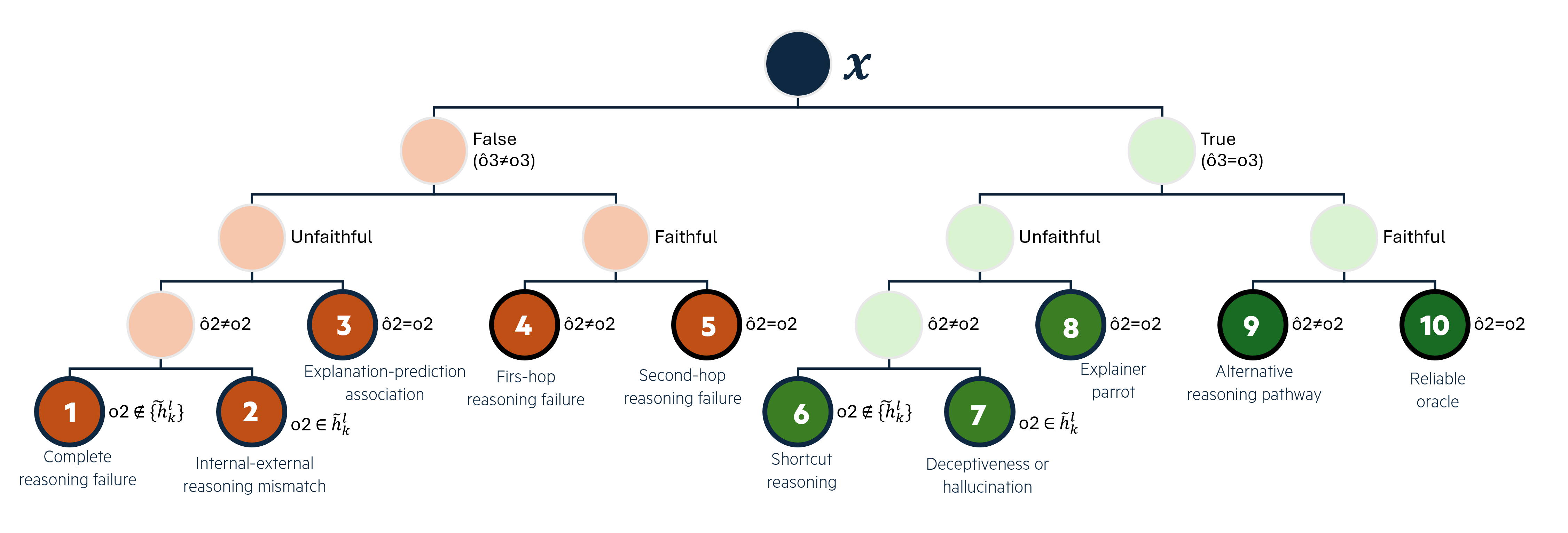}
\caption{Detailed taxonomy of $f$ behavior in two-hop reasoning, based on the status of the prediction, the self-NLE and the latent reasoning.}  
\label{fig:taxonomy}
\end{center}
\end{figure}

\begin{itemize}[align=left, leftmargin=*]
\item \textbf{($C_{1}$) Complete reasoning failure.} $\widehat{o_{3}}\neq o_3$, $F(x,e)=0$, $\widehat{o_{2}} \neq o_{2}$ and $\forall (k,\ell) \in \Gamma, o_{2} \notin \tilde{h^\ell_k}$. 
\textit{Observation:} Wrong prediction, incorrect unfaithful explanation, ground-truth bridge object undetected in circuit $\Gamma$. 
\textit{Interpretation:} Evidence suggests failure in first-hop reasoning, with neither explanation nor internal representations containing the expected bridge object.
    
    \item \textbf{($C_{2}$) Internal-external reasoning mismatch.} $\widehat{o_{3}} \neq o_3$, $F(x,e)=0$, $\widehat{o_{2}} \neq o_{2}$ and $\exists (k,\ell) \in \Gamma, o_{2} \in \tilde{h^\ell_k}$. 
\textit{Observation:} Wrong prediction, incorrect unfaithful explanation, but ground-truth bridge object detected in circuit $\Gamma$. 
\textit{Interpretation:} The model appears to have correct internal knowledge but generates inconsistent explanations, indicating potential reasoning-explanation dissociation because of either deceptiveness or hallucination.
    \item \textbf{($C_{3}$) Explanation-prediction association.} $\widehat{o_{3}} \neq o_3$, $F(x,e)=0$ and $\widehat{o_{2}} = o_{2}$. 
\textit{Observation:} Wrong prediction with unfaithful but correct explanation. 
\textit{Interpretation:} The model associates the correct bridge object with incorrect prediction during explanation generation, suggesting superficial pattern matching without genuinely resolving the first hop of the 2-hop reasoning.
    \item \textbf{($C_{4}$) First-hop reasoning failure.} $\widehat{o_{3}} \neq o_3$, $F(x,e)=1$ and $\widehat{o_{2}} \neq o_{2}$. 
\textit{Observation:} Wrong prediction with faithful but incorrect explanation. 
\textit{Interpretation:} The model consistently follows an incorrect reasoning pathway, indicating error in first-hop reasoning or concept misunderstanding.
    \item \textbf{($C_{5}$) Second-hop reasoning failure.} $\widehat{o_{3}} \neq o_3$, $F(x,e)=1$ and $\widehat{o_{2}} = o_{2}$. 
\textit{Observation:} Wrong prediction with faithful and correct explanation. 
\textit{Interpretation:} The model correctly identifies he bridge object but fails in second reasoning step, suggesting error occurs after successful first-hop completion.
    \item \textbf{($C_{6}$) Shortcut learning.} $\widehat{o_{3}}=o_3$, $F(x,e)=0$, $\widehat{o_{2}} \neq o_{2}$ and $\forall (k,\ell) \in \Gamma, o_{2} \notin \tilde{h^\ell_k}$. 
\textit{Observation:} Correct prediction with unfaithful incorrect explanation, ground-truth bridge object undetected. 
\textit{Interpretation:} Evidence suggests direct $o_1 \rightarrow o_3$ association, consistent with shortcut learning behavior that bypasses intermediate reasoning steps.
    \item \textbf{($C_{7}$) Deceptiveness or hallucination.} $\widehat{o_{3}}=o_3$, $F(x,e)=0$, $\widehat{o_{2}} \neq o_{2}$ and $\exists (k,\ell) \in \Gamma, o_{2} \in \tilde{h^\ell_k}$. 
\textit{Observation:} Correct prediction with unfaithful incorrect explanation, but ground-truth bridge object detected internally. 
\textit{Interpretation:} The model possesses correct internal knowledge but generates deceptive (or hallucinated) explanations, suggesting reasoning-explanation dissociation or alternative reasoning pathways. This case is expected to be rare, otherwise highlighting a case where $f$ is not honest in its self-NLE while "knowing" the ground truth bridge object, raising a problem in $f$ alignment.
    \item \textbf{($C_{8}$) Explainer parrot.} $\widehat{o_{3}}=o_3$, $F(x,e)=0$ and $\widehat{o_{2}} = o_{2}$. 
\textit{Observation:} Correct prediction with unfaithful but correct explanation. 
\textit{Interpretation:} The model generates the expected explanation without corresponding detectable internal reasoning, suggesting post-hoc explanation generation (i.e. "explainer parrot" behavior).
    \item \textbf{($C_{9}$) Alternative reasoning pathway.} $\widehat{o_{3}}=o_3$, $F(x,e)=1$ and $\widehat{o_{2}} \neq o_{2}$. 
\textit{Observation:} Correct prediction with faithful but incorrect explanation. 
\textit{Interpretation:} The model uses a consistent but non-canonical reasoning pathways, indicating bias or alternative reasoning mechanism that leads to correct outcomes through unexpected intermediate steps.
    \item \textbf{($C_{10}$) Reliable oracle.} $\widehat{o_{3}}=o_3$, $F(x,e)=1$ and $\widehat{o_{2}} = o_{2}$. 
\textit{Observation:} Correct prediction with faithful and correct explanation. 
\textit{Interpretation:} Strong evidence for expected canonical reasoning pathway, representing the most interpretable and reliable case for knowledge extraction and model understanding.

\end{itemize}

While this taxonomy relies on the interpreter ability to decode the model's internal activity, the systematic patterns observed across different models and tasks provide convergent evidence for these behavioral categories. We give examples of this taxonomy in Appendix~\ref{sec::examples}


\begin{table}[t]
\centering
\small
\begin{tabular}{l*{5}{c}}
\toprule
\multicolumn{6}{c}{\textbf{Incorrect Predictions}} \\
\midrule
\textbf{Model} & \textbf{C1} & \textbf{C2} & \textbf{C3} & \textbf{C4} & \textbf{C5} \\
& \scriptsize{Complete} & \scriptsize{Internal-external} & \scriptsize{Explanation-} & \scriptsize{First-hop} & \scriptsize{Second-hop} \\
& \scriptsize{reasoning failure} & \scriptsize{reasoning mismatch} & \scriptsize{prediction assoc.} & \scriptsize{reasoning failure} & \scriptsize{reasoning failure} \\
\midrule
\textbf{\texttt{gemma-2-2b}} & 23.8\% & 4.2\% & 14.4\% & 16.0\% & 41.6\% \\
\textbf{\texttt{gemma-2-9b}} & 26.1\% & 4.5\% & 14.5\% & 14.1\% & 40.9\% \\
\textbf{\texttt{gemma-2-27b}} & 22.8\% & 4.4\% & 12.5\% & 17.1\% & 43.1\% \\
\textbf{\texttt{mistral-3-7b}} & 46.5\% & 13.7\% & 7.1\% & 8.4\% & 24.2\% \\
\bottomrule
\end{tabular}
\caption{Distribution of categories for incorrect predictions across model sizes.}
\label{tab:incorrect_predictions}
\end{table}


\begin{table}[t]
\centering
\small
\begin{tabular}{l*{5}{c}}
\toprule
\multicolumn{6}{c}{\textbf{Correct Predictions}} \\
\midrule
\textbf{Model} & \textbf{C6} & \textbf{C7} & \textbf{C8} & \textbf{C9} & \textbf{C10} \\
& \scriptsize{Shortcut} & \scriptsize{Deceptiveness or} & \scriptsize{Explainer} & \scriptsize{Alternative} & \scriptsize{Reliable} \\
& \scriptsize{learning} & \scriptsize{hallucination} & \scriptsize{parrot} & \scriptsize{reasoning pathway} & \scriptsize{oracle} \\
\midrule
\textbf{\texttt{gemma-2-2b}} & 27.8\% & 6.1\% & 17.7\% & 8.1\% & 40.3\% \\
\textbf{\texttt{gemma-2-9b}} & 14.9\% & 4.2\% & 20.0\% & 6.8\% & 54.0\% \\
\textbf{\texttt{gemma-2-27b}} & 12.8\% & 3.1\% & 15.3\% & 6.4\% & 62.4\% \\
\textbf{\texttt{mistral-3-7b}} & 20.6\% & 4.0\% & 19.3\% & 5.2\% & 50.9\% \\
\bottomrule
\end{tabular}
\caption{Distribution of categories for correct predictions across model sizes.}
\label{tab:categ_distrib_correct_predictions}
\end{table}

\subsubsection{2-hop reasoning detailed results.} 
\label{sec::appendix_2_hop_detailed_results}

Here we detail the results outlined in Section~\ref{sec:2_hop} and~\ref{sec:linear_detection} by breaking down the results at the category-level as introduced above. 

\paragraph{Taxonomy Descriptive Analysis.}
Table~\ref{tab:incorrect_predictions} highlights that the categories C1 (complete reasoning failure) and C5 (second-hop reasoning failure) are the most represented across the models. The distributions are overall highly stable for incorrect predictions whereas the category C10 (reliable oracle) increases with the model size for accurate predictions. The category C7 (deceptiveness or hallucination) tends to decrease with model size, but still represents a non negligible part of the self-NLE. 

\paragraph{Taxonomy Faithfulness Enhancement Analysis.}
Table~\ref{tab:categ_distib_steering_faithfulness},\ref{tab:categ_distib_steering_faithfulness_2} and~\ref{tab:categ_distib_steering_faithfulness_3}  respectively show the detailed impact of hallucination inhibition, linear faithfulness amplification and deceptiveness inhibition steering on self-NLE faithfulness. Categories C2 and C7 are the most prone to be turned into faithful self-NLE overall. Hallucination and faithfulness steering lead to significantly better results as compared to deceptiveness steering overall. Linear faithfulness amplification gives slightly better results than hallucination inhibition. Hallucination inhibition obtains slightly better results than linear faithfulness amplification for C1 and C6.

\begin{table}[H]
\centering
\small
\begin{tabular}{l*{6}{c}}
\toprule
\multicolumn{7}{c}{\textbf{New Faithfulness After Steering Through Hallucination Inhibition (\%)}} \\
\midrule
\textbf{Model} & \textbf{C1} & \textbf{C2} & \textbf{C3} & \textbf{C6} & \textbf{C7} & \textbf{C8} \\
& \scriptsize{Complete} & \scriptsize{Internal-external} & \scriptsize{Explanation-} & \scriptsize{Shortcut} & \scriptsize{Deceptiveness or} & \scriptsize{Explainer} \\
& \scriptsize{reasoning failure} & \scriptsize{reasoning mismatch} & \scriptsize{prediction assoc.} & \scriptsize{learning} & \scriptsize{hallucination} & \scriptsize{parrot} \\
\midrule
\textbf{\texttt{gemma-2-2b}} & 10.8\% & \textbf{33.3}\% & 5.6\% & 7.4\% & \textbf{36.2}\% & 3.0\% \\
\textbf{\texttt{gemma-2-9b}} & 9.2\% & \textbf{13.4}\% & 5.1\% & 16.1\% & \textbf{35.9}\% & 1.6\% \\
\textbf{\texttt{gemma-2-27b}} & 12.0\% & \textbf{22.7}\% & 4.8\% & 12.5\% & \textbf{48.9}\% & 2.6\% \\
\bottomrule
\end{tabular}
\caption{Percentage of initially unfaithful explanations that become faithful after hallucination steering interventions, by category and model size.}
\label{tab:categ_distib_steering_faithfulness}
\end{table}

\begin{table}[H]
\centering
\small
\begin{tabular}{l*{6}{c}}
\toprule
\multicolumn{7}{c}{\textbf{New Faithfulness After Steering Through Linear Faithfulness Amplification (\%)}} \\
\midrule
\textbf{Model} & \textbf{C1} & \textbf{C2} & \textbf{C3} & \textbf{C6} & \textbf{C7} & \textbf{C8} \\
& \scriptsize{Complete} & \scriptsize{Internal-external} & \scriptsize{Explanation-} & \scriptsize{Shortcut} & \scriptsize{Deceptiveness or} & \scriptsize{Explainer} \\
& \scriptsize{reasoning failure} & \scriptsize{reasoning mismatch} & \scriptsize{prediction assoc.} & \scriptsize{learning} & \scriptsize{hallucination} & \scriptsize{parrot} \\
\midrule
\textbf{\texttt{gemma-2-2b}} & 8.1\% & \textbf{38.0}\% & 5.1\% & 9.3\% & \textbf{39.5}\% & 1.1\% \\
\textbf{\texttt{gemma-2-9b}} & 6.9\% & \textbf{28.4}\% & 3.7\% & 10.3\% & \textbf{57.8}\% & 2.0\% \\
\textbf{\texttt{gemma-2-27b}} & 9.3\% & \textbf{19.7}\% & 5.3\% & 10.9\% & \textbf{40.4}\% & 1.3\% \\
\bottomrule
\end{tabular}
\caption{Percentage of initially unfaithful explanations that become faithful after linear faithfulness steering interventions, by category and model size.}
\label{tab:categ_distib_steering_faithfulness_2}
\end{table}

\begin{table}[H]
\centering
\small
\begin{tabular}{l*{6}{c}}
\toprule
\multicolumn{7}{c}{\textbf{New Faithfulness After Steering Through Deceptiveness Inhibition (\%)}} \\
\midrule
\textbf{Model} & \textbf{C1} & \textbf{C2} & \textbf{C3} & \textbf{C6} & \textbf{C7} & \textbf{C8} \\
& \scriptsize{Complete} & \scriptsize{Internal-external} & \scriptsize{Explanation-} & \scriptsize{Shortcut} & \scriptsize{Deceptiveness or} & \scriptsize{Explainer} \\
& \scriptsize{reasoning failure} & \scriptsize{reasoning mismatch} & \scriptsize{prediction assoc.} & \scriptsize{learning} & \scriptsize{hallucination} & \scriptsize{parrot} \\
\midrule
\textbf{\texttt{gemma-2-2b}} & 5.1\% & 3.2\% & 2.7\% & 4.7\% & \textbf{13.8}\% & 6.0\% \\
\textbf{\texttt{gemma-2-9b}} & 6.1\% & \textbf{11.7}\% & 0.6\% & 8.0\% & \textbf{18.7}\% & 1.4\% \\
\textbf{\texttt{gemma-2-27b}} & 5.0\% & 6.3\% & 1.7\% & 2.7\% & \textbf{9.1}\% & 3.3\% \\
\bottomrule
\end{tabular}
\caption{Percentage of initially unfaithful explanations that become faithful after deceptiveness steering interventions, by category and model size (third experimental condition).}
\label{tab:categ_distib_steering_faithfulness_3}
\end{table}

\begin{figure}[H]
    \centering
    \begin{subfigure}{0.45\textwidth}
        \centering
        \includegraphics[width=\linewidth]{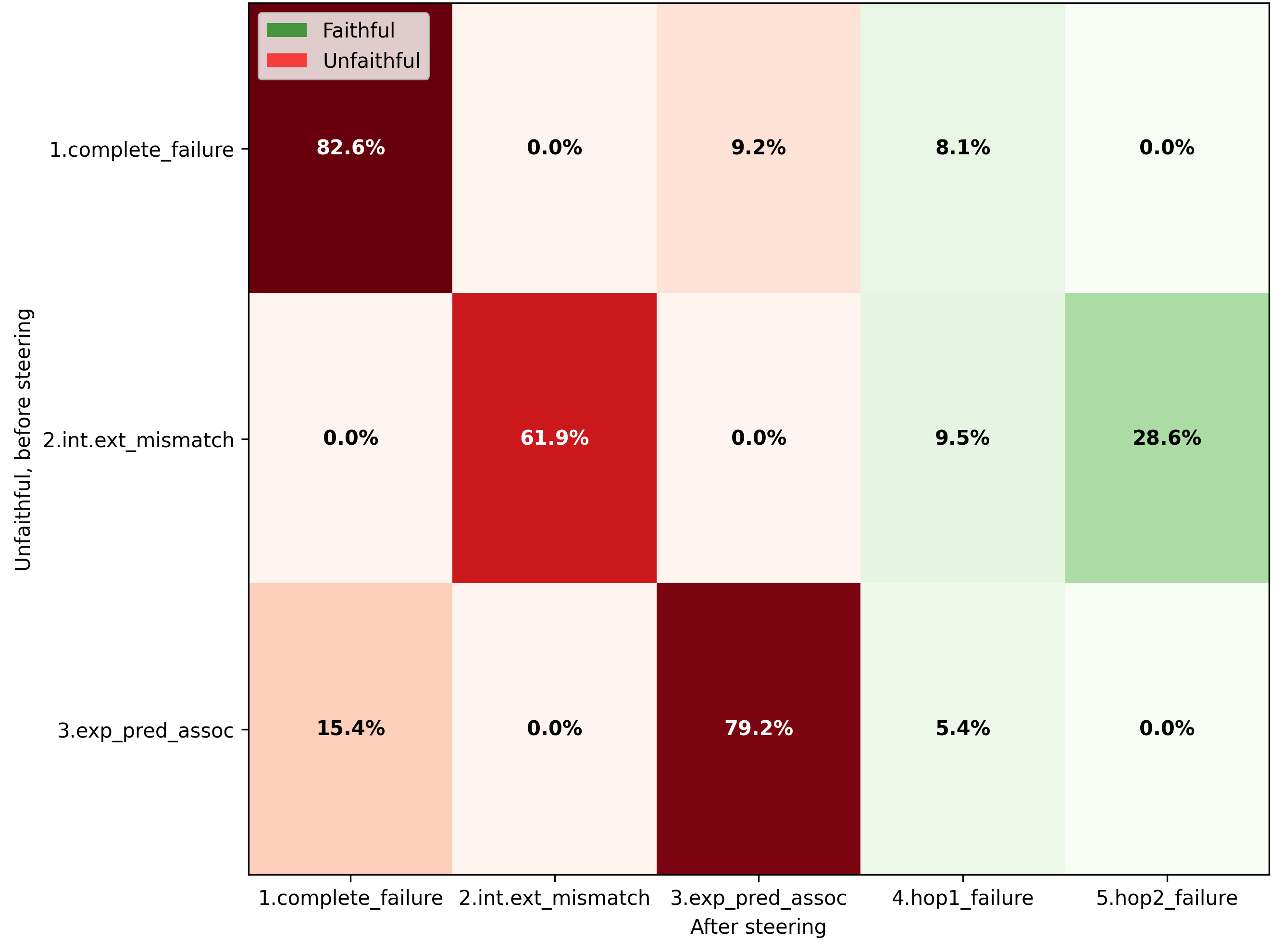}
        \caption{Initially inaccurate predictions.}
        \label{fig:chart1}
    \end{subfigure}
    \hfill
    \begin{subfigure}{0.45\textwidth}
        \centering
        \includegraphics[width=\linewidth]{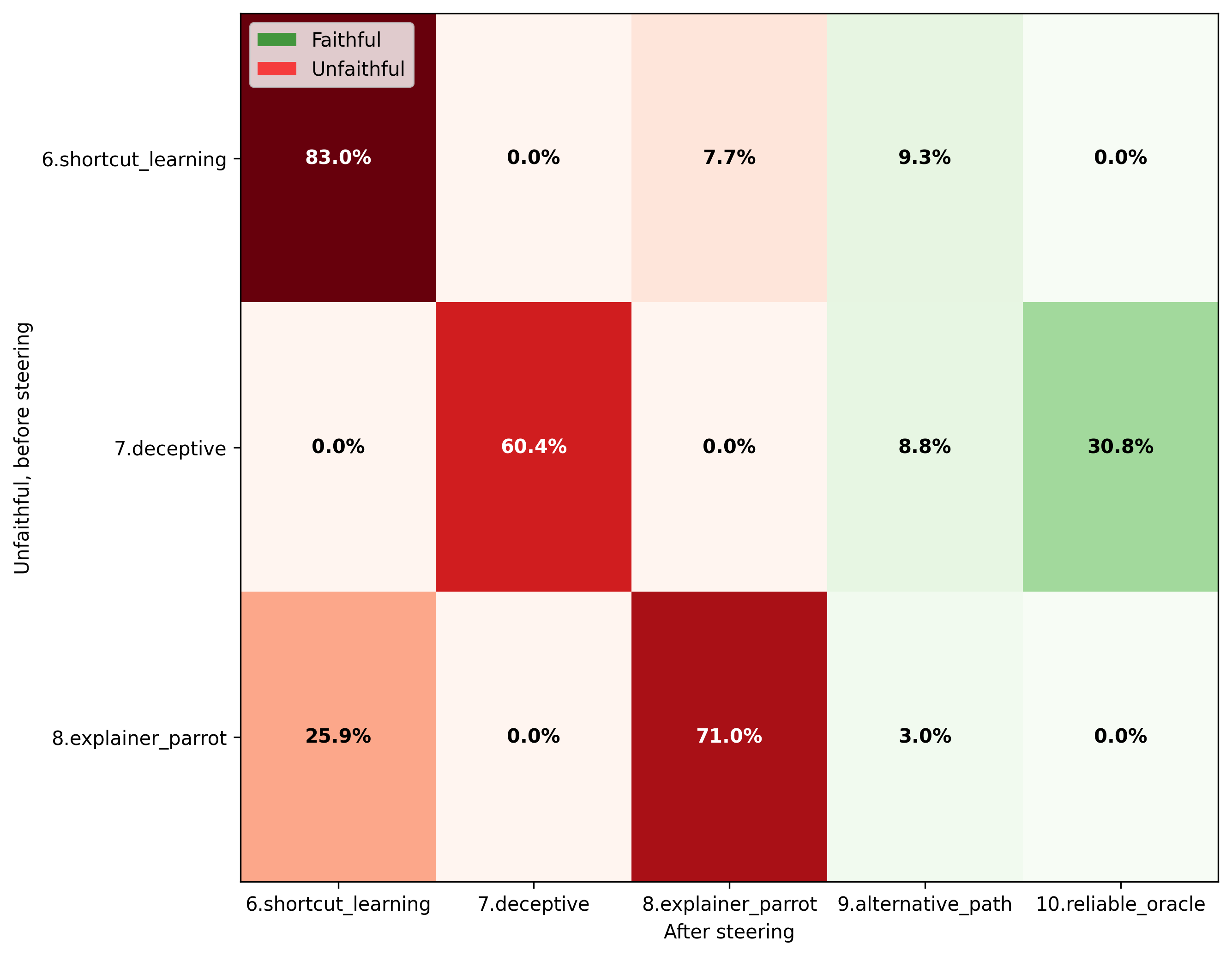}
        \caption{Initially accurate predictions.}
        \label{fig:chart2}
    \end{subfigure}
    \caption{Detailed taxonomy transition state analysis, before and after \method\ linear faithfulness steering on \texttt{gemma-2-2b} on 2-hop reasoning.}
    \label{fig:transition_2b}
\end{figure}

\begin{figure}[H]
    \centering
    \begin{subfigure}{0.45\textwidth}
        \centering
        \includegraphics[width=\linewidth]{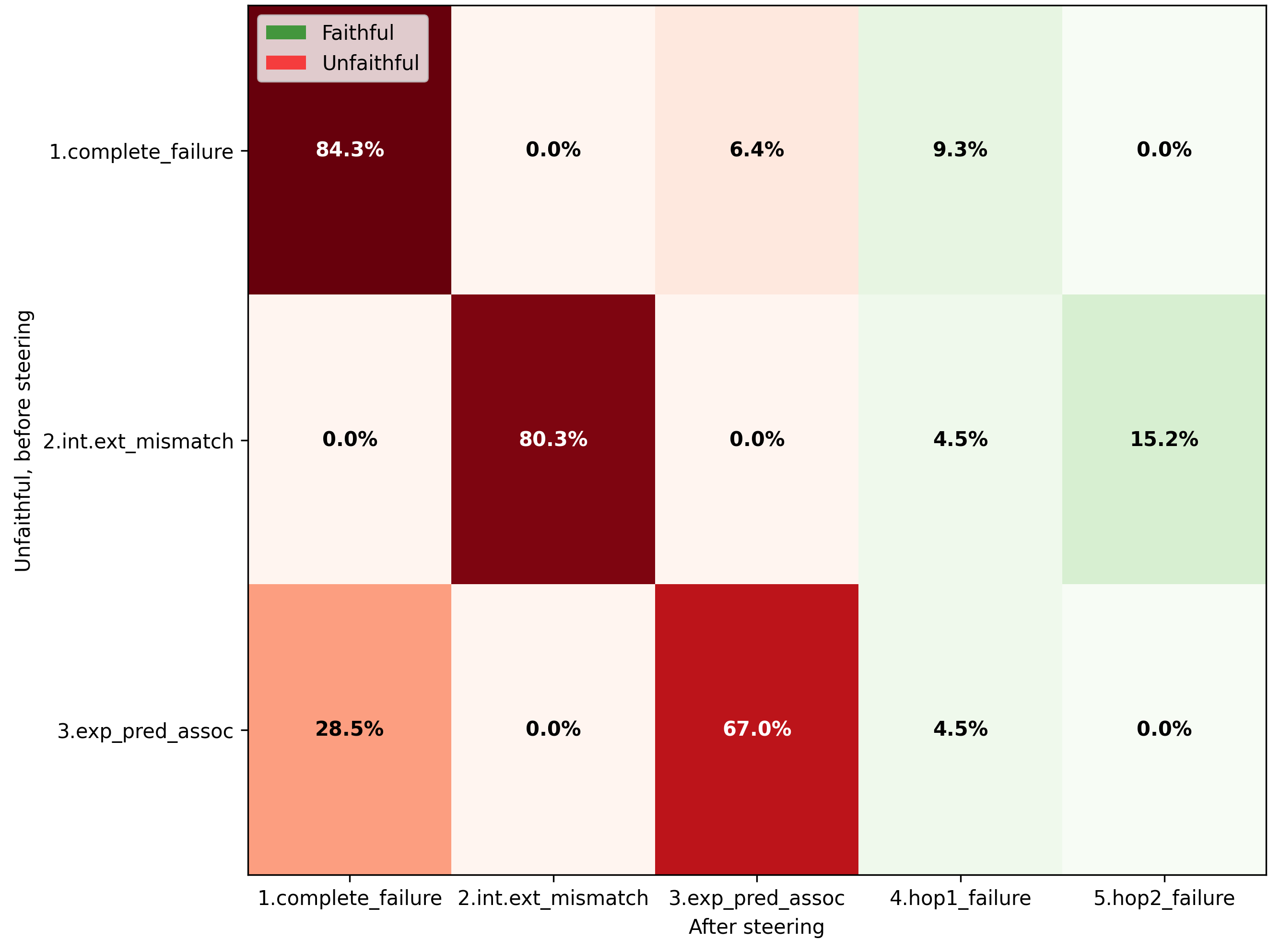}
        \caption{Initially inaccurate predictions.}
        \label{fig:chart1}
    \end{subfigure}
    \hfill
    \begin{subfigure}{0.45\textwidth}
        \centering
        \includegraphics[width=\linewidth]{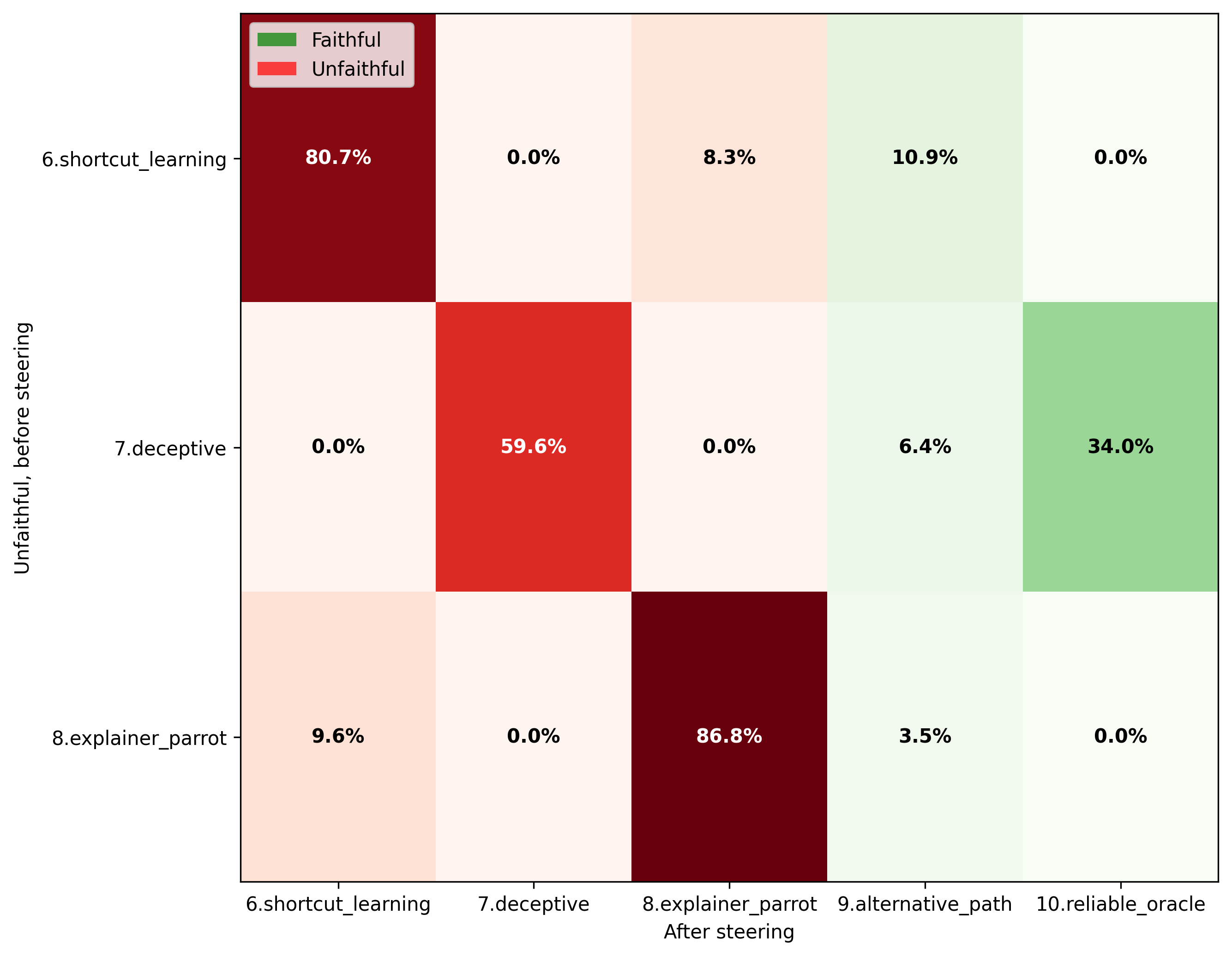}
        \caption{Initially accurate predictions.}
        \label{fig:chart2}
    \end{subfigure}
    \caption{Detailed taxonomy transition state analysis, before and after \method\ linear faithfulness steering on \texttt{gemma-2-9b} on 2-hop reasoning.}
    \label{fig:transition_9b}
\end{figure}

\begin{figure}[H]
    \centering
    \begin{subfigure}{0.45\textwidth}
        \centering
        \includegraphics[width=\linewidth]{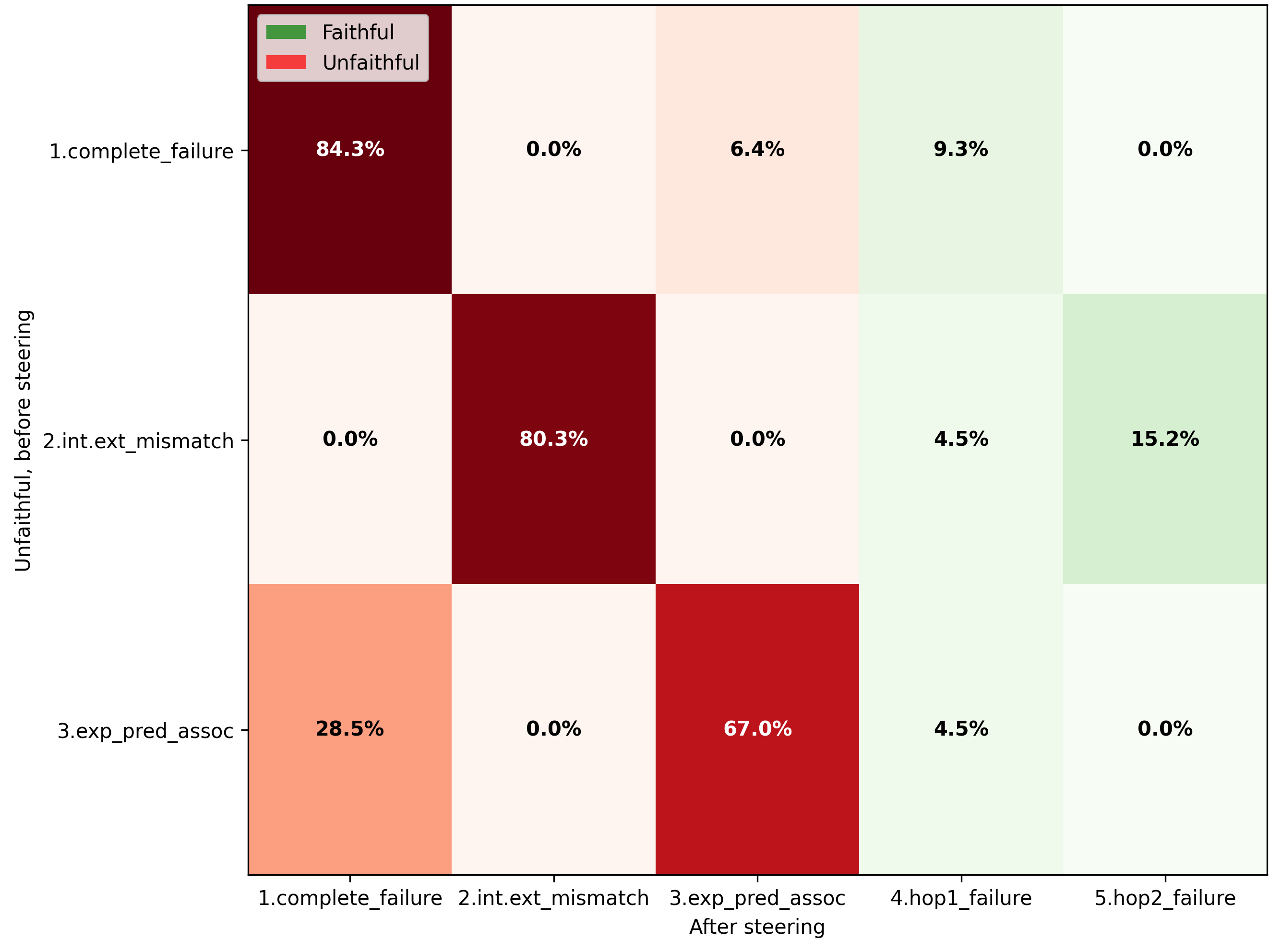}
        \caption{Initially inaccurate predictions.}
        \label{fig:chart1}
    \end{subfigure}
    \hfill
    \begin{subfigure}{0.45\textwidth}
        \centering
        \includegraphics[width=\linewidth]{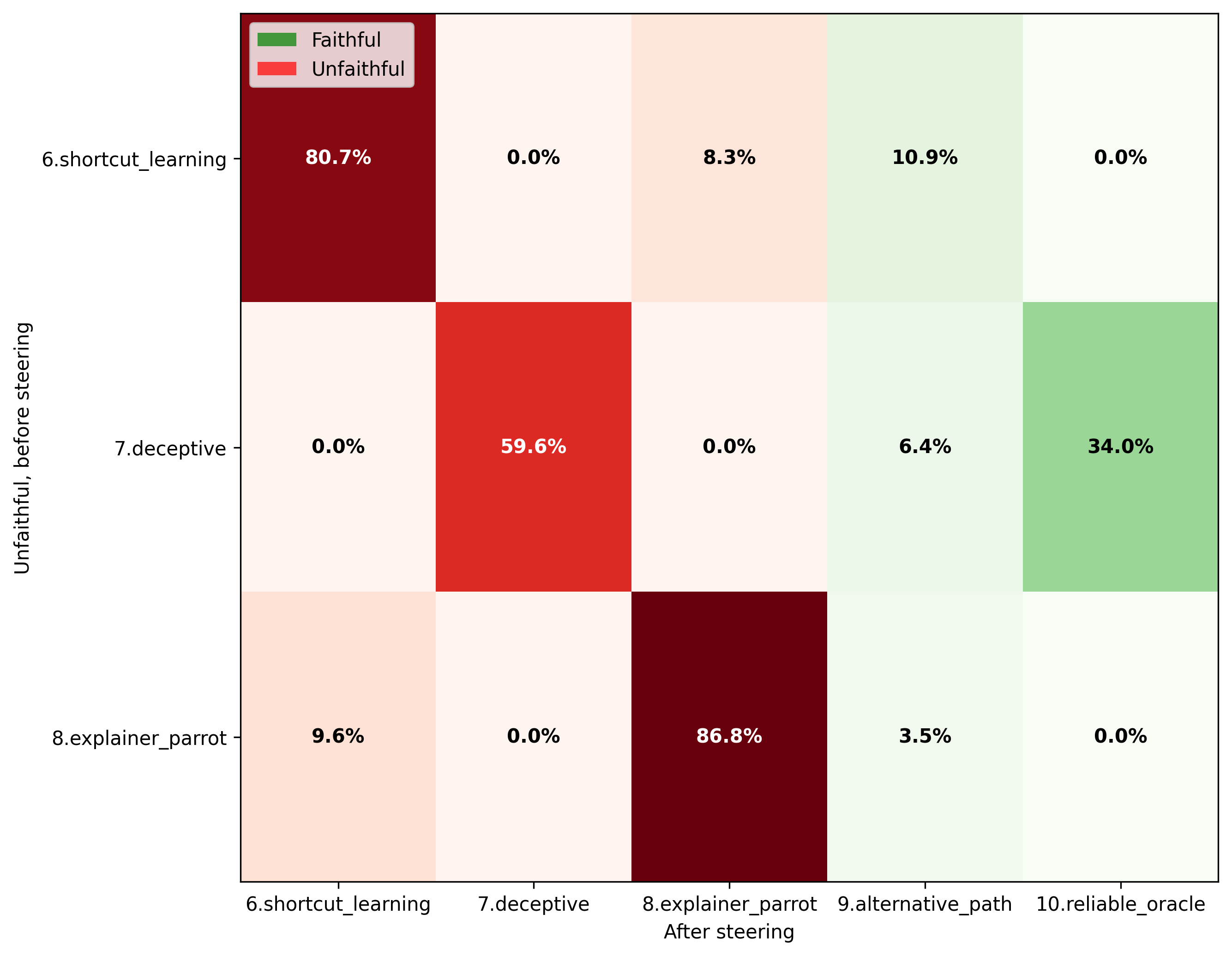}
        \caption{Initially accurate predictions.}
        \label{fig:chart2}
    \end{subfigure}
    \caption{Detailed taxonomy transition state analysis, before and after \method\ linear faithfulness steering on \texttt{gemma-2-27b} on 2-hop reasoning.}
    \label{fig:transition_27b}
\end{figure}

As shown in Figures~\ref{fig:transition_2b}, \ref{fig:transition_9b}, and~\ref{fig:transition_27b}, we observe consistent transition patterns when steering makes unfaithful explanations faithful. For incorrect predictions, category C1 (complete reasoning failure) consistently transitions to C4 (first-hop reasoning failure), achieving faithfulness approximately 10\% of the time under \method\ linear faithfulness amplification and hallucination inhibition. Category C2 (internal-external reasoning mismatch) predominantly transitions to C5 (second-hop reasoning failure) when steering succeeds. Category C3 (explanation-prediction association) exclusively leads to C5 (second-hop reasoning failure) upon becoming faithful.

Accurate predictions follow similar patterns. Category C6 (shortcut learning) transitions to C9 (alternative reasoning pathway) when made faithful, while category C8 (explainer parrot) becomes C10 (reliable oracle). Category C7 (deceptiveness or hallucination) can transition to either C9 or C10, though it more commonly becomes a reliable oracle (C10). We give several examples of unfaithful self-NLE made faithful in Appendix~\ref{sec::examples}.

\begin{figure}[H]
    \centering
    \begin{subfigure}{0.45\textwidth}
        \centering
        \includegraphics[width=\linewidth]{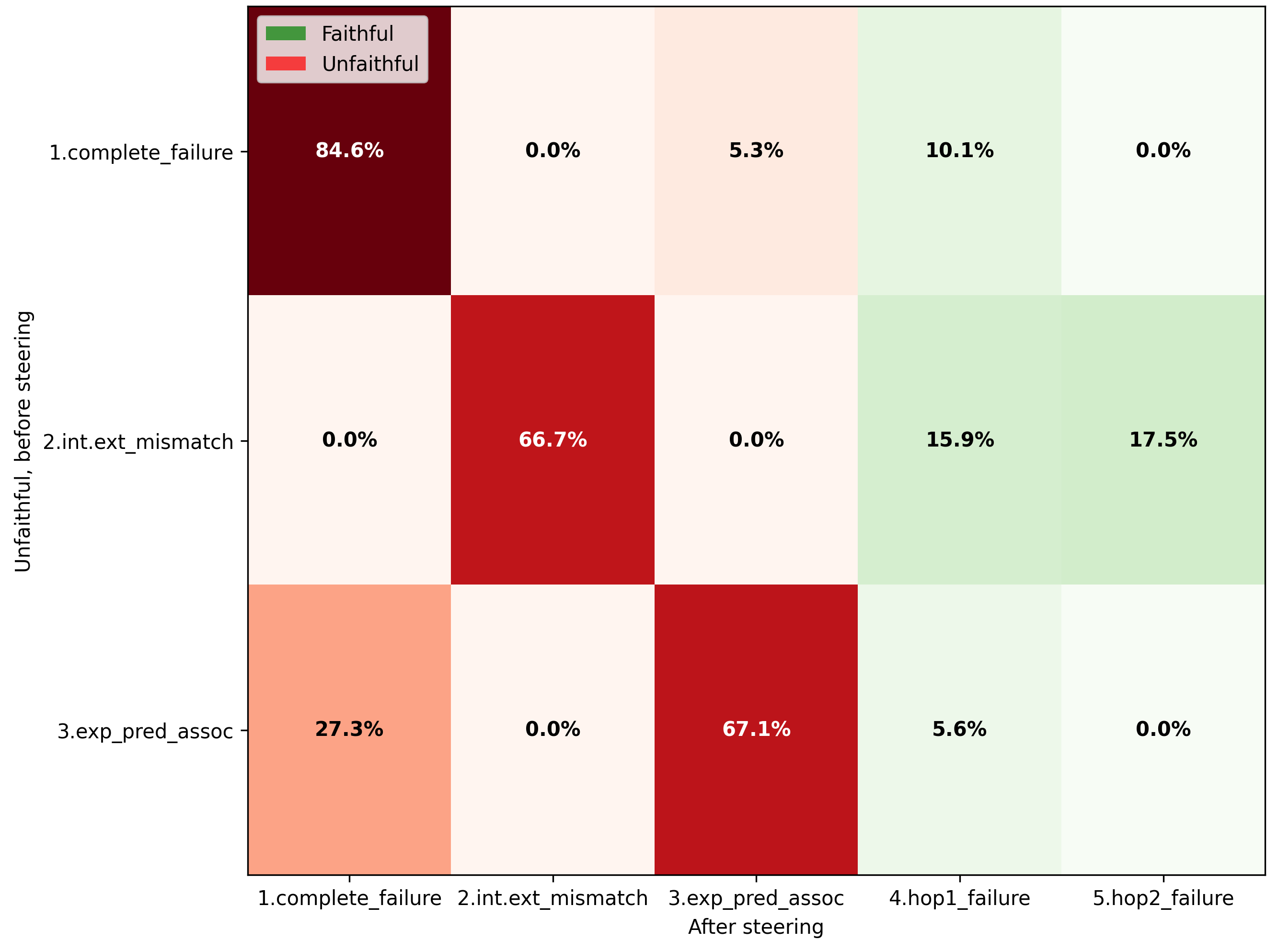}
        \caption{Initially inaccurate predictions.}
        \label{fig:chart1}
    \end{subfigure}
    \hfill
    \begin{subfigure}{0.45\textwidth}
        \centering
        \includegraphics[width=\linewidth]{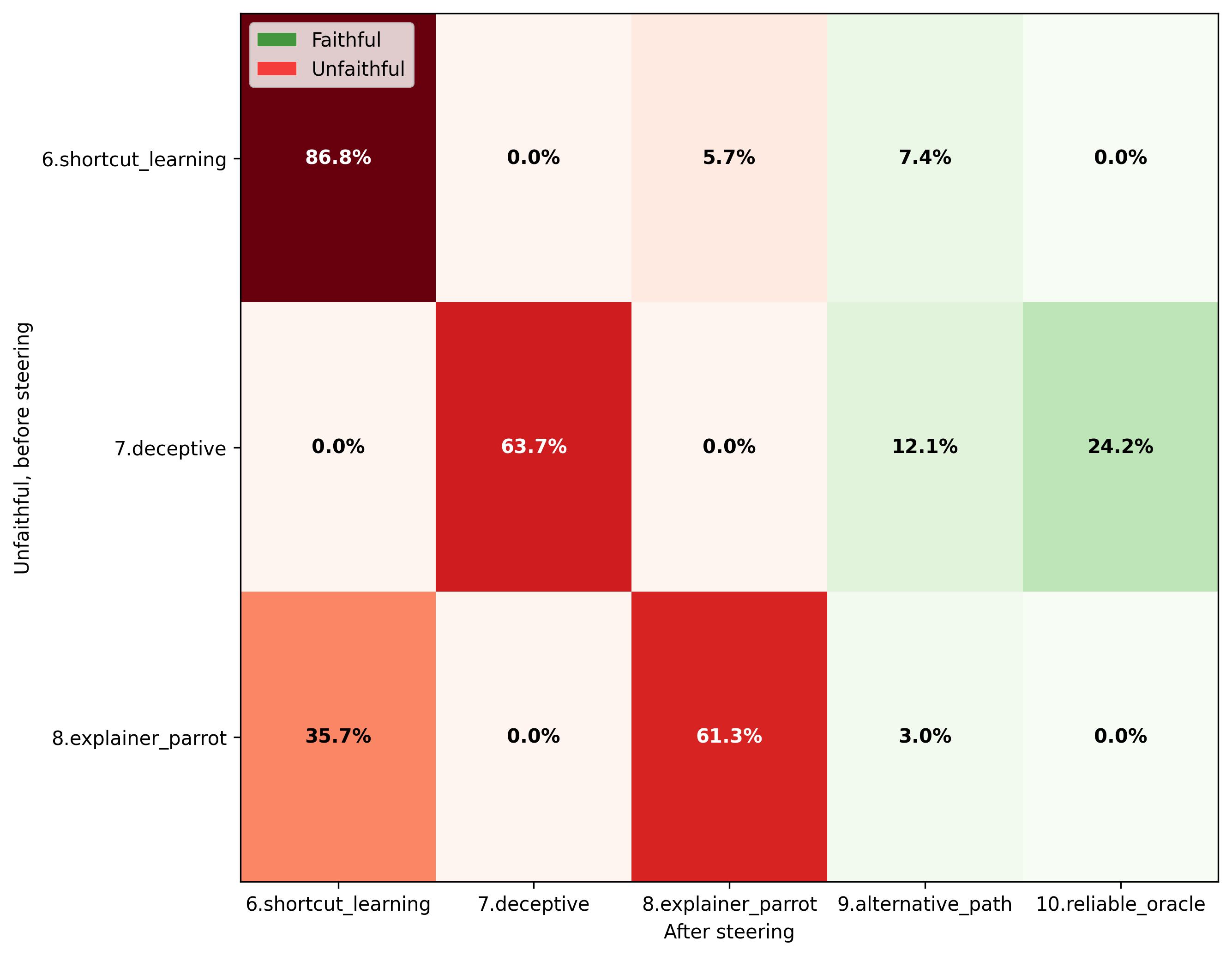}
        \caption{Initially accurate predictions.}
        \label{fig:chart2}
    \end{subfigure}
    \caption{Detailed taxonomy transition state analysis, before and after hallucination inhibition on \texttt{gemma-2-2b} on 2-hop reasoning.}
    \label{fig:transition_2b_hallucination}
\end{figure}

\begin{figure}[H]
    \centering
    \begin{subfigure}{0.45\textwidth}
        \centering
        \includegraphics[width=\linewidth]{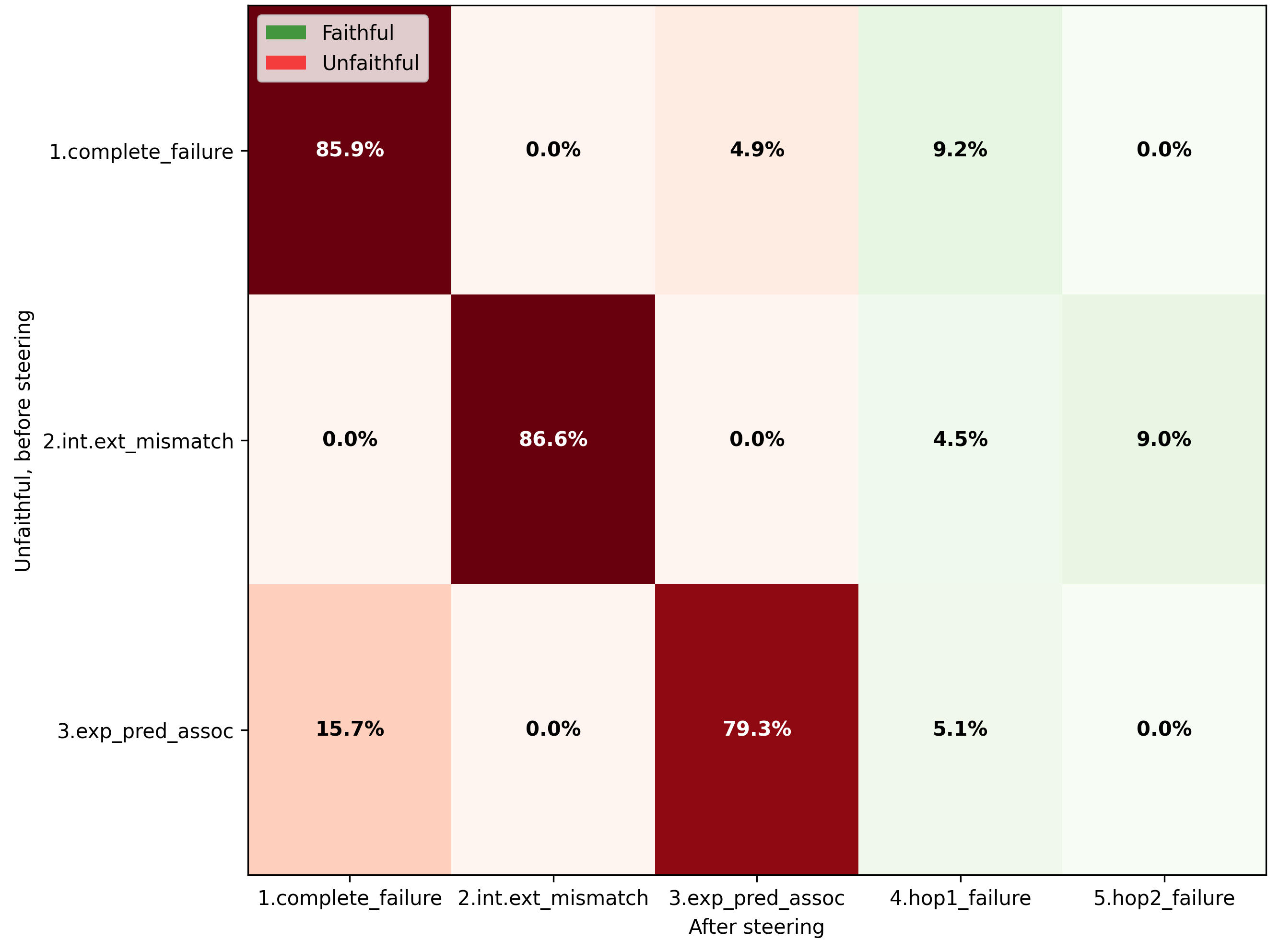}
        \caption{Initially inaccurate predictions.}
        \label{fig:chart1}
    \end{subfigure}
    \hfill
    \begin{subfigure}{0.45\textwidth}
        \centering
        \includegraphics[width=\linewidth]{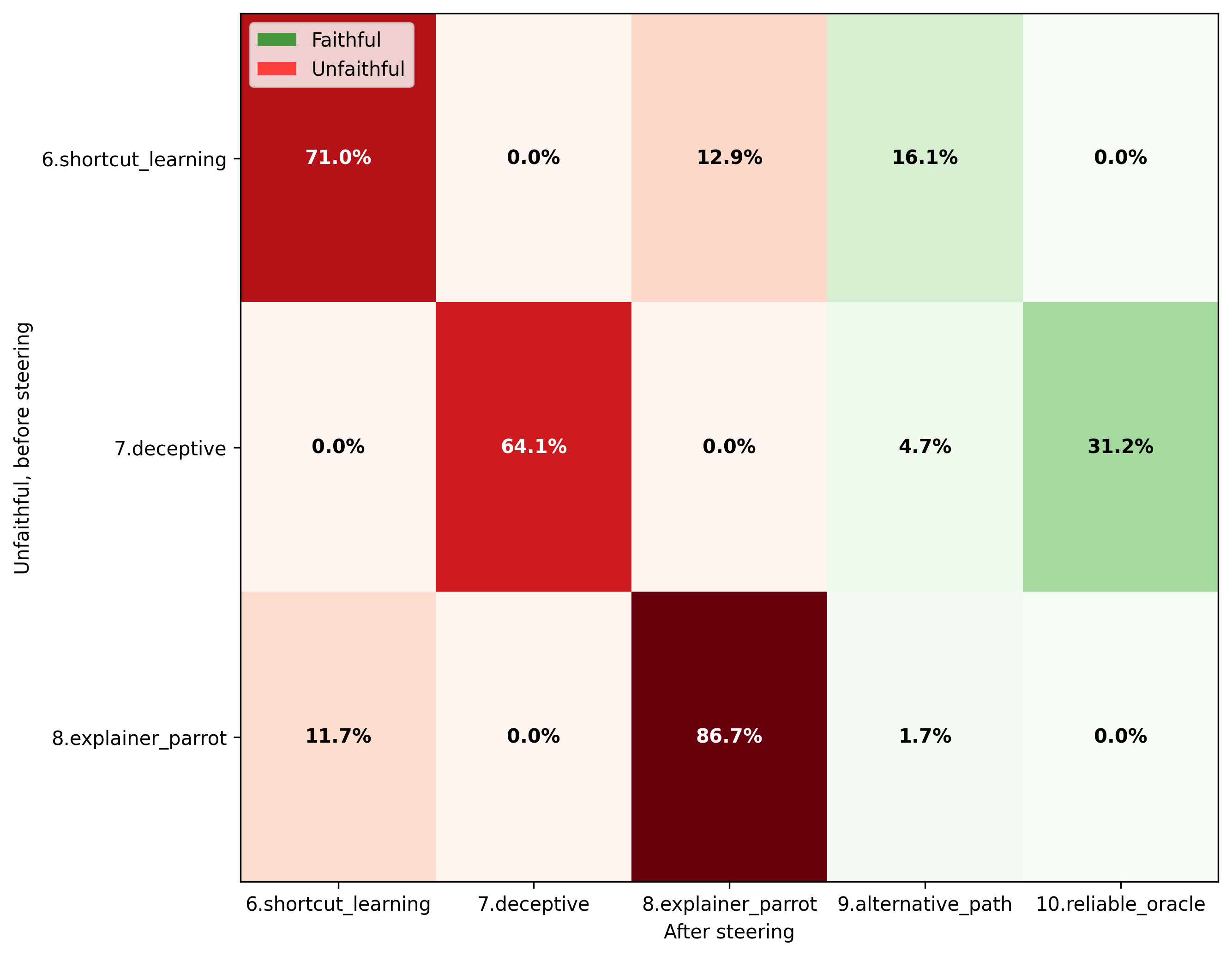}
        \caption{Initially accurate predictions.}
        \label{fig:chart2}
    \end{subfigure}
    \caption{Detailed taxonomy transition state analysis, before and after hallucination inhibition on \texttt{gemma-2-27b} on 2-hop reasoning.}
    \label{fig:transition_9b_hallucination}
\end{figure}

\begin{figure}[H]
    \centering
    \begin{subfigure}{0.45\textwidth}
        \centering
        \includegraphics[width=\linewidth]{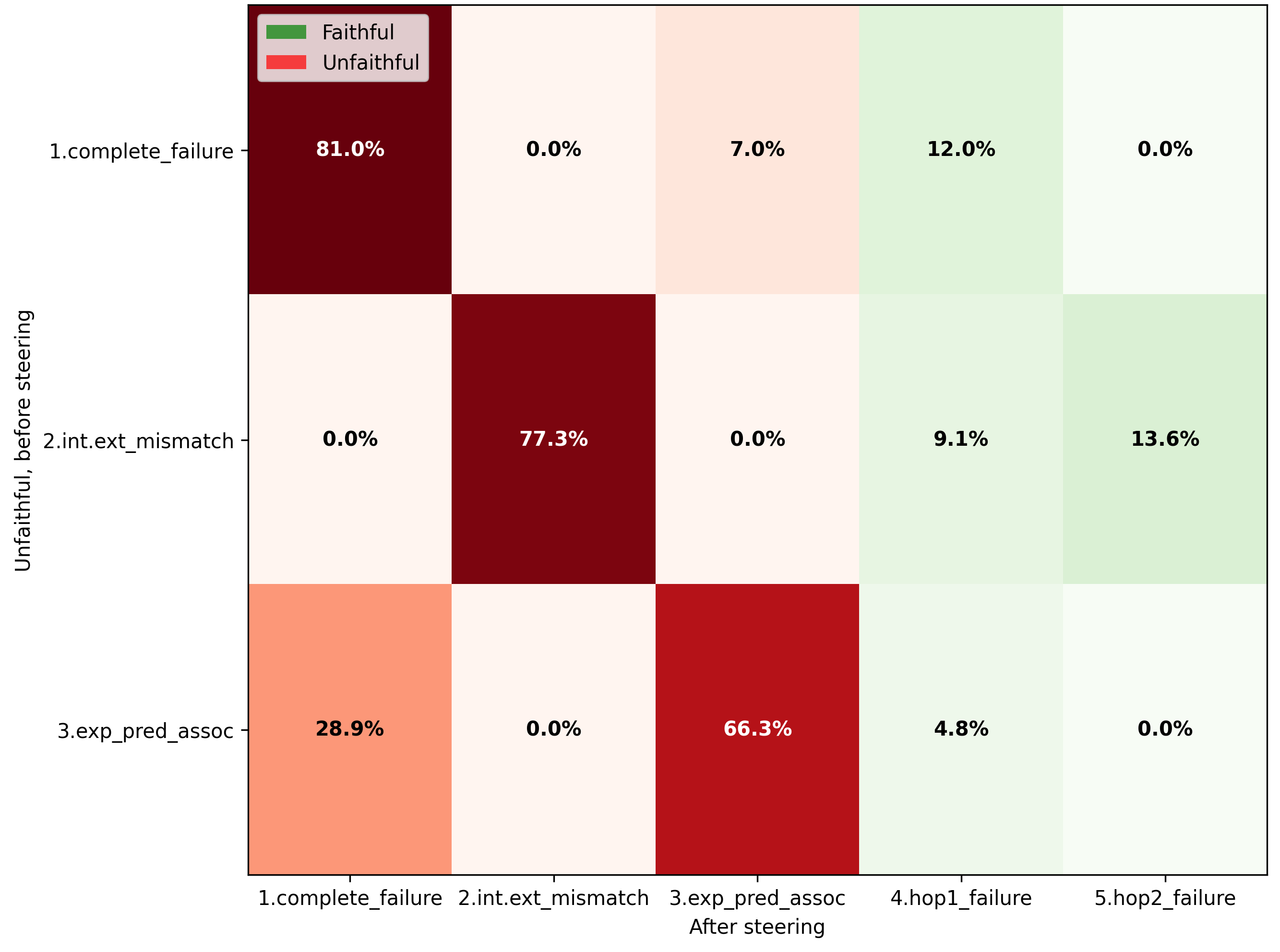}
        \caption{Initially inaccurate predictions.}
        \label{fig:chart1}
    \end{subfigure}
    \hfill
    \begin{subfigure}{0.45\textwidth}
        \centering
        \includegraphics[width=\linewidth]{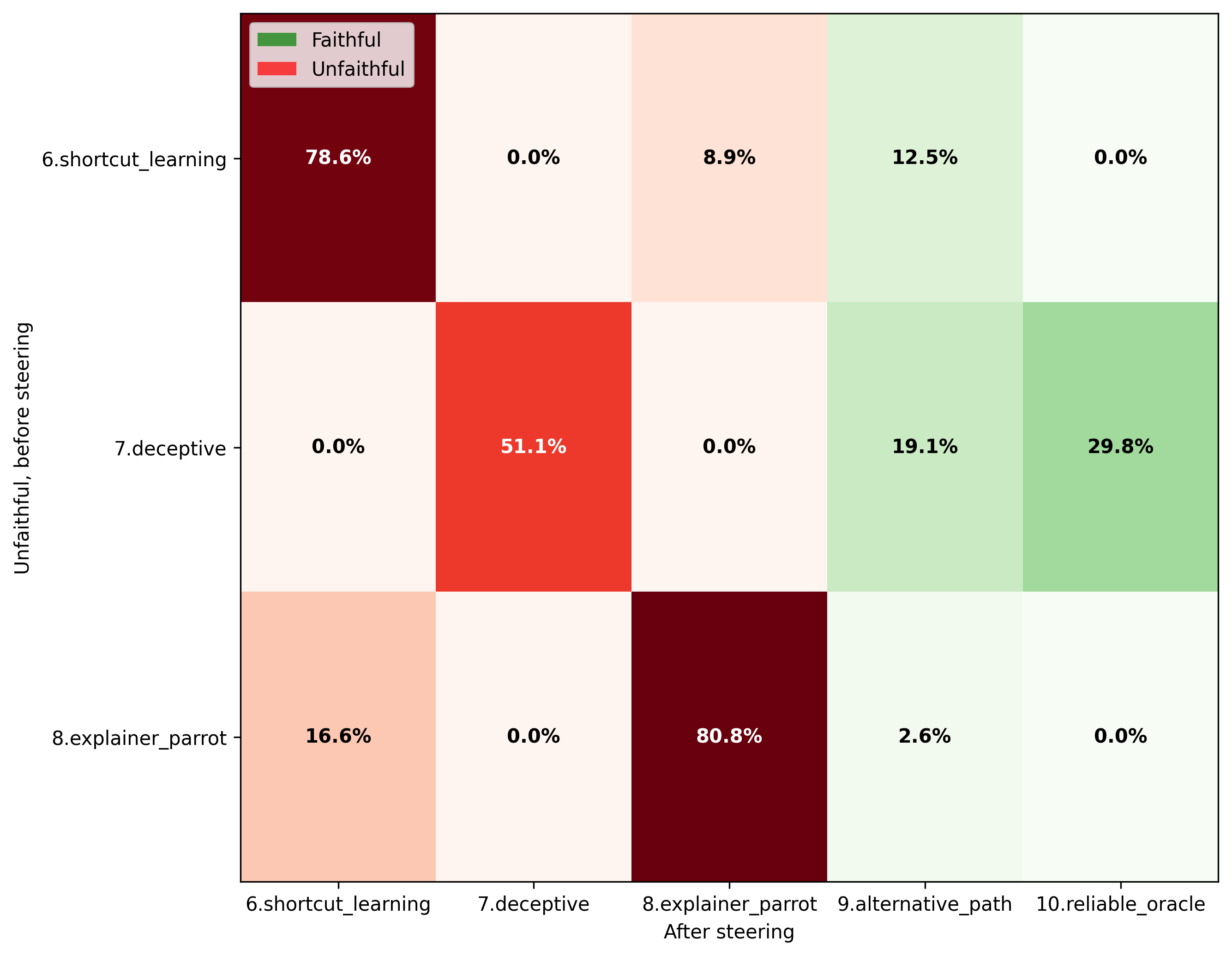}
        \caption{Initially accurate predictions.}
        \label{fig:chart2}
    \end{subfigure}
    \caption{Detailed taxonomy transition state analysis, before and after hallucination inhibition on \texttt{gemma-2-2b} on 2-hop reasoning.}
    \label{fig:transition_27b_hallucination}
\end{figure}

\subsection{\method\ Faithfulness Measure Comparison}
\label{sec:sota_comparison}

In this section we examine how faithfulness as measured by \method\ compares to existing Counterfactual Intervention (CI)~\cite{atanasova_faithfulness_2023} and Attribution Agreement (AA) approaches~\cite{parcalabescu_faithfulness} for evaluating LLM self-NLE faithfulness. We focus on 2-hop reasoning and begin with a qualitative comparison that illustrates potential differences between these approaches, followed by a quantitative analysis that provides evidence for these observations.

\subsubsection{2-hop Reasoning Faithfulness Qualitative Comparison.}
\label{sec:appendix_quantitative_comparison}

\begin{figure*}[]{\centering}
\begin{center}
\includegraphics[width=0.8\linewidth]{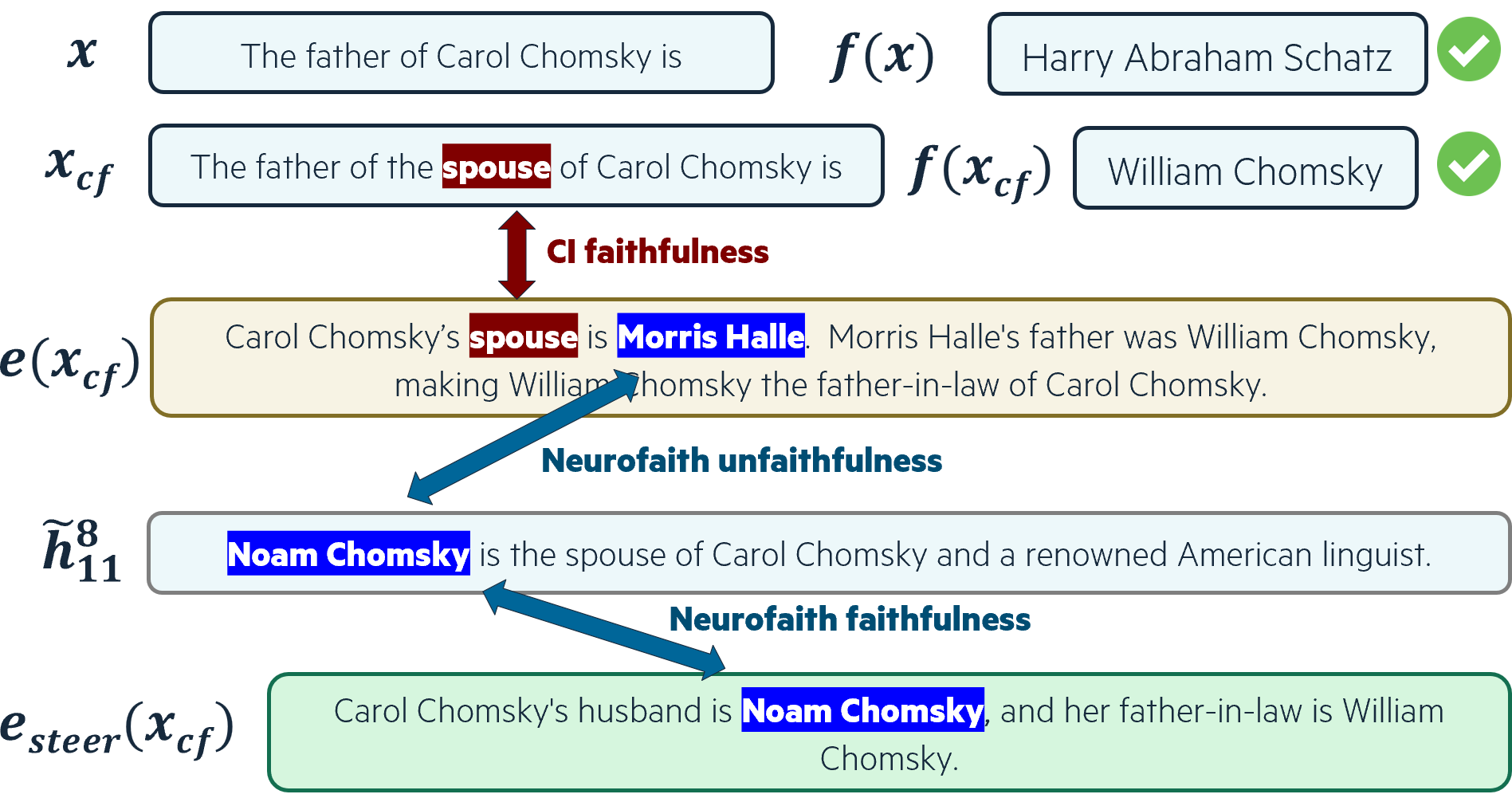}
\caption{\label{fig:sota_comparaison}
Qualitative comparison between \method\ and CI faithfulness.}
\end{center}
\end{figure*}

To illustrate the critical differences between existing faithfulness evaluation approaches and \method, we examine a concrete 2-hop reasoning example (see Figure~\ref{fig:sota_comparaison}) that illustrates fundamental limitations in current methods (CI and AA). Given input text $x$, prediction  $f(x)$, and self-generated explanation $e(x)$, we compare how different faithfulness approaches evaluate the same case. We analyse the following example: 
\begin{itemize}
    \item Input: $x$ = "\textit{The father of Carol Chomsky is}"
    \item Prediction: $f(x)$ = "\textit{Harry Abraham Schatz}"
    \item Counterfactual Intervention: $x_{cf}$ = "\textit{The father of the spouse of Carol Chomsky is}"
    \item Counterfactual prediction: $f(x_{cf})$ = "\textit{William Chomsky}"
\end{itemize}

Counterfactual Intervention (CI) would assess the self-NLE $e(x_{cf})$ as faithfulness. This evaluation is based solely on the presence of the intervention term "spouse" within $e(x_{cf})$, establishing consistency between the input modification and explanation content. 
Attribution Analysis (AA) methods consists in comparing attribution scores (e.g., using SHAP) to highlight important tokens (e.g. "father", "spouse", "Carol", "Chomsky") for the prediction and the self-NLE. High correlation coefficients between the attribution vectors for prediction and self-NLE would similarly classify the self-NLE as faithful. 

On the contrary, \method\ rejects $e(x_{cf})$ as unfaithful because the bridge object ("Morris Halle") is not contained in the decoded  hidden states ($\{\tilde{h^\ell_{k}}\}$). This mismatch between internal computation and self-NLE content reveals unfaithfulness. The steered self-NLE obtained with linear faithful steering (see Section~\ref{sec:linear_detection}) is evaluated as faithful by \method, due to shared bridge object ("Noam Chomsky") between the new self-NLE and the decoded hidden states.

This qualitative example illustrates that both CI and AA approaches may employ more lenient evaluation standards compared to \method\ and can miss unfaithful self-NLE. This observation is corroborated by our analysis in the next paragraph.

\subsubsection{2-hop Reasoning Faithfulness Quantitative Comparison.}
\label{sec:appendix_quantitative_comparison}

We propose an experimental protocol to evaluate the practical utility of different faithfulness measures for model analysis. The protocol is motivated by two common applications of explanations in AI systems: (1) troubleshooting models by identifying the source of errors in wrong predictions~\cite{biecek2024position}, and (2) detecting potential biases or shortcuts in correct predictions to ensure they align with expected reasoning processes~\cite{ribeiro2016should}.

We test whether faithful explanations, as identified by \method\ and CI methods, provide better diagnostic information for these purposes. The key hypothesis is that if a faithfulness measure accurately captures the model's internal reasoning, then explanations deemed faithful should better localize reasoning failures and identify non-canonical reasoning pathways.

\paragraph{Experimental Framework.} 

\begin{figure*}[t]{\centering}
\begin{center}
\includegraphics[scale=0.35]{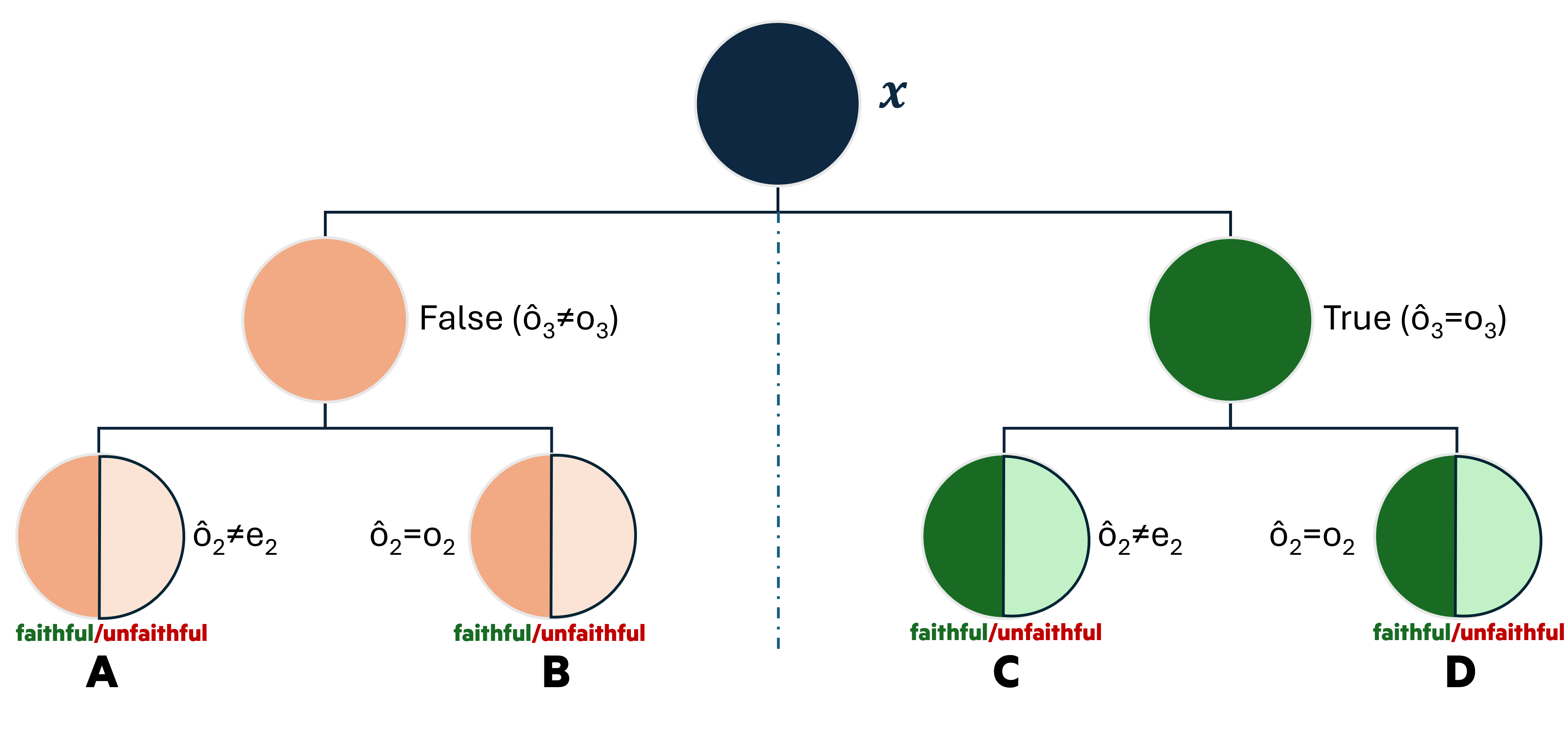}
\caption{Simplified taxonomy of $f$ behavior in two-hop reasoning, based on the status of the prediction and the self-NLE.}  
\label{fig:simplified_taxonomy}
\end{center}
\end{figure*}

Given an input text $x= (o_{1},r_1,\tiny{\blacktriangle{}},r_2,\bullet)$ requiring 2-hop reasoning and the model's reasoning trace $(o_{1},r_1,\widehat{o_{2}}, r_2,\widehat{o_{3}})$, we categorize the model behavior using a simplified taxonomy (see~Figure~\ref{fig:simplified_taxonomy}) based on prediction correctness and bridge object accuracy: \begin{itemize}
    \item Category A: Wrong prediction ($\widehat{o}_{3}\neq o_3$), wrong bridge object ($\widehat{o}_{2}\neq o_2$) $\rightarrow$ likely first-hop failure
    \item Category B: Wrong prediction ($\widehat{o}_{3}\neq o_3$), correct bridge object ($\widehat{o}_{2} = o_2$) $\rightarrow$ likely second-hop failure
    \item Category C: Correct prediction ($\widehat{o}_{3} = o_3$), wrong bridge object ($\widehat{o}_{2}\neq o_2$) $\rightarrow$ alternative reasoning pathway
    \item Category D: Correct prediction ($\widehat{o}_{3} = o_3$), correct bridge object ($\widehat{o}_{2} = o_2$) $\rightarrow$ canonical reasoning
\end{itemize}
Each category can be further subdivided with respect to faithfulness, enabling comparison between faithful and unfaithful self-NLE within each reasoning pattern. In the following we propose 3 evaluation protocols based on these 4 categories and assess if faithful self-NLE lead to better model debugging and bias targeting than unfaithful self-NLE.

\paragraph{First Hop Hint.} 
We test whether faithful self-NLE better identify first-hop reasoning failures through a targeted intervention. For each input $x$ having led to a wrong prediction ($\widehat{o}_{3}\neq o_3$), we create a modified version $x_{hint1} = (o_{1},r_1,o_{2},x)$ that explicitly provides the correct bridge object. For example: \begin{itemize}
    \item $x = $ "The country of origin of the movie maker that directed Persona is"
    \item $x_{hint1} = $  "The movie maker that directed Persona is Ingmar Bergman. The country of origin of the movie maker that directed Persona is"
\end{itemize}

If faithful self-NLE accurately reflect internal reasoning, then providing first-hop hints should differentially improve performance for Category A (first-hop failures) versus Category B (second-hop failures), and this difference should be stronger for faithful self-NLE. We define the performance ratio under first-hop hints as: \begin{equation}
PR(A,B,hint1) = \frac{ACC(A,hint1)}{ACC(B,hint1)}
\end{equation} where  $ACC(A,hint1)$ represents the accuracy of Category A examples when given first-hop hints. We compute separate ratios for faithful and unfaithful explanations:  \begin{equation}
PR(A,B,hint1,faithful) = \frac{ACC(A,hint1, faithful)}{ACC(B,hint1, faithful)}
\end{equation} Finally, the Compound Accuracy Score (CAS) quantifies whether faithful explanations provide better error localization: \begin{equation}
CAS(A,B,hint1) = \log\Bigl(\frac{PR(A,B,hint1,faithful)}{PR(A,B,hint1,unfaithful) }\Bigr)
\end{equation}

A positive CAS indicates that faithful self-NLE better identify first-hop failures. This metric is conceptually close to the In-Context Editing instantiation of the approach proposed by~\citet{zaman-srivastava-2025-causal} to compare faithfulness measures. In the following, we extend beyond this proposal with two other metrics. 

\begin{table*}[!htbp]
\centering
\small

\begin{tabular*}{\textwidth}{@{\extracolsep{\fill}}lcccccccc@{}}
\toprule
\textbf{Quality Metric} & \multicolumn{3}{c}{\texttt{gemma-2-2b}} & \multicolumn{3}{c}{\texttt{gemma-2-9b}} & \multicolumn{2}{c}{\texttt{gemma-2-27b}} \\
\cmidrule(lr){2-4} \cmidrule(lr){5-7} \cmidrule(lr){8-9}
& \method\ & CI & AA & \method\ & CI & AA & \method\ & CI \\
\midrule
hint1 & \textbf{0.04} & -0.07 & -0.23 & 0.06 & -1.10 & \textbf{0.40} & \textbf{0.75} & -0.99 \\
hint2 & \textbf{-0.44} & -0.55 & -1.23 & \textbf{0.28} & -0.03 & 0.15 & -0.35 & \textbf{1.25} \\
$r_2 \rightarrow r'_2$ & 0.09 & -2.62 & \textbf{1.32} & \textbf{0.17} & -0.32 & 0.01 & \textbf{0.31} & -0.26 \\
\bottomrule
\end{tabular*}
\caption{\method\ comparison to CI and AA (higher is better) across models on 352 CI-compatible samples on 2-hop reasoning.}
\label{tab:eval_overall_faithfulness_sota}
\vspace{1.5em}

\begin{tabular*}{\textwidth}{@{\extracolsep{\fill}}lccc@{}}
\toprule
\textbf{Quality Metric} & \texttt{gemma-2-2b} & \texttt{gemma-2-9b} & \texttt{gemma-2-27b} \\
\midrule
hint1 & 0.48 & 0.14 & 0.22 \\
hint2 & 0.03 & 0.33 & 0.88 \\
$r_2 \rightarrow r'_2$ & -0.04 & 1.18 & 0.40 \\
\bottomrule
\end{tabular*}
\caption{\method\ evaluation across models on the overall 2-hop reasoning dataset.}
\label{tab:eval_overall_faithfulness}
\vspace{1.5em} 
\end{table*}


\paragraph{Second Hop Hint.} 
We then test whether faithful self-NLE better identify second-hop reasoning failures through a targeted intervention. Following the logic introduced above, for each input $x$ having led to a wrong prediction ($\widehat{o}_{3}\neq o_3$), we create a modified version $x_{hint2} = (o_{2},r_2,o_{3},x)$ that explicitly provides the second part of the 2-hop reasoning. For example: \begin{itemize}
    \item $x = $ "The country of origin of the movie maker that directed Persona is"
    \item $x_{hint2} = $  "The country of origin of Ingmar Bergman is Sweden. The country of origin of the movie maker that directed Persona is"
\end{itemize}

If faithful self-NLE accurately reflect internal reasoning, then providing second-hop hints should differentially improve performance for Category B (second-hop failures) versus Category A (first-hop failures), and this difference should be stronger for faithful self-NLE. We build our second metric following the notations introduced above, based on the Compound Accuracy Score: \begin{equation}
CAS(B,A,hint2) = log\Bigl(\frac{PR(B,A,hint2,faithful)}{PR(B,A,hint2,unfaithful) }\Bigr)
\end{equation}

Here, a positive CAS indicates that faithful self-NLE better identify second-hop failures.

\paragraph{Second Relation Modification.}
We test whether faithful self-NLE better identify non-canonical reasoning pathways through the modification of the second step of the reasoning trace ($r_2$). Intuitively, a non-canonical reasoning pathway is more prone to leading to false reasoning when changing one step of the 2-hop reasoning. For each input $x$ having led to a correct prediction ($\widehat{o}_{3} = o_3$), we create a modified version $x_{r_2 \rightarrow r'_2} = (o_{1},r_1,\tiny{\blacktriangle{}},r_2',\bullet)$ that changes the second relation of the 2-hop reasoning input. For example: \begin{itemize}
    \item $x = $ "The country of origin of the movie maker that directed Persona is"
    \item $x_{r_2 \rightarrow r'_2} = $  "The father of the movie maker that directed Persona is"
\end{itemize}

If faithful self-NLE accurately reflect internal reasoning, then changing $r_2$ should significantly decrease performance for Category C (alternative reasoning pathway) versus Category D (canonical reasoning), and this difference should be stronger for faithful self-NLE. We build our third metric following the notations introduced above, based on the Compound Accuracy Score: \begin{equation}
CAS(D,C,r_2 \rightarrow r'_2) = log\Bigl(\frac{PR(D,C,r_2 \rightarrow r'_2,faithful)}{PR(D,C,r_2 \rightarrow r'_2,unfaithful) }\Bigr)
\end{equation}

A positive CAS indicates that faithful self-NLE better identify non-canonical reasoning pathways.

\paragraph{Experimental Results.}

We evaluate our three faithfulness indicators to compare \method\ to commonly used Attribution Agreement (AA) and Counterfactual Intervention (CI) methods. The Wikidata-2-hop dataset provides natural support for computing $x_{r_2 \rightarrow r'_2}$ instances, as it contains multiple reasoning chains involving similar objects. Additionally, the dataset includes counterfactual interventions through variations in reasoning chains, enabling straightforward CI computation as in~\citet{atanasova_faithfulness_2023} across multiple instances. We implement AA based on gradient-based attributions as in~\citet{wiegreffe_faithfulness}. Due to prohibitive cost for 27-billion models, we only apply AA to \texttt{Gemma-2-2B} and \texttt{Gemma-2-9B}.

Our evaluation reveals 2 key findings. First, Table~\ref{tab:eval_overall_faithfulness_sota} shows \method\ outperforms CI and AA on average on the subset of CI-compatible samples. Second, Table~\ref{tab:eval_overall_faithfulness} presents \method\ results on the complete dataset, showing positive metric values across all models except \texttt{gemma-2-2b}.

\subsection{Classification Examples}
\label{sec::examples_classification}
This section gives two examples of instances characterized as either faithful or unfaithful in classification case.

\begin{figure}[H]  
    \centering
    \includegraphics[width=0.3\linewidth]{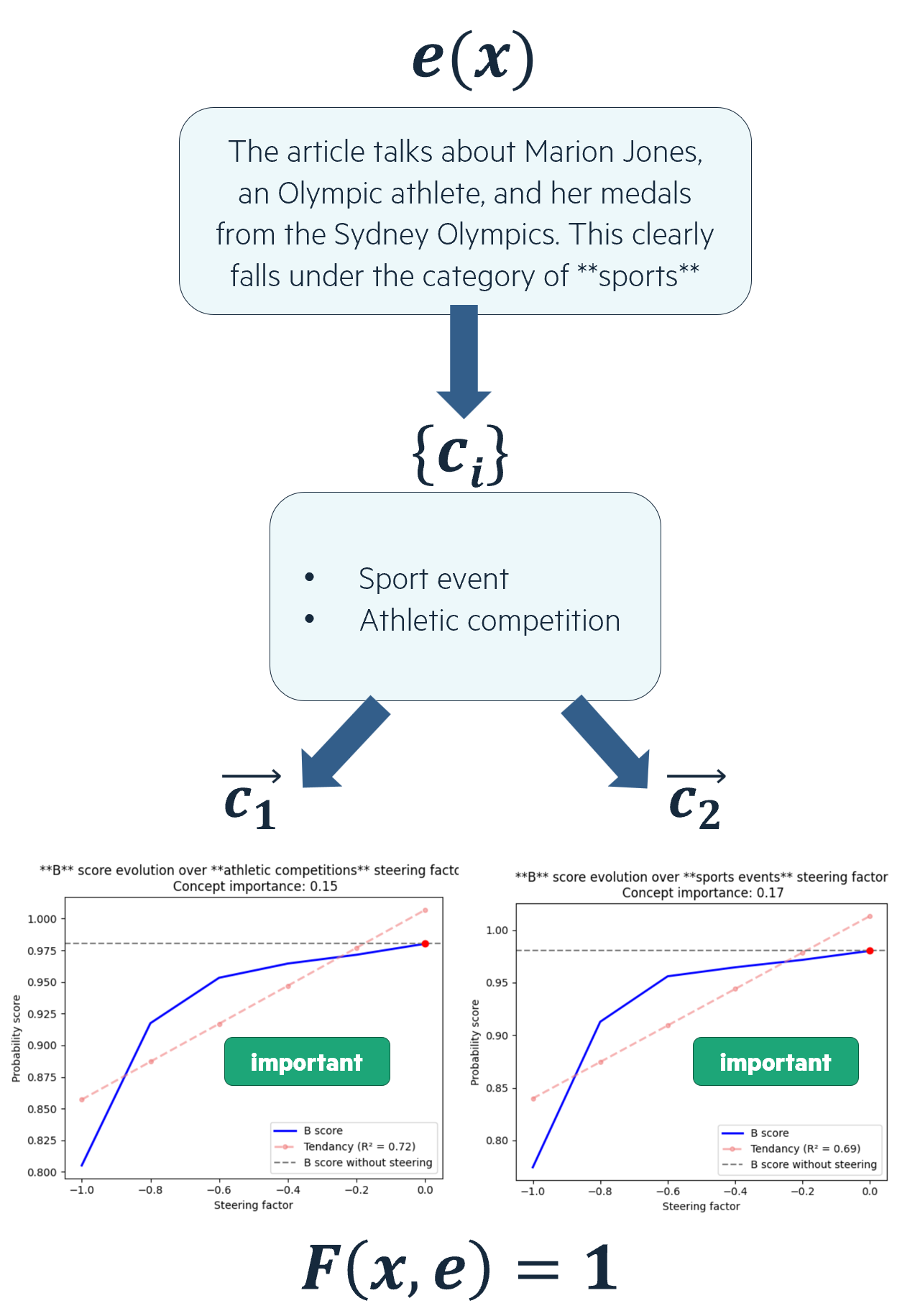}
    \caption{Example from the AGNews dataset where we detect "sport event" and "athletic competition" as relevant concepts from the explanation. These two concepts are assessed as important for the prediction, making this explanation faithful ($F(x,e)=1$).}
    \label{fig:label}
\end{figure}

\begin{figure}[H]  
    \centering
    \includegraphics[width=0.5\linewidth]{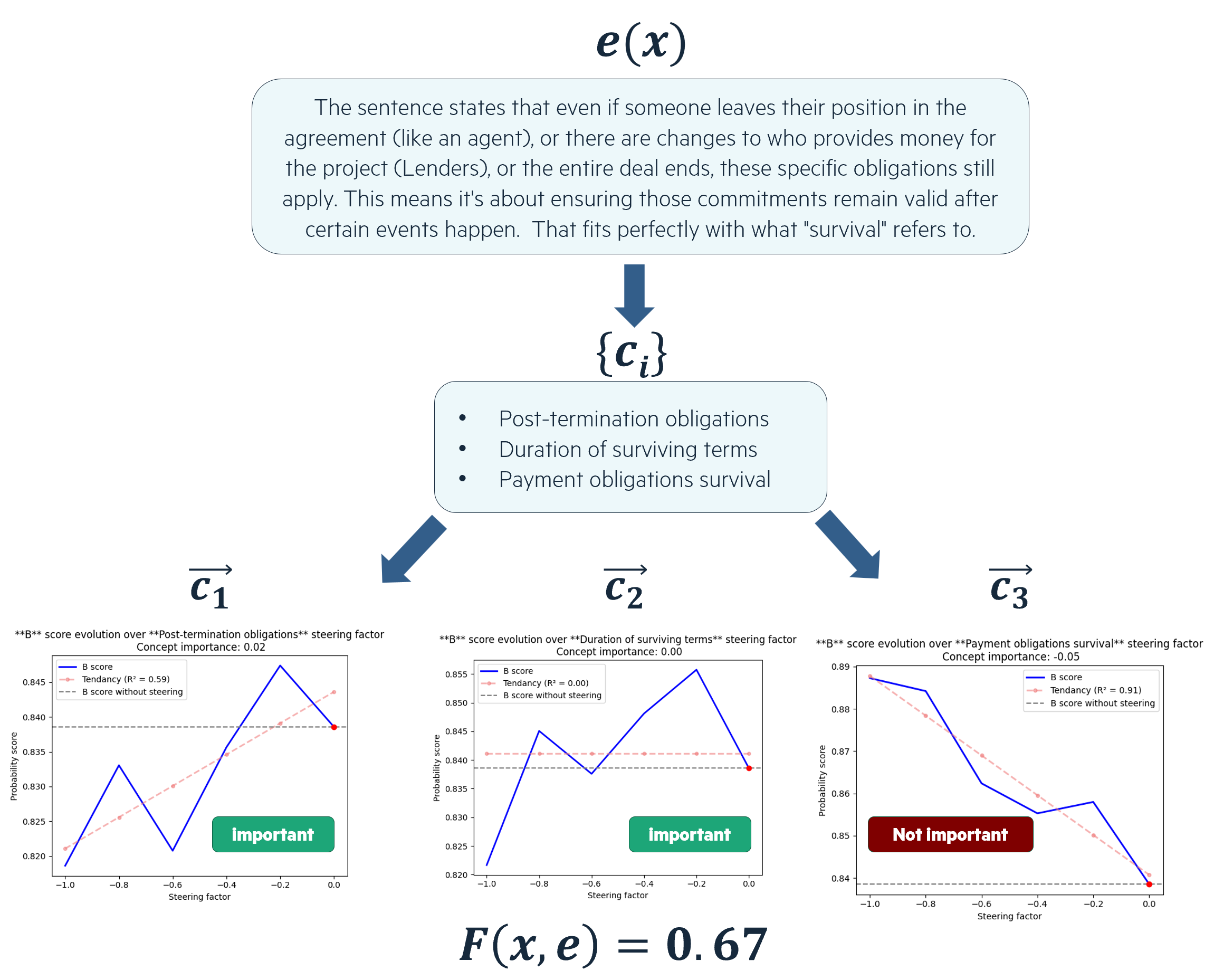}
    \caption{Example from the Ledgar dataset where we detect "post-termination obligations", "duration of surviving terms" and "payment obligations survival" as relevant concepts from the explanation. Two concepts over three are assessed as important, giving an explanation faithfulness score at 0.67.}
    \label{fig:label}
\end{figure}

\subsection{2-hop Reasoning Taxonomy Examples}
\label{sec::examples}
In this section we give examples of characterized instances based on the taxonomy introduced in Appendix~\ref{sec::caract}. We also give examples of unfaithful self-NLE made faithful, characterized by the same taxonomy.

\begin{figure}[H]  
    \centering
    \includegraphics[width=0.75\linewidth]{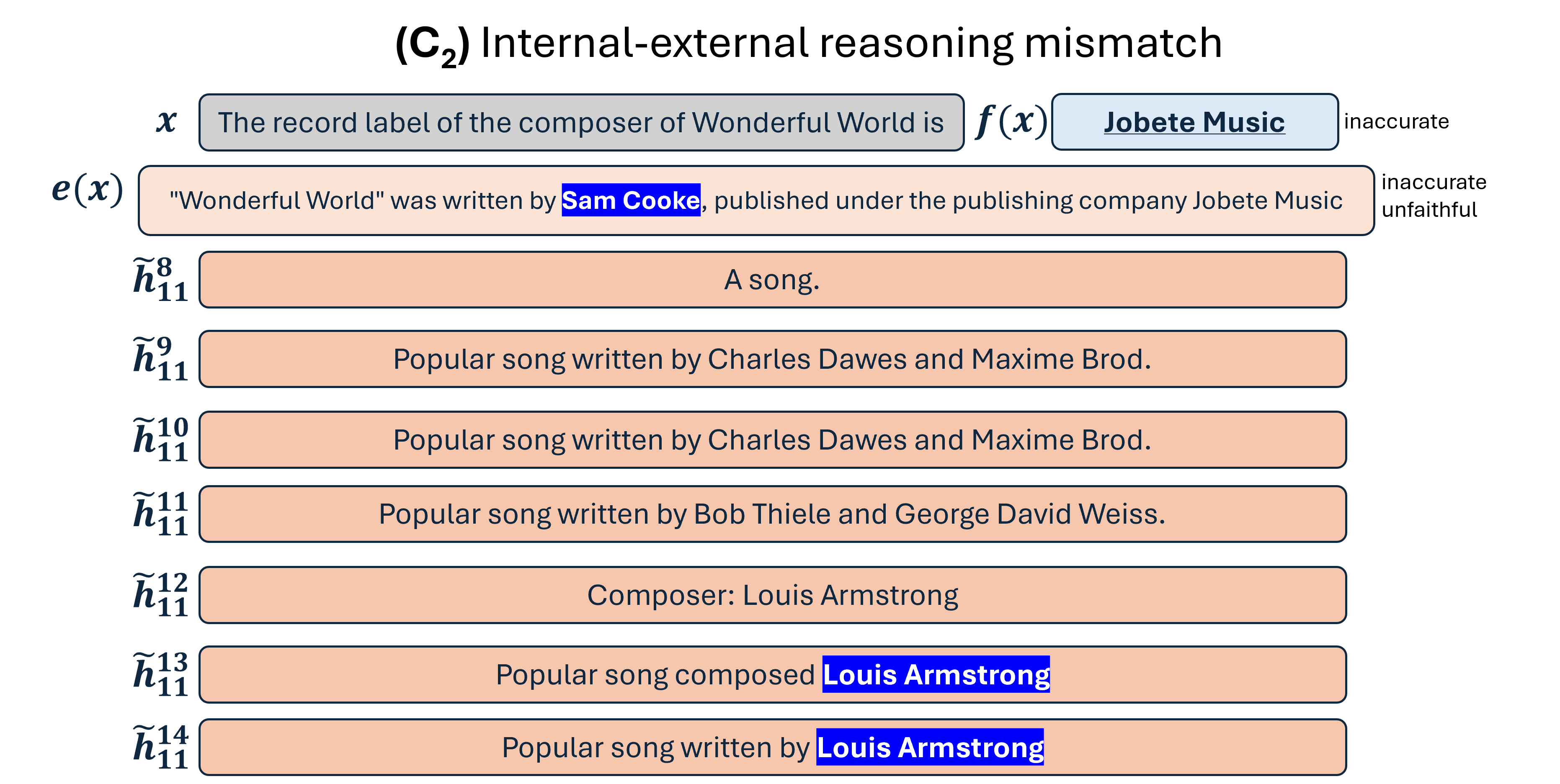}
    \caption{Example from the Wikidata-2-hop dataset where we observe an internal-external reasoning mismach. The model incorrectly answers "Jobete Music" and provides a self-NLE referencing to the bridge object "Sam Cooke". This bridge object is false and does not appear in the set of natural language interpretations of $f$ latent states. However, the ground truth bridge object "Louis Armstrong" is decoded at token index 11 and layers 13 and 14.}
    \label{fig:label}
\end{figure}

\begin{figure}[H]  
    \centering
    \includegraphics[width=0.75\linewidth]{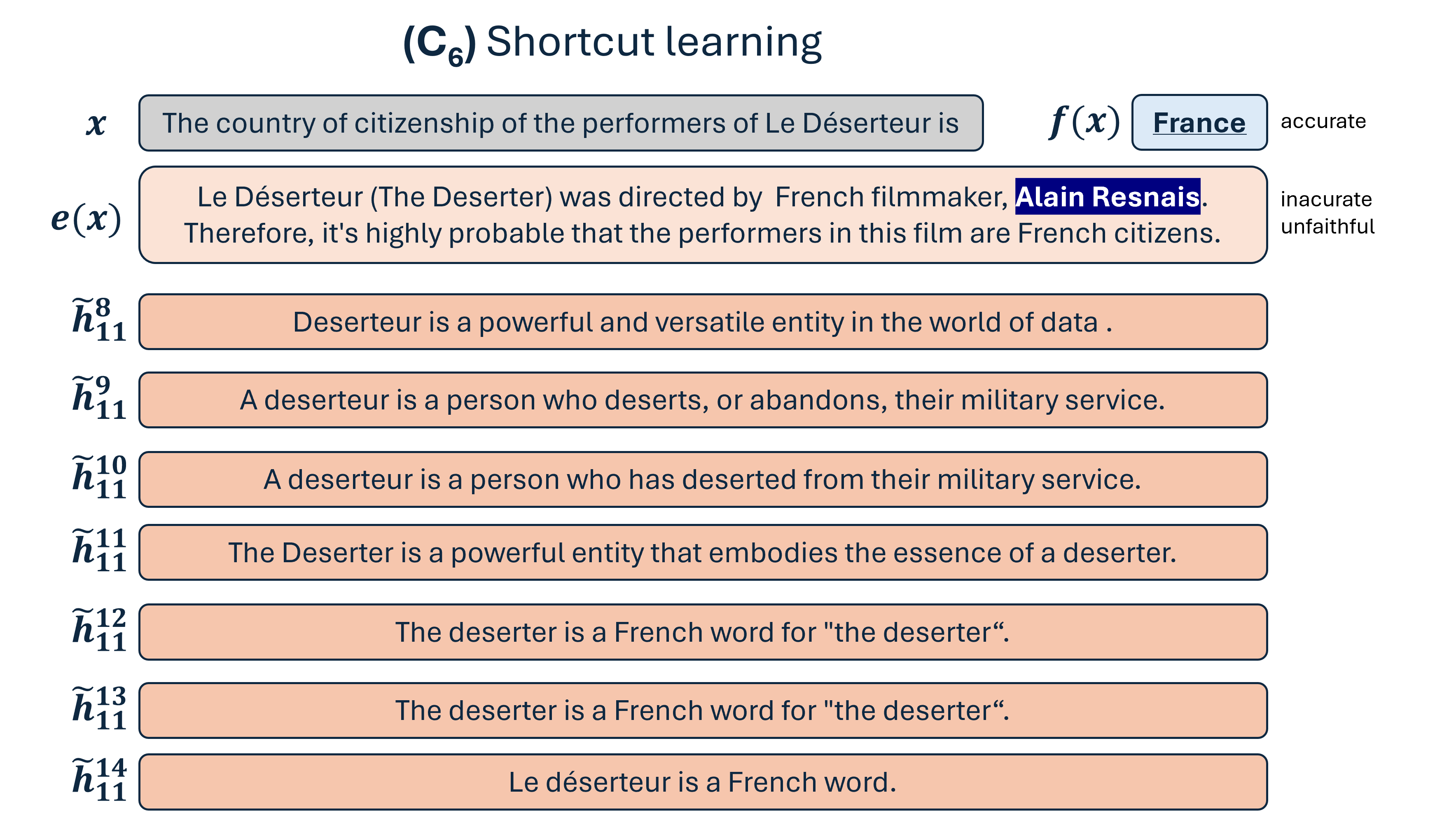}
    \caption{Example from the Wikidata-2-hop dataset where we observe shortcut learning. The model correctly answers "France" and provides a self-NLE referencing to "Alain Resnais". This bridge object is incorrect and does not appear in the set of natural language interpretations of $f$ latent states.}
    \label{fig:label}
\end{figure}

\begin{figure}[H]  
    \centering
    \includegraphics[width=0.75\linewidth]{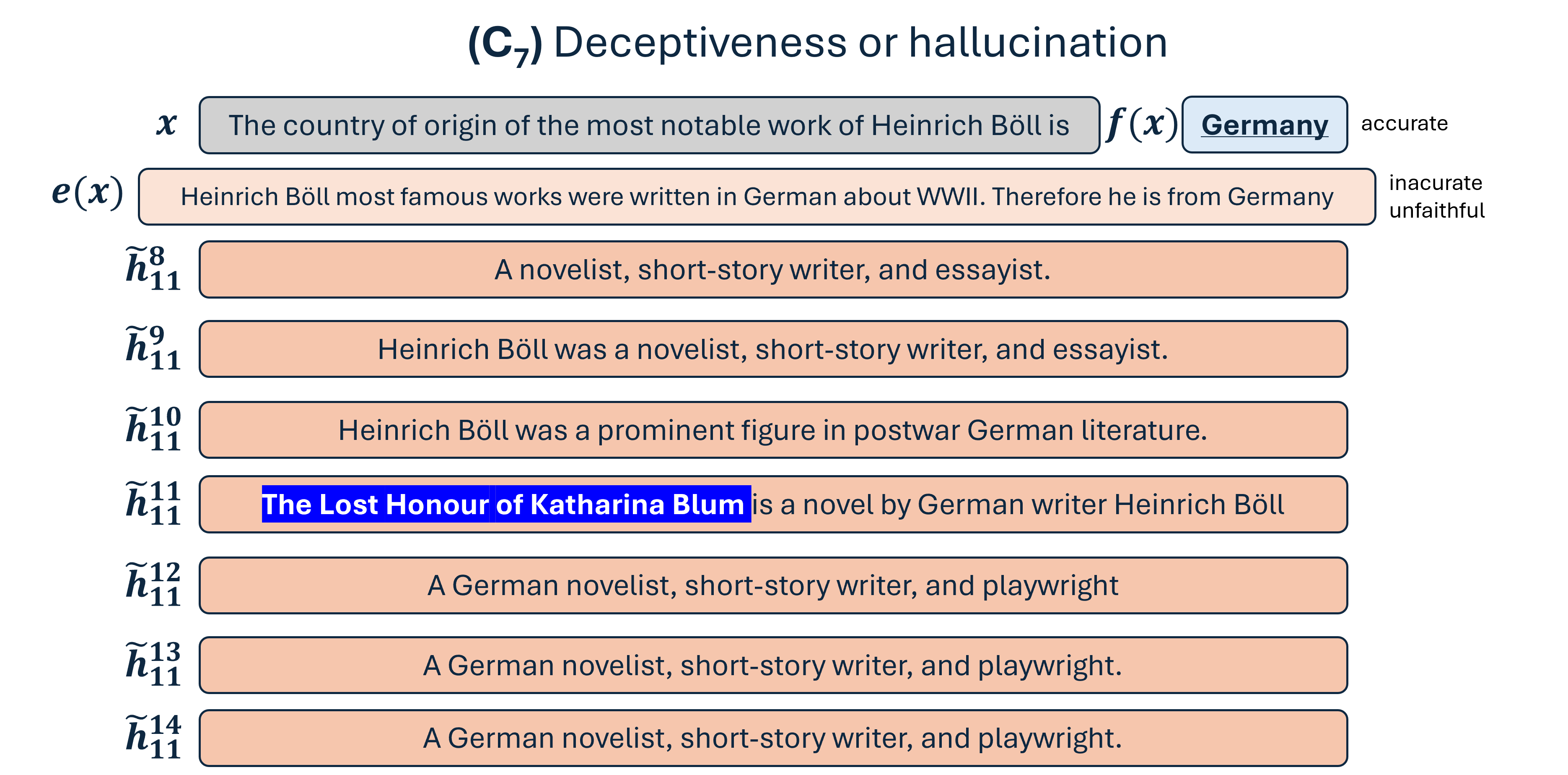}
    \caption{Example from the Wikidata-2-hop dataset where we observe shortcut learning. The model correctly answers "Germany" without providing any bridge object in its self-NLE. The expected bridge object "The Lost Honour of Katharina Blum" is however decoded from the representation space.}
    \label{fig:label}
\end{figure}

\begin{figure}[H]  
    \centering
    \includegraphics[width=0.75\linewidth]{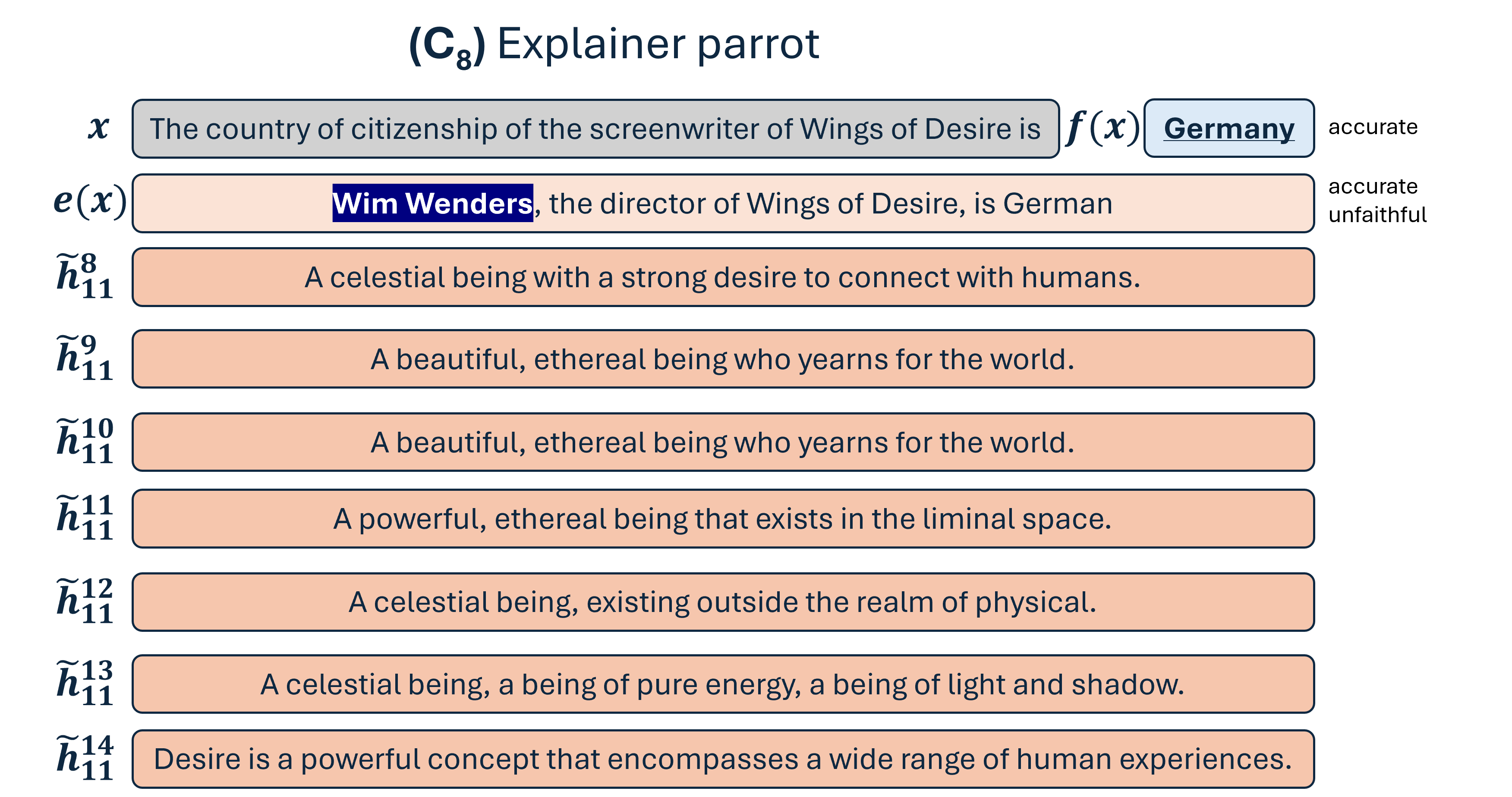}
    \caption{Example from the Wikidata-2-hop dataset where we observe an explainer parrot case. The model correctly answers "Germany" and provides a self-NLE referencing to "Wim Wenders". This bridge object is correct but does not appear in the set of natural language interpretations of $f$ latent states.}
    \label{fig:label}
\end{figure}

\begin{figure}[H]  
    \centering
    \includegraphics[width=0.75\linewidth]{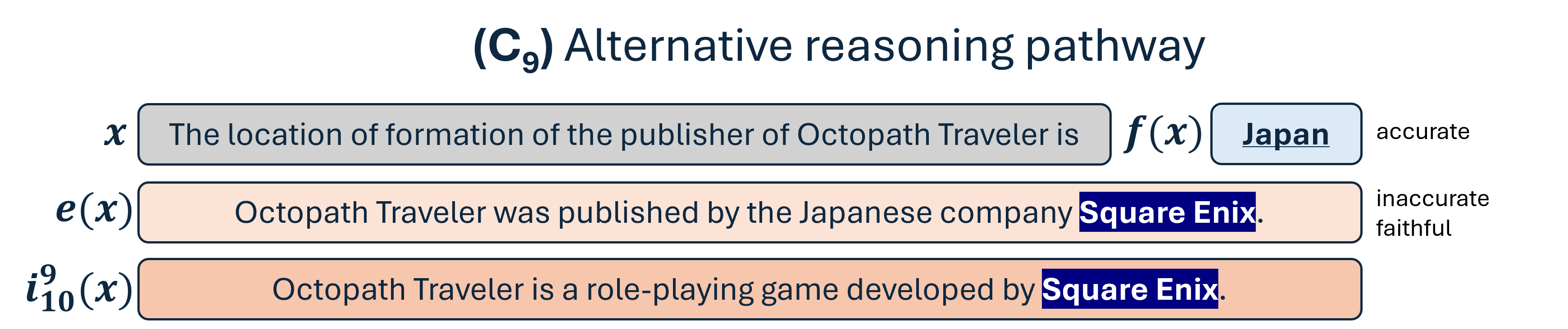}
    \caption{Example from the Wikidata-2-hop dataset where we observe an alternative reasoning pathway. The model correctly answers "Japan" and provides a self-NLE referencing to "Square Enix". This bridge object is incorrect and also appears in the set of natural language interpretations of $f$ latent states.}
    \label{fig:label}
\end{figure}

\begin{figure}[H]  
    \centering
    \includegraphics[width=0.75\linewidth]{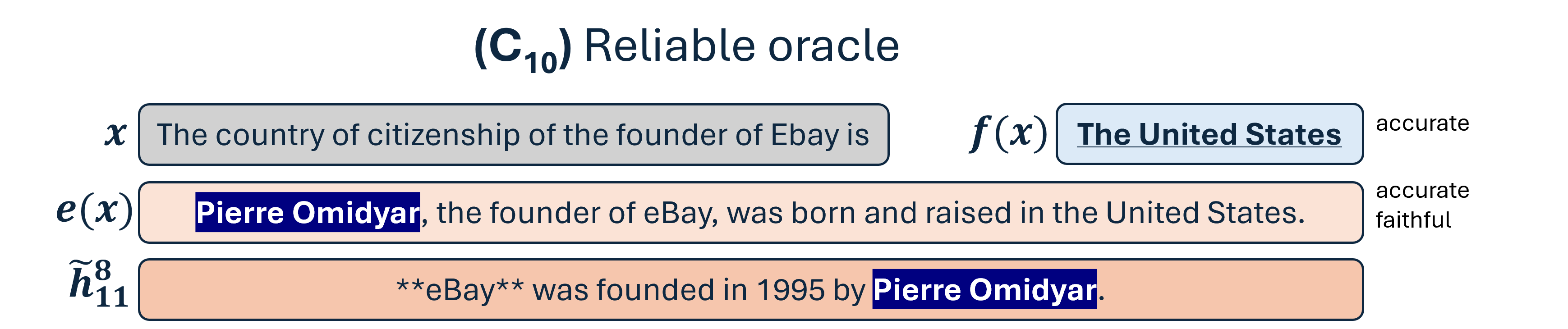}
    \caption{Example from the Wikidata-2-hop dataset where we observe an alternative reasoning pathway. The model correctly answers "USA" and provides a self-NLE referencing to "Pierre Omidyar". This bridge object is correct and also appears in the set of natural language interpretations of $f$ latent states.}
    \label{fig:label}
\end{figure}

\begin{figure}[H]  
    \centering
    \includegraphics[width=0.75\linewidth]{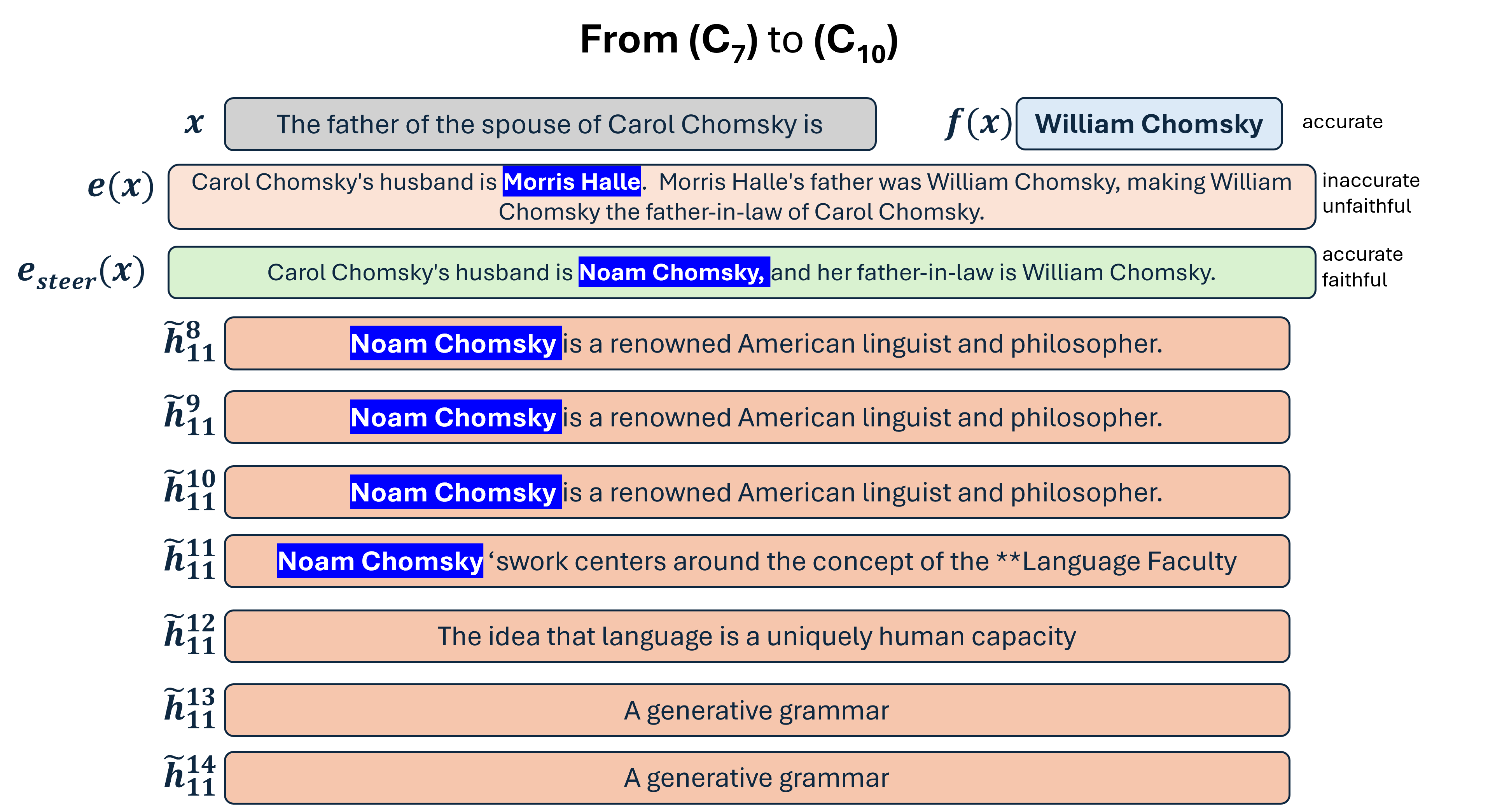}
    \caption{Example from the Wikidata-2-hop dataset where we start from an incorrect unfaithful explanation and go to a correct and faithful explanation through \method\ linear faithfulness steering.}
    \label{fig:label}
\end{figure}

\begin{figure}[H]  
    \centering
    \includegraphics[width=0.75\linewidth]{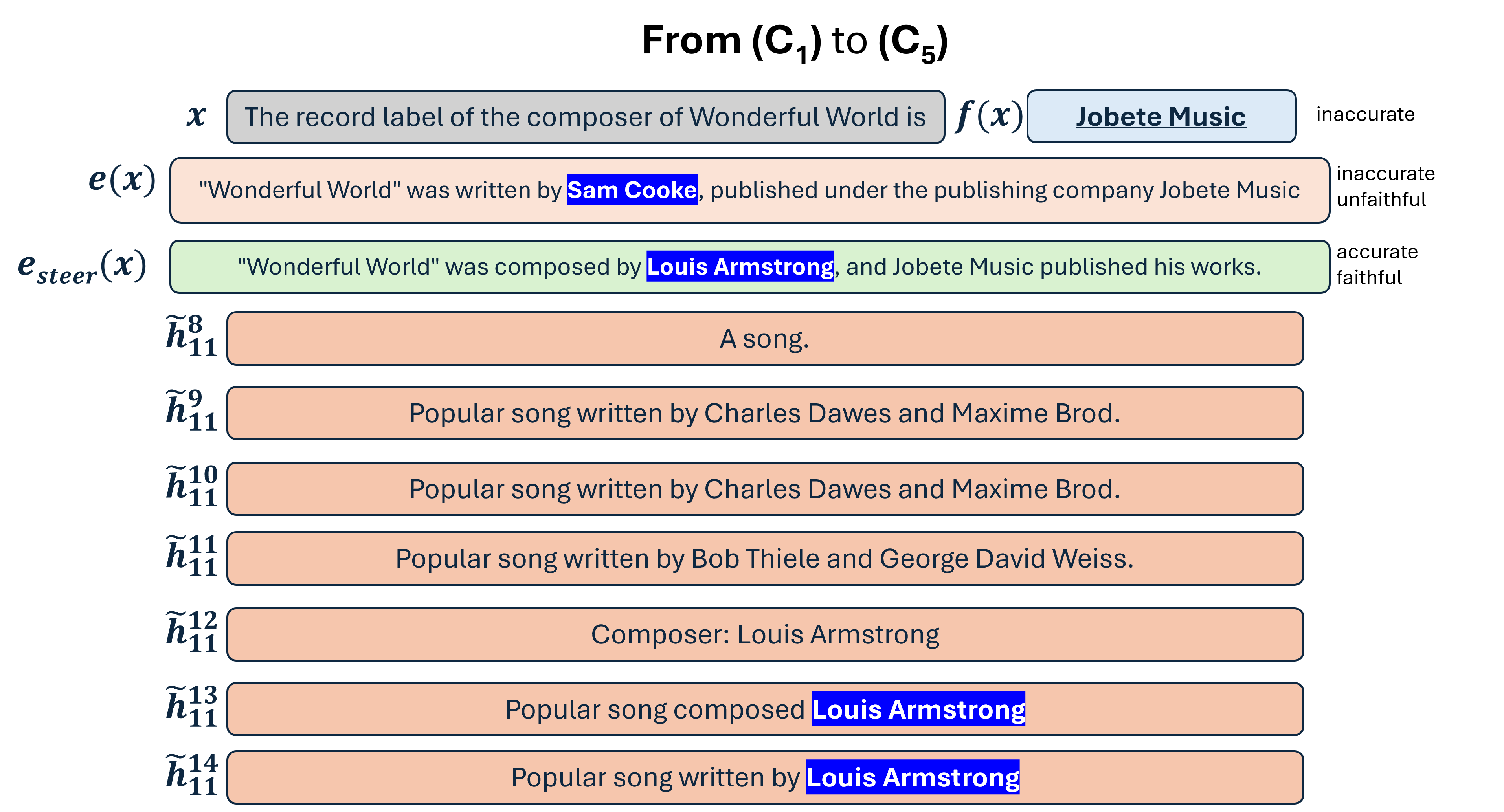}
    \caption{Example from the Wikidata-2-hop dataset where we start from an incorrect unfaithful explanation and go to a correct and faithful explanation through \method\ linear faithfulness steering.}
    \label{fig:label}
\end{figure}

\begin{figure}[H]  
    \centering
    \includegraphics[width=0.75\linewidth]{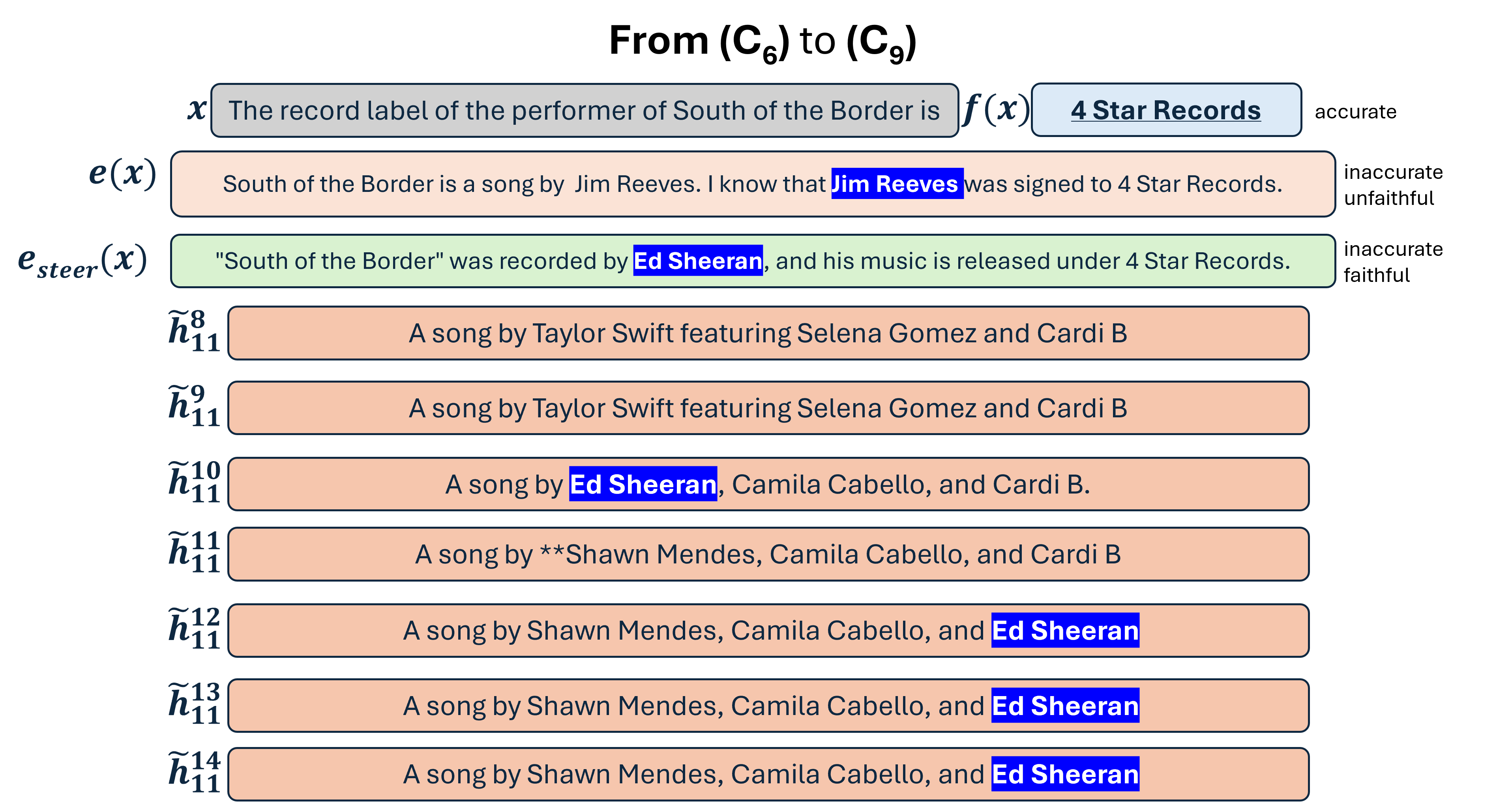}
    \caption{Example from the Wikidata-2-hop dataset where we start from an incorrect unfaithful explanation and go to a still incorrect but faithful explanation through \method\ linear faithfulness steering.}
    \label{fig:label}
\end{figure}


\end{document}